%% file: main.tex
\documentclass[lettersize,journal]{IEEEtran}

\IEEEoverridecommandlockouts                            
                          
\usepackage{color}
\usepackage{colortbl}
\usepackage{graphicx} 
\usepackage[american]{babel}
\usepackage{subfigure}		
\usepackage{balance}
\usepackage{cite}
\usepackage{makecell}
\usepackage{xcolor}
\usepackage{hyperref}
\usepackage{amsmath}
\usepackage{bbding}
\usepackage{bm}
\usepackage{fontawesome}
\usepackage{nicematrix}
\usepackage{array}
\usepackage{dblfloatfix}   
\usepackage{stackengine}

\usepackage{amssymb}

\usepackage{algorithm}
\usepackage{algpseudocode}
\usepackage{multirow} 
\newcommand{\head}[2]{\multicolumn{1}{>{\centering\arraybackslash}p{#1}}{\textbf{#2}}}

\title{\LARGE \bf
CineMPC: A Fully Autonomous Drone Cinematography System \\Incorporating Zoom, Focus, Pose, and Scene Composition
}

\author{\centering Pablo Pueyo, Juan Dendarieta, Eduardo Montijano, Ana C. Murillo and Mac Schwager
\thanks{This work was supported by a DGA scholarship and  project  T45\_23R; Spanish projects PID2019-105390RB-I00 and PID2021-125514NB-I00, funded by MCIN/AEI/10.13039/501100011033, by ERDF A way of making Europe and by the European Union NextGenerationEU/PRTR, and by ONR grant N00014-18-1-2830.}
\thanks{P. Pueyo, J. Dendarieta, E. Montijano and A. C. Murillo are associated with the Instituto de Investigaci\'on en Ingenier\'ia de Arag\'on and DIIS, Universidad de Zaragoza, Spain 
\texttt{\small \{ppueyor, 538123, emonti, acm\}@unizar.es}}
\thanks{M. Schwager is associated with Dept. of Aeronautics and Astronautics, Stanford University, USA
\texttt{\small \{schwager\}@stanford.edu}}
}

\begin{document}

\maketitle
\thispagestyle{empty}
\pagestyle{empty}

\begin{abstract}
We present CineMPC, a complete cinematographic system that autonomously controls a drone to film multiple targets recording user-specified aesthetic objectives. 
Existing solutions in autonomous cinematography control only the camera extrinsics, namely its position, and orientation. In contrast, CineMPC is the first solution that includes the camera intrinsic parameters in the control loop, which are essential tools for controlling cinematographic effects like focus, depth-of-field, and zoom.
The system estimates the relative poses between the targets and the camera from an RGB-D image and optimizes a trajectory for the extrinsic and intrinsic camera parameters to film the artistic and technical requirements specified by the user. The drone and the camera are controlled in a nonlinear Model Predicted Control (MPC) loop by re-optimizing the trajectory at each time step in response to current conditions in the scene. 
The perception system of CineMPC can track the targets' position and orientation despite the camera effects.
Experiments in a photo-realistic simulation and with a real platform demonstrate the capabilities of the system to achieve a full array of cinematographic effects that are not possible without the control of the intrinsics of the camera.  Code for CineMPC is implemented following a modular architecture in ROS and released to the community\footnote{\url{https://github.com/ppueyor/CineMPC_ros}}.

\end{abstract}

\textbf{Keywords - }Aerial Robotics Applications, Autonomous Drone Cinematography, Camera Intrinsics, MPC.
\section{Introduction}
\label{sec_intro}
\input{01_Intro}

\section{Related work}
\label{sec_related}
\input{02_Related_work}


\section{System Overview}
\label{sec_system_overview}
\input{03_System_Overview.tex}

 \section{Cinematographic agents}
\label{sec_agents}
\input{04_Agents}

\section{Control Module}
\label{sec_control_problem}
\input{05_Control_Problem}

\section{Perception Module}
\label{sec_perception}
\input{06_Perception}

\section{Implementation}
\label{sec_implementation}
\input{07_Implementation}

\section{Experiments}
\label{sec_experiments}
\input{08_Experiments}

\section{Future work and limitations}
\label{sec_future_work}

\input{09_Future}

\section{Conclusions}
\label{sec_conclusions}
\input{10_Conclusions}

\bibliographystyle{IEEEtran}
\bibliography{references}

\begin{IEEEbiography}[{\includegraphics[width=1in,height=1.25in,clip,keepaspectratio]{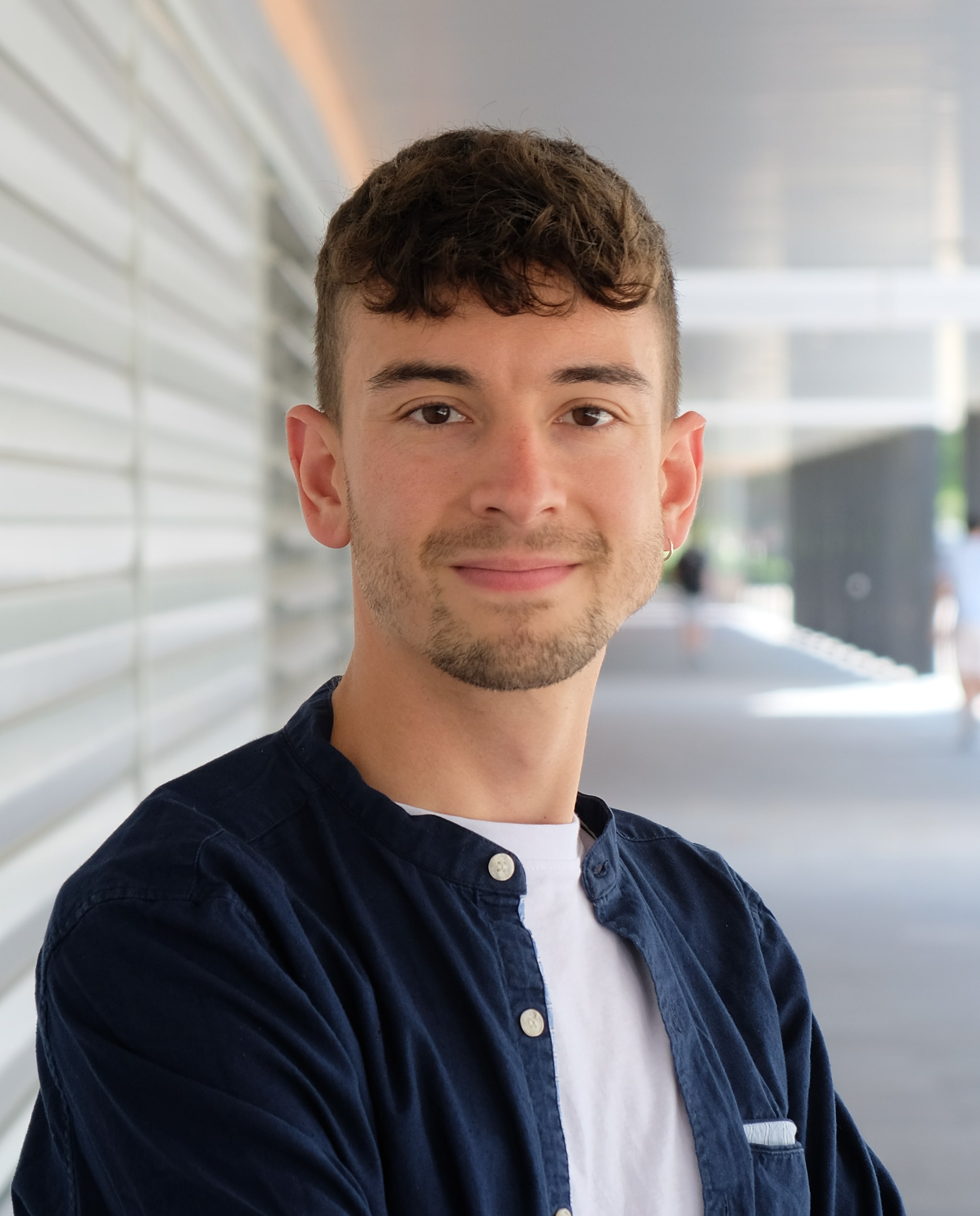}}]{Pablo Pueyo}
(Student Member,
IEEE) received the B.Eng. degree in computer science in 2015 from the Universidad de Zaragoza,
Spain. He attained an M.Eng. in Mobile Engineering in 2018,
from the Universidad Pontificia de Salamanca. He is pursuing a Ph.D. degree in robotics and systems engineering with the Departamento de Informática e
Ingenieria de Sistemas, Universidad de Zaragoza, funded by a DGA
grant. His research interests include aerial cinematography, robotic simulators, MPC, perception, multi-robot systems, and non-linear control.
\end{IEEEbiography}

\begin{IEEEbiography}[{\includegraphics[width=1in,height=1.25in,clip,keepaspectratio]{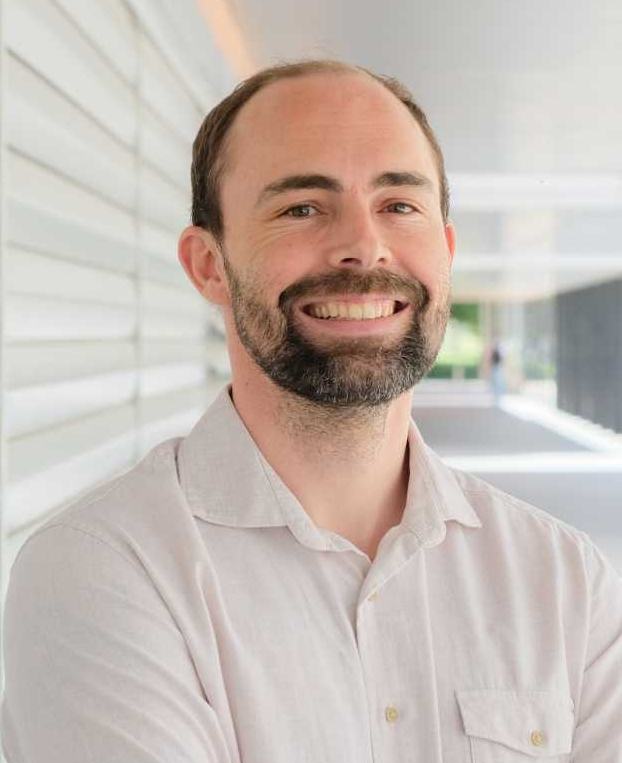}}]{Juan Dendarieta} earned a B.Eng. degree in Industrial Engineering and is currently pursuing a Master's degree in Robotics, Graphics, and Computer Vision at the Universidad de Zaragoza, Spain. He is actively engaged within the Robotics, Computer Vision, and Artificial Intelligence Group at the Universidad de Zaragoza, working on projects involving drones. Juan actively contributes to the progressive advancements in autonomous drone applications, particularly in the realms of navigation and cinematography.
\end{IEEEbiography}

\begin{IEEEbiography}[{\includegraphics[width=1in,height=1.25in,clip,keepaspectratio]{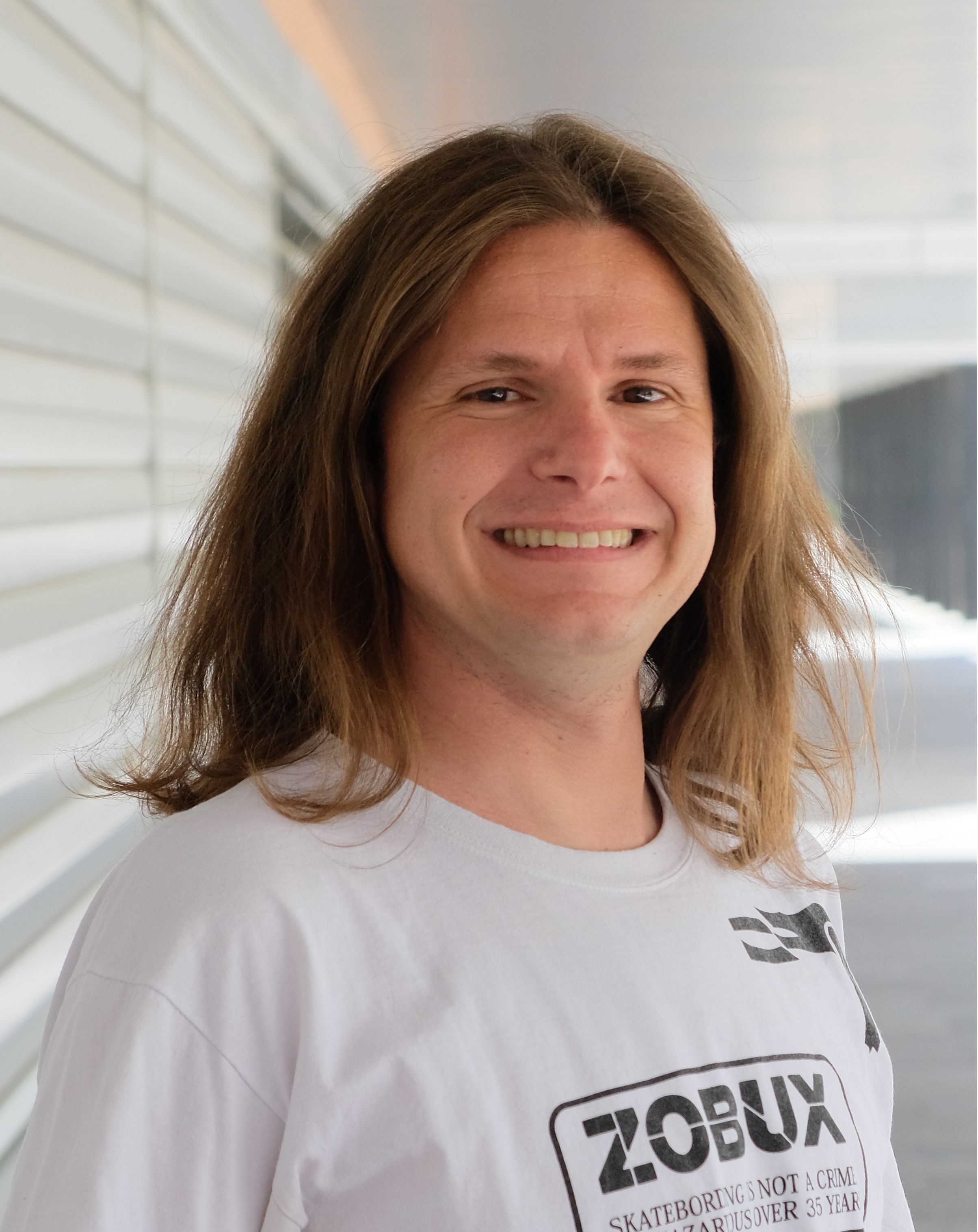}}]{Eduardo Montijano}
(Member, IEEE) received the
M.Sc. degree in computer science and a Ph.D. degree in robotics and systems engineering from the Universidad de Zaragoza, Spain, in 2008
and 2012, respectively.
He is currently Associate Professor with the Departamento de Informática e Ingeniería de Sistemas, Universidad de Zaragoza. He was a faculty member with Centro Universitario de la Defensa, Zaragoza, between 2012 and 2016. His main research interests include distributed algorithms and automatic control in perception problems. 
\end{IEEEbiography}

\begin{IEEEbiography}[{\includegraphics[width=1in,height=1.25in,clip,keepaspectratio]{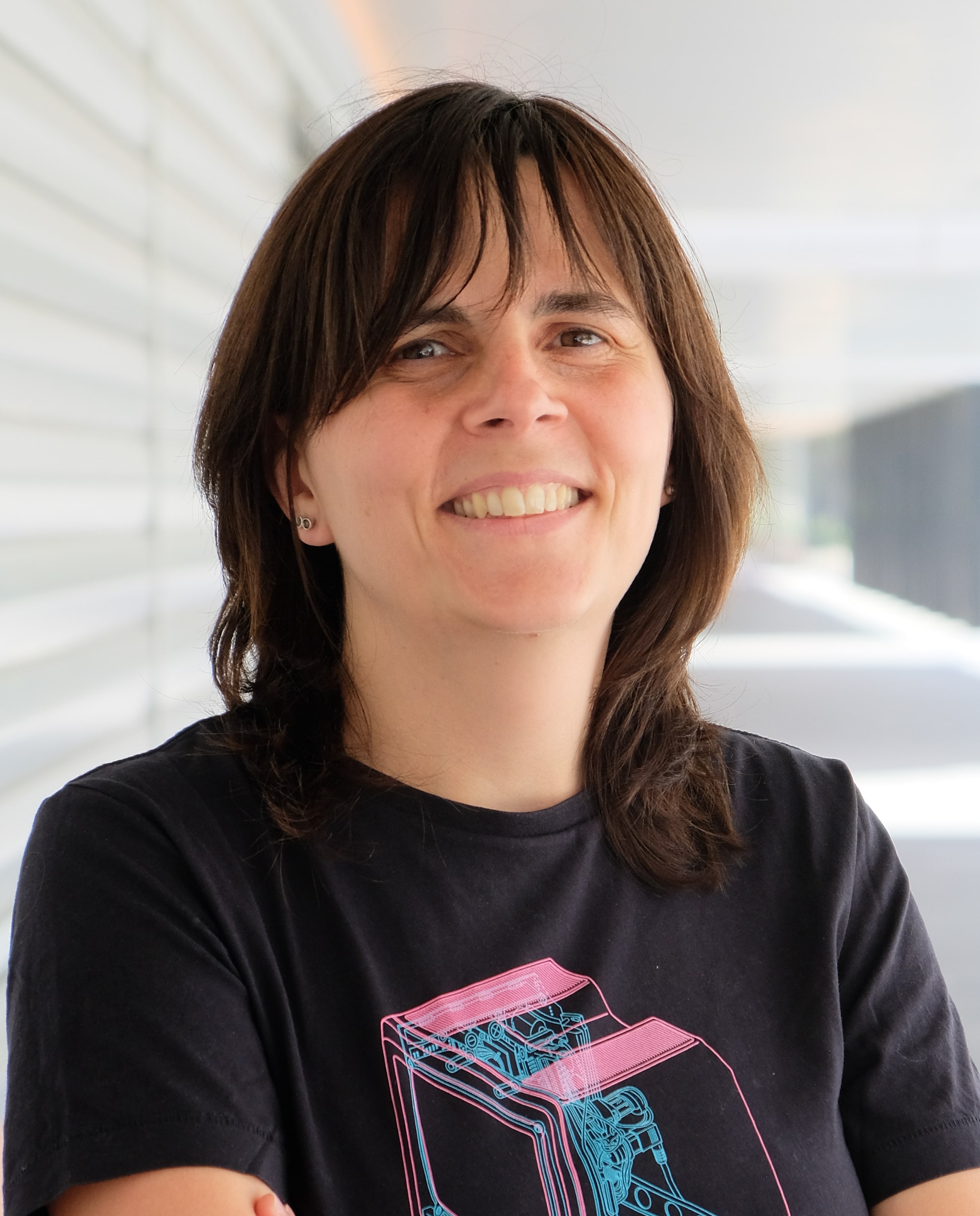}}]{Ana Cristina Murillo}
(Member, IEEE) received the Ph.D. degree in computer science from the University of Zaragoza, Spain, in 2008. 
She is currently researcher and Associate Professor at the Computer Science department at the University of Zaragoza. Her current
research interests are in the area of computer vision and machine learning, with particular interest in scene-understanding problems for robotic and medical applications.
\end{IEEEbiography}

\begin{IEEEbiography}[{\includegraphics[width=1in,height=1.25in,clip,keepaspectratio]{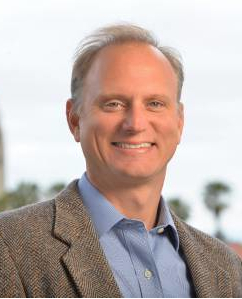}}]{Mac Schwager}
(M’09) received the B.S. degree
in mechanical engineering from Stanford University, Stanford, USA, in 2000, and the M.S.
and Ph.D. degrees in mechanical engineering from Massachusetts Institute of Technology (MIT), Cambridge, in 2005 and 2009, respectively. He is an Associate Professor with the Aeronautics and Astronautics Department, Stanford University. 
His research interests include distributed algorithms for control, perception, and learning in groups of robots. He received the National Science Foundation CAREER Award in 2014.
\end{IEEEbiography}

\end{document}

%% file: 01_Intro.tex
\begin{table*}[bh!]
\begin{center}
\caption{\textbf{Related work.} Comparison of Existing Cinematographic Platforms' Main Properties} 
\scriptsize{
\begin{tabular}{| c | c | c | c | c | c | c | c | c | c | c | c | c | c |}
\hline \multirow{1.5}{*}{\bf{Existing platforms}}
   & \head{1.1cm}{Control Extrinsics}  
   &  \head{1.2cm}{Real Perception} 
   & \head{1.1cm}{Dynamic targets} 
   & \head{1.1cm}{Multitarget} 
   & \head{1.0cm}{Image comp.}   
   & \head{1.1cm}{Obstacle avoidance} 
   & \head{1.1cm}{Occlusion avoidance} 
   & \head{1.2cm}{Public ROS Code} 
   & \head{1.0cm}{Control DoF} 
   & \head{1.0cm}{Control Intrinsics} \vline \\ \hline
  
\addstackgap[4pt]{\bf{CineMPC}} 
& \cellcolor{green!15}\faCheckSquare 
& \cellcolor{green!15}\faCheckSquare
& \cellcolor{green!15}\faCheckSquare
& \cellcolor{green!15}\faCheckSquare
& \cellcolor{green!15}\faCheckSquare 
& \cellcolor{green!15}\faCheckSquare 
& \cellcolor{green!15}\faCheckSquare 
& \cellcolor{green!15}\faCheckSquare 
& \cellcolor{green!15}\faCheckSquare
& \cellcolor{green!15}\faCheckSquare \\ \hline  

\addstackgap[4pt]{Alcantara et al.(2021) [2]} 
& \cellcolor{green!15}\faCheckSquare
& \cellcolor{red!25}\faClose 
& \cellcolor{green!15}\faCheckSquare
& \cellcolor{red!25}\faClose
& \cellcolor{red!25}\faClose 
& \cellcolor{green!15}\faCheckSquare  
& \cellcolor{green!15}\faCheckSquare 
& \cellcolor{green!15}\faCheckSquare 
& \cellcolor{red!25}\faClose
& \cellcolor{red!25}\faClose \\ \hline 

\addstackgap[4pt]{Huang et al.(2018) [33]}  
& \cellcolor{green!15}\faCheckSquare
& \cellcolor{green!15}\faCheckSquare
& \cellcolor{green!15}\faCheckSquare
& \cellcolor{red!25}\faClose
& \cellcolor{olive!25}\faEllipsisH
& \cellcolor{olive!25}\faEllipsisH 
& \cellcolor{red!25}\faClose  
& \cellcolor{red!25}\faClose 
& \cellcolor{red!25}\faClose 
& \cellcolor{red!25}\faClose\\ \hline 

\addstackgap[4pt]{Bonatti et al.(2020) [1]} 
& \cellcolor{green!15}\faCheckSquare
& \cellcolor{green!15}\faCheckSquare
& \cellcolor{green!15}\faCheckSquare 
& \cellcolor{red!25}\faClose
& \cellcolor{olive!25}\faEllipsisH 
& \cellcolor{green!15}\faCheckSquare 
& \cellcolor{green!15}\faCheckSquare 
& \cellcolor{red!25}\faClose 
& \cellcolor{red!25}\faClose
& \cellcolor{red!25}\faClose \\ \hline 

\addstackgap[4pt]{Bucket et al.(2021) [34]} 
& \cellcolor{green!15}\faCheckSquare 
& \cellcolor{green!15}\faCheckSquare
& \cellcolor{green!15}\faCheckSquare
& \cellcolor{red!25}\faClose
& \cellcolor{red!25}\faClose 
& \cellcolor{green!15}\faCheckSquare 
& \cellcolor{green!15}\faCheckSquare 
& \cellcolor{red!25}\faClose
& \cellcolor{red!25}\faClose
& \cellcolor{red!25}\faClose \\ \hline 

\addstackgap[4pt]{Joubert et al.(2016) [29]} 
& \cellcolor{green!15}\faCheckSquare  
& \cellcolor{red!25}\faClose
& \cellcolor{red!25}\faClose
& \cellcolor{green!15}\faCheckSquare
 & \cellcolor{green!15}\faCheckSquare  
 & \cellcolor{red!25}\faClose 
 & \cellcolor{red!25}\faClose 
 & \cellcolor{red!25}\faClose
& \cellcolor{red!25}\faClose 
 & \cellcolor{red!25}\faClose \\ \hline 

\addstackgap[4pt]{Nagëli et al.(2017) [31]} 
& \cellcolor{green!15}\faCheckSquare 
& \cellcolor{red!25}\faClose
& \cellcolor{red!25}\faClose
& \cellcolor{green!15}\faCheckSquare
& \cellcolor{green!15}\faCheckSquare  
& \cellcolor{green!15}\faCheckSquare  
& \cellcolor{green!15}\faCheckSquare
& \cellcolor{red!25}\faClose 
& \cellcolor{red!25}\faClose
& \cellcolor{red!25}\faClose\\ \hline 
 \end{tabular}
 }
\label{table:competitors}
\end{center}
\end{table*}

Remotely piloted multi-rotor aircraft have already been widely adopted as mobile camera platforms for videography in television, film, commercial, and hobby applications. These platforms have also expanded the locations from which aerial footage can be obtained to include geographical regions previously inaccessible.  However, a skilled human pilot and human camera operator are still essential to operate these systems.  

\begin{figure}[!t]
\centering
\begin{tabular}{c}
    \includegraphics[width=0.99\columnwidth]{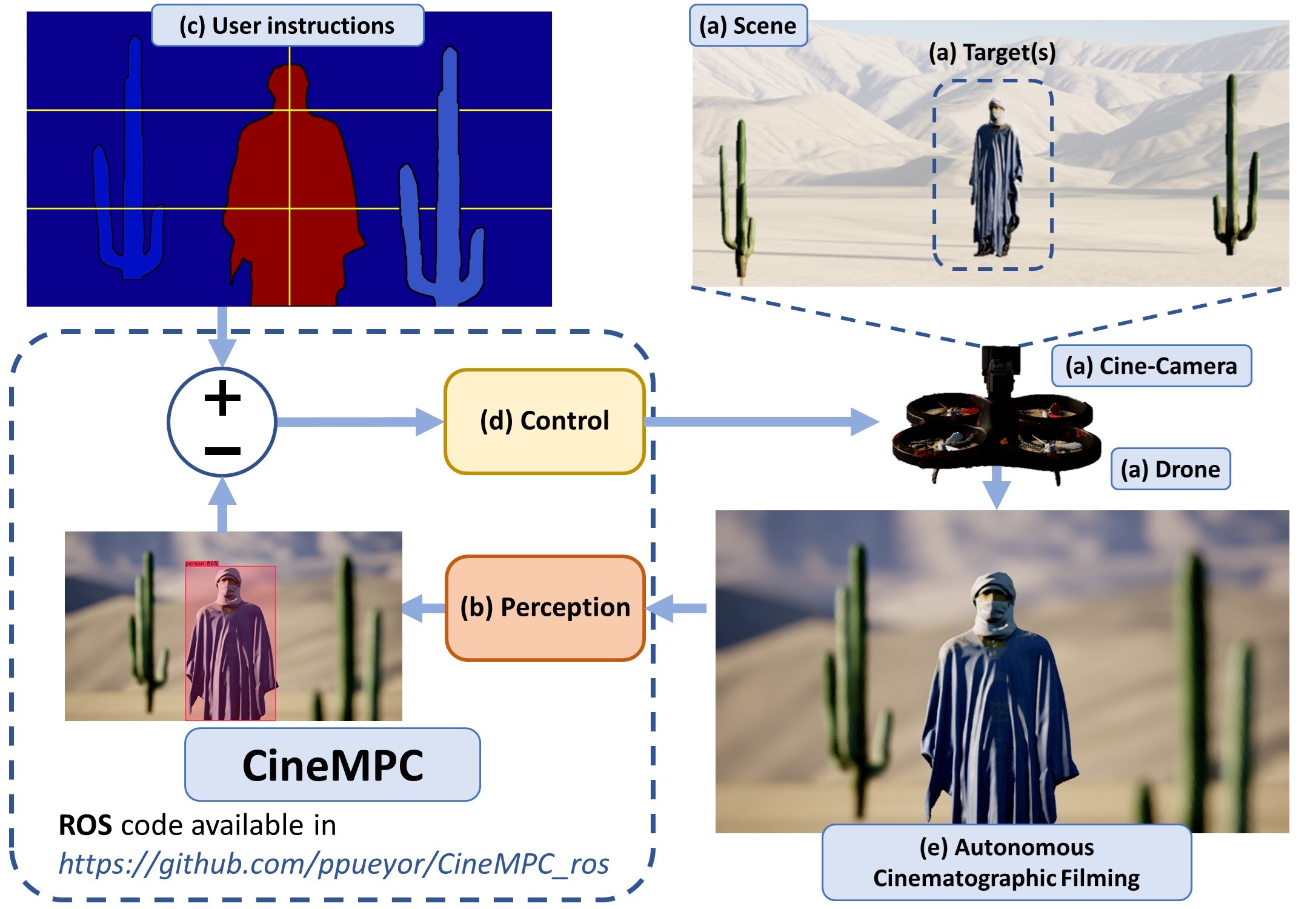}
\end{tabular}
\caption{\footnotesize{\textbf{CineMPC pipeline}. (a) The drone holds a cinematographic camera, capturing footage of targets in a scene. (b) The perception module processes the recorded images to extract the targets' pose and calculates the error in comparison to user instructions.  (c) Visual representation of user instructions, with the focused area highlighted in red, the blurry area in blue, and yellow lines depict the desired image position for the top and lower parts of the target. (d) The calculated error is the input to the control module, which determines the next $N$ steps for both the drone and the camera to minimize the error. (e) This process results in a new image acquired by the drone, restarting the loop for continuous refinement, producing Autonomous Cinematographic Filming.
}}
\label{fig:main}
\end{figure}

In this paper, we propose CineMPC as a step toward making drone aerial videography truly autonomous.  In contrast with other research in autonomous videography \cite{bonatti2020autonomous, alcantara2020optimal}, CineMPC controls both the camera pose as well as the focus, depth-of-field, and zoom---the so-called camera intrinsics---thereby doing the job of both the pilot and the camera operator. We accomplish this through a thin-lens model~\cite{lu2020camera} of the camera optics, which exposes these camera intrinsic properties as control inputs.  We then optimize a sequence of control inputs for both the camera pose and camera intrinsics while constraining the trajectory to be dynamically feasible for the drone.  We close the loop by detecting the poses of the multiple targets in the scene in real-time, and re-optimizing the trajectory in an MPC loop as new images are acquired. CineMPC is able to track and record multiple dynamic targets (such as humans, animals, cars, or other aircraft) while taking footage to optimize artistic and technical objectives specified by the user. This framework is depicted in Fig.~\ref{fig:main}.

The core idea of the control module is to adapt the classic cinematographic concepts~\cite{thompson2009grammar} to mathematical expressions that can be optimized using control techniques. The specifications are optimized thanks to a nonlinear MPC formulation that transforms them into instructions to autonomously control the drone and the camera while recording footage. The drone position and orientation are controlled together with the intrinsic parameters of the camera lens in one unified control problem.

CineMPC's perception module identifies and estimates the pose of targets in images captured by a thin-lens cinematographic monocular camera. In images from this camera model, targets may appear blurry or distorted, posing a challenge to pose estimation. To address this issue, the module employs a neural network to extract target positions from RGB-D images and uses a Kalman Filter and vector algebra to determine target orientation based on movement direction. Existing solutions \cite{lewandowski2019deep,chen2020monocular,nageli2018flycon} for 3D orientation estimation with monocular cameras often require invasive wearables or involve heavy deep learning, making them impractical for real-time aerial cinematography of targets in the wild.

We release a modular implementation of the whole solution in ROS that also incorporates the tools and instructions to test it using CinemAirSim \cite{pueyo2020cinemairsim}, an extension for cinematographic purposes of the robotics simulator AirSim\cite{shah2018airsim}. 
In this environment, we conduct a battery of photorealistic experiments that, along with real-world experiments, demonstrate the potential of our approach.

The main contributions of this work are the following:

\begin{enumerate}
\item \textbf{Optimal Control Problem:} We propose a novel optimal control problem within an MPC framework. This enables autonomous control not only over the extrinsic but also the intrinsic parameters of a drone and cinematographic camera.  This facilitates capturing previously unattained cinematographic effects while handling different constraints,

\item \textbf{Integration with Perception Module:} 
The control solution is integrated with a perception module capable of tracking 3D poses for multiple moving targets from RGB-D images. This capability remains unaffected by image distortions resulting from the modification of intrinsic camera parameters,

\item \textbf{ROS Implementation:} A mature ROS implementation is released with a modular software architecture, facilitating the adaptation of CineMPC to new drones and diverse aerial videography applications,

\item \textbf{Practical Implementation Aspects:} We delve into practical considerations associated with implementing a fully autonomous cinematographic platform in a real setup. We present an extensive array of experiments, in scenarios with both a real drone and camera, along with challenging filming sequences conducted in simulation.

\end{enumerate}

This work is \emph{an evolved version of \cite{pueyo2021cinempc}}, extending the core optimal control problem addressed in the conference paper. The control module now handles significant cinematography and robotics constraints, including collisions and occlusions. Additionally, we introduce an extra cost term ($J_f$) to achieve a broader range of cinematographic effects and a low-level controller to ensure smooth trajectory execution. The remaining contributions are primarily introduced in this extended work.

%% file: 02_Related_work.tex
The design of new user-friendly interfaces to direct drones with cinematographic purposes is essential for their use in cinematography. There are several efforts on this topic, for example,  \cite{gebhardt2016airways} shows a new way to introduce a simplified trajectory, which is extended in \cite{gebhardt2018wyfiwyg} allowing the introduction of aesthetic requirements. In \cite{joubert2015interactive}, a complete tool helps expert and novice cinematographers to achieve a visually pleasant drone trajectory, based on key-frames and some aesthetic user inputs. Other works, like~\cite{lan2017xpose} and \cite{kang2018flycam}, develop touch interfaces to specify how to record a target.
Although it is not the focus of this work, our implementation includes a user interface to ease the introduction of the control objectives.

In order to autonomously record aesthetically attractive footage while satisfying some cinematographic constraints
some works present mathematical expressions to measure how good or bad the aesthetics of an image are \cite{perona2021application, mai2011rule, fang2020perceptual}. These formulas are used to move a regular camera  to a position that satisfies instructions from cinematographers in an autonomous way~\cite{he1996virtual, li2005interactive,lino2011director}
or to find optimal views considering a static scene, enabling canonical static shots, like the rule of thirds~\cite{xiong2017automatic}. In contrast to CineMPC, these solutions only control the extrinsics of the camera, limiting the number of cinematographic options.

Other approaches focus on making the trajectory of the drone smoother while recording. Given a set of way-points, different MPC formulations are used to control the drone to avoid unstable trajectories while passing through the established points \cite{rousseau2018quadcopter, gebhardt2018optimizing}. In \cite{galvane2018directing}, the director introduces the desired shot composition or a set of viewpoints, and a team of multiple drones records targets following a smooth trajectory, ensuring collisions and occlusion avoidance. Recent works attempt to imitate the trajectories run by a professional cinematographer.

In \cite{ashtari2020capturing} the drone imitates a walking camera operator that moves following one of the predefined patterns of movement to record a first-person view of a person, and \cite{huang2019learning} and \cite{huang2021one} use imitation learning to reproduce the cinematographers' operations. These drone platforms are intended to take a single photo of a static scene or record footage either reproducing a determined kind of shot or following a predetermined trajectory of viewpoints, for a delimited amount of time.
In opposition, CineMPC can track static and moving targets 
during an indeterminate time. Moreover, the control of the drone is not limited to the trajectory of the drone itself, but also the trajectory of the intrinsic camera parameters.

Other works can deal with multiple targets. For instance,  \cite{joubert2016towards} guides a drone to record multiple targets following the rules of a predefined set of shots. In \cite{alcantara2020optimal}, a multidrone platform lets the user choose among a list of canonical drone shots according to \cite{smith2016photographer}. MPC is used in \cite{nageli2017real} to film scenes while tracking and recording multiple targets, according to some cinematographic standards, i.e., the position of targets on the image. The multi-drone platform presented in \cite{nageli2017realmulti} uses MPC to record different targets, optimizing a trajectory of predefined viewpoints, while avoiding occlusions and collisions. 
Although these approaches represent substantial advances in cinematographic platforms, they do not use real perception to extract the pose of the targets. 

The solutions presented in \cite{huang2018act} and \cite{bonatti2020autonomous} use real perception to track and record people doing different kinds of activities while following a visually pleasant trajectory.
The authors of \cite{bucker2021you} use some of the principles of \cite{bonatti2020autonomous} to present a multi-drone approach, enabling multi-view of the target and avoiding collisions, occlusions, and view-point similarities between drones.
All these solutions focus on getting the best shots of human targets by only optimizing the extrinsic parameters, e.g., the position and orientation of the drones. 
Compared to them, our approach is the first that introduces an essential factor for high-quality photography into the control problem: the intrinsic parameters of the camera lens. 
Besides, we also use real perception to track different types of targets.

Table \ref{table:competitors} shows the contributions of the most relevant existing works compared to CineMPC. 
The titles of the headers of the columns are reduced for the sake of space. The complete titles are, respectively: Control of Extrinsics, Real Perception, Dynamic Targets, Multitarget, Control Image Composition (position of elements in the image), Obstacle avoidance, Occlusion avoidance, Public ROS code, Control of the Depth of Field (DoF), and Control of Intrinsic Parameters.

%% file: 03_System_Overview.tex
\begin{figure*}[tb!]
\centering
         \includegraphics[width=0.97\linewidth]{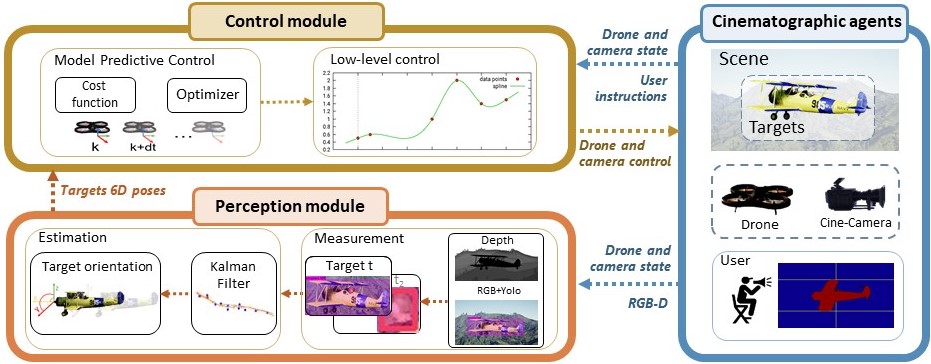}
	\caption{\footnotesize \textbf{CineMPC System Overview}. A schematic summary of the platform, its modules, and their interactions. The cinematographic agents comprise the scene (containing the target(s)), the drone, the cine-camera, and the user providing instructions. The perception module utilizes camera images to extract the pose of targets, which are then fed to the control module. This module calculates the trajectory for the next $N$ steps for both the camera and drone, optimizing the cost function within an MPC framework. This trajectory is transmitted through a low-level controller, ensuring smooth recording.
 }
	\label{fig:system_overview}
\end{figure*}

The complete CineMPC system is represented in Fig.~\ref{fig:system_overview}. This figure shows a high-level schematic of the modules involved in the system and the communication between them. 

\paragraph{Cinematographic agents} These components are found in any real cinematographic setup involving drones; the \textit{user}, that gives instructions, (e.g., a movie director, director of photography), the \textit{drone}, responsible for holding and maneuvering the cinematographic \textit{camera} through the \textit{scene} or environment, where the \textit{targets} are found. Section \ref{sec_agents} provides detailed explanations of these agents.

\paragraph{Control module} This module is in charge of performing the computations that give autonomy to the drone. It transforms the users' requirements in actions to apply to the drone and the camera to achieve the cinematographic objectives. The \textit{MPC solver} solves the control problem, using the \textit{optimizer} to resolve a \textit{cost function}, and gives actions that the \textit{low-level controller} transforms into commands sent to the drone and the camera to take actions accordingly. Section \ref{sec_control_problem} describes each component in detail.
\label{contol_module}

\paragraph{Perception module} This module processes \textit{RGB} and \textit{Depth} images from the RGB-D camera, estimating the \textit{6D pose of the targets} and providing the poses to the Control Module. Target position is determined through \textit{Yolo} detector~\cite{redmon2018yolov3} detection in the image and the depth map. A \textit{Kalman filter} predicts the next $N$ position and velocity values, used for calculating the target's orientation. Section \ref{sec_perception} details the 
 steps for this module.
\label{perception_module}

%% file: 04_Agents.tex
This section describes the cinematographic agents present in a real-world setup with drones, namely the user, the drone, the camera, and the scene. 
\subsection{Drone and Gimbal - Extrinsic parameters}
\label{sec:drone}
The drone is the flying vehicle that holds and moves the recording camera around the 3D space.
In this paper, we consider a simplified model of this vehicle and leave the accurate control of its high-order complex dynamics to the low-level control module (Sec.~\ref{Sec:lowlevel}). This way, CineMPC remains flexible to be used with different platforms, as long as the manufacturers provide suitable low-level controllers.

We define the position and velocity of the drone at discrete time instant $k$ by $\mathbf{p}_{d,k}$ and $\mathbf{v}_{d,k} \in \mathbb{R}^{3}$ respectively.
We decouple the orientation of the camera from that of the drone, assuming the presence of a gimbal.
While a gimballed camera is somewhat unusual in hobby drones, it is standard in high-quality cinematography drones~\cite{kang2018flycam}.
Most gimbals also implement image stabilization strategies, as a high velocity on the movement of the drone produces aggressive motions that can lead to shaky recordings. 
Nevertheless, for the sake of simplicity, we denote this orientation by $\mathbf{R}_{d,k} \in SO(3)$.
Therefore, the extrinsic parameters of the system are 
\begin{equation}
\mathbf{x}_{d,k} = (\mathbf{p}_{d,k}, \mathbf{v}_{d,k}, \mathbf{R}_{d,k}).
\end{equation}

The actuators in the simplified model are the drone acceleration, and the angular velocity of the gimbal, and are represented by  $\mathbf{a}_{d,k} \in \mathbb{R}^{3}$ and $\mathbf{\Omega}_{d,k}  \in \mathbb{R}^{3}$ and are grouped into the drone actuators vector $\mathbf{u}_{d,k}$, 
\begin{equation}
\mathbf{u}_{d,k}=(\mathbf{a}_{d,k}, \mathbf{\Omega}_{d,k}).
\end{equation}
According to this, for the optimal control problem, we consider double integrator dynamics for the position-velocity pair,

\begin{equation}
\mathbf{p}_{d,k+1} = \mathbf{p}_{d,k}+ \Delta_T \mathbf{v}_{d,k},\quad
\mathbf{v}_{d,k+1} = \mathbf{v}_{d,k} + \Delta_T\mathbf{a}_{d,k},
\label{eq:dynamics_drone1}
\end{equation}
where $\Delta_T$ is the sampling time of the discrete model, and the rotation evolves according to

\begin{equation}
\mathbf{R}_{d,k+1} = \mathbf{R}_{d,k}  \exp\left(\Delta_T\mathbf{\Omega}_{d,k}^{\wedge}\right),
\label{eq:dynamics_drone2}
\end{equation}
where $\exp(\cdot)$ is the exponential map, used to compute the rotation matrix obtained by rotation at constant angular speed $\mathbf{\Omega}_{d,k}$ for $\Delta_T$ seconds.

\subsection{Cinematographic Camera - Intrinsic parameters}
The cinematographic camera, a key component of CineMPC, captures the scene. Traditional cameras use the pin-hole camera model, considering only projection and geometric parameters. In this model, all the image rays pass through an aperture (hole) at the center of the sensor, showing the whole image in focus. However, the aperture of any real camera has a finite diameter, it is not a pin-hole. A higher fidelity model of a camera is given by the thin-lens camera model, where a lens replaces the sensor's hole. This substitution allows the control of the image focus, the depth of field, and other artistic features by adjusting the camera's intrinsics \cite{baba2002thin}. 

The camera's intrinsics that we model in CineMPC are the focus distance, the focal length, and the aperture. 

\begin{figure}[!h]
\centering
\begin{tabular}{ccc}
    \includegraphics[width=0.48\columnwidth]{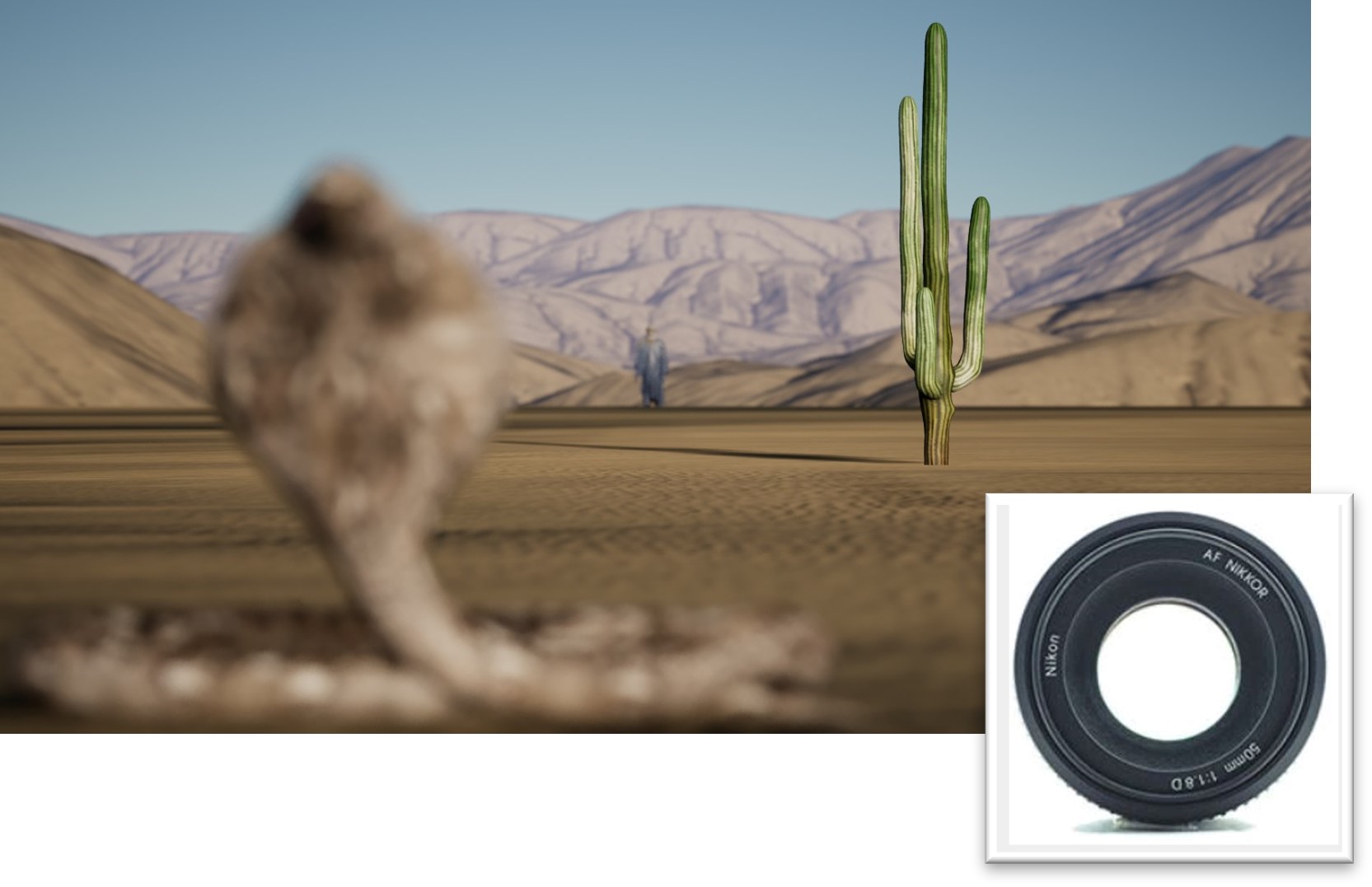}
    & 
    \includegraphics[width=0.48\columnwidth]{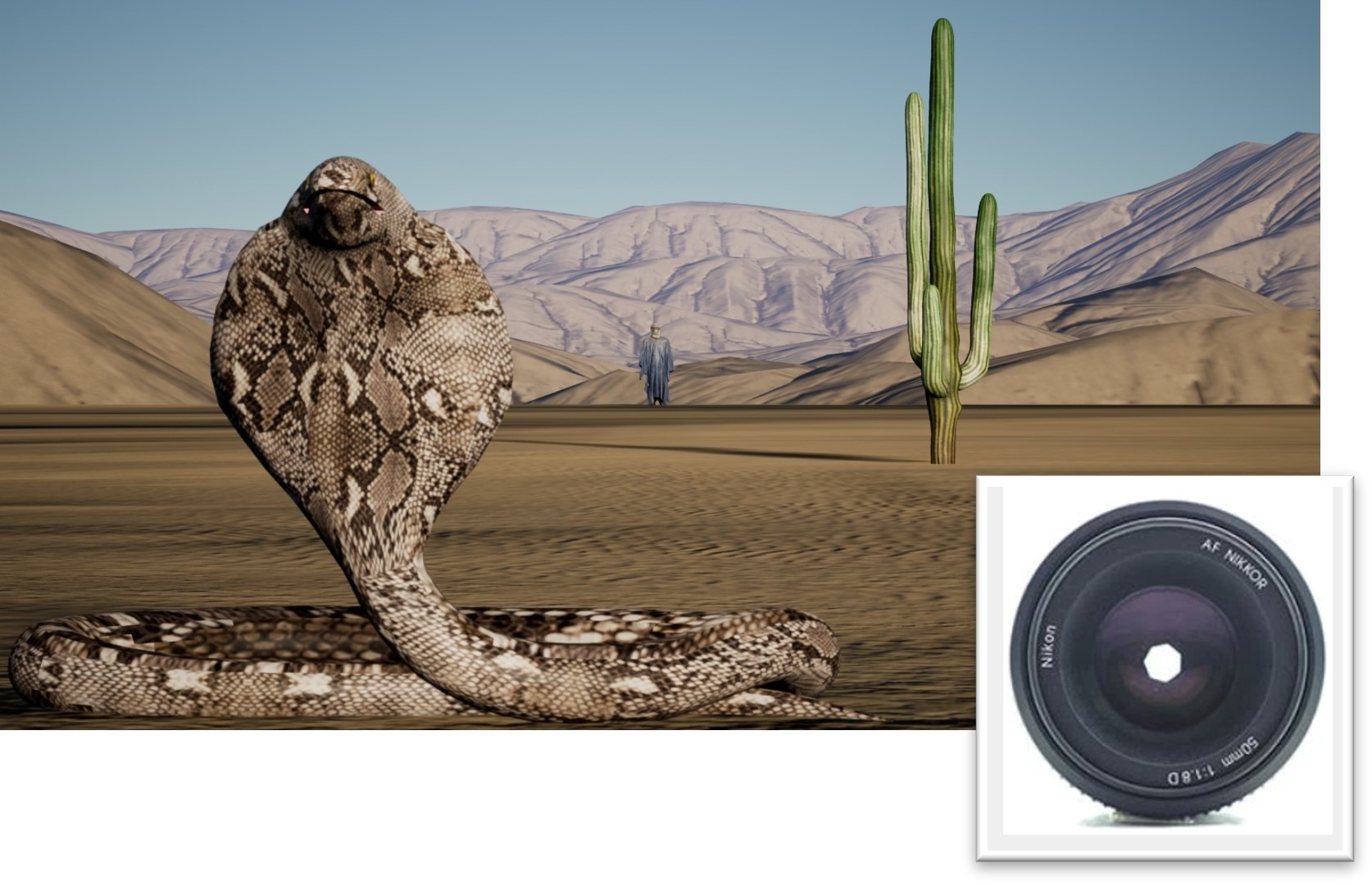} 
     \\ \footnotesize $A$ = f/1.2 & \footnotesize $A$ = f/22 \\
     
    \includegraphics[width=0.48\columnwidth]{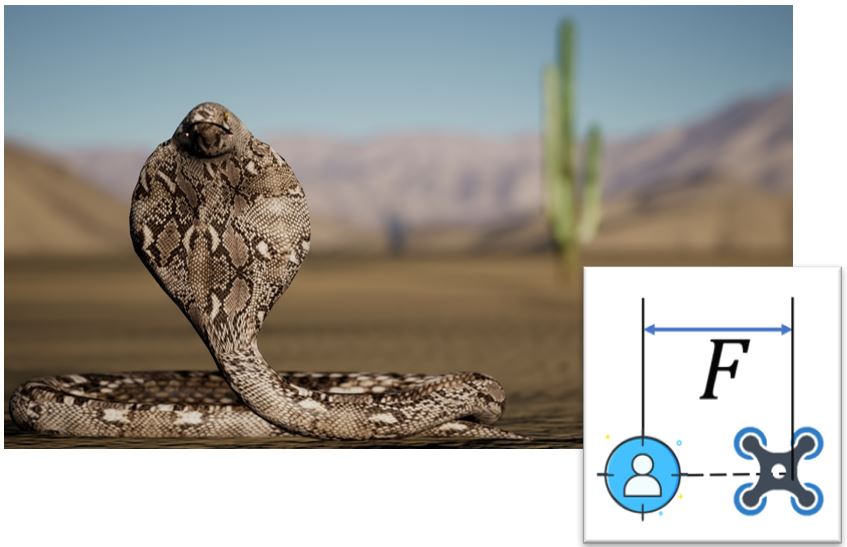}
    & 
    \includegraphics[width=0.48\columnwidth]{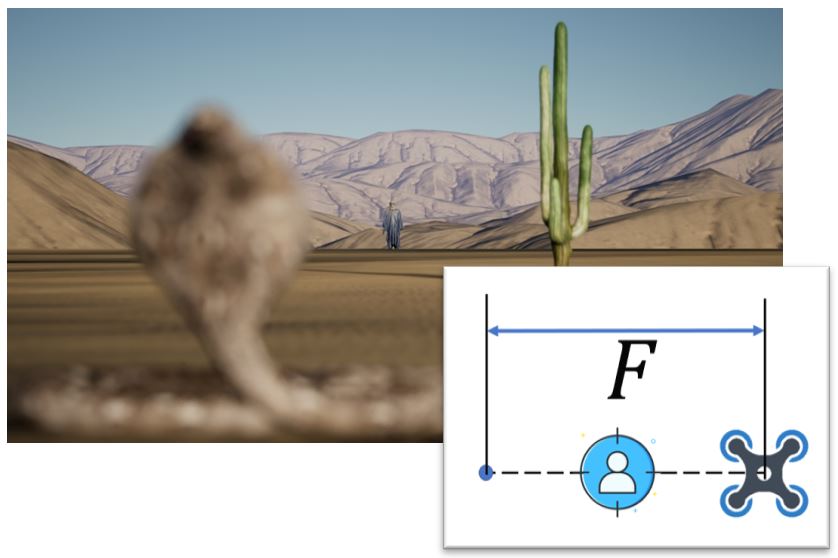} 
     \\ \footnotesize $F$ = 3 m & \footnotesize $F$ = 90 m \\
     \includegraphics[width=0.48\columnwidth]{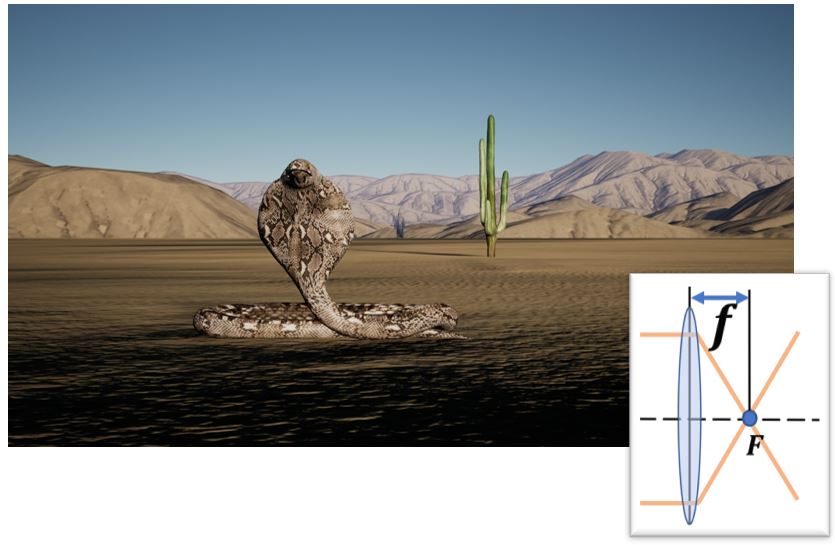}
    & 
    \includegraphics[width=0.48\columnwidth]{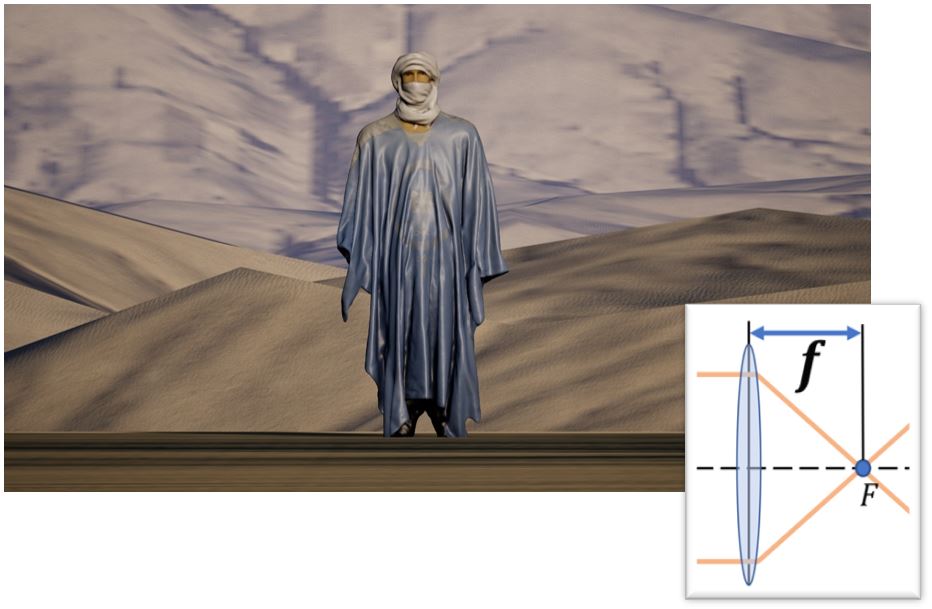} 
     \\ \footnotesize $f$ = 30 mm & \footnotesize $f$ = 400 mm \\
\end{tabular}
\caption{\footnotesize{\textbf{Effect of intrinsics in the final image}. 
 The first row compares two aperture ($A$) values, affecting the portion of the scene shown in focus (depth of field). The left side, with a low f-stop, has a wider aperture and shallow depth of field. The right side, with a high f-stop, has a narrow aperture and a larger depth of field. The second row contrasts two focus distance ($F$ - distance from the camera to the center of the depth of field) values, focusing on a closer distance (left) and a further distance (right). The third row compares different focal length ($f$) values, affecting the zoom, field of view, and depth of field. The left side has a small focal length, providing a wide-angle view. The right side has a large focal length, resulting in a highly zoomed image. The camera maintains the same pose (same extrinsics) across all images.} 
}
\label{fig:agents_aperture_dof}
\end{figure}

The \textbf{focus distance}, $F_k$, represents the distance from the camera where the elements appear in perfect focus. The definition of the \textbf{focal length}, $f_k$ is different in the pin-hole model and the thin-lens model. In pin-hole, it represents the distance in millimeters between the aperture and the sensor. In the thin-lens camera model, it is the distance from the optical center of the lens and the point of focus, where the parallel rays from the image intersect. The focal length affects different artistic effects, such as the field of view 
and the depth of field (part of the scene that is in focus). The \textbf{lens aperture}, $A_k$, controls light intake by adjusting the size of the opening through which image light passes to the camera sensor. 
Expressed through the f-number (or f-stop), it influences image brightness, exposure, Bokeh effect, and depth of field.
Fig.~\ref{fig:agents_aperture_dof} shows a graphical explanation of the effect of the intrinsics in the final image.

The vector  $\mathbf{x}_{c,k}$ represents the state of the intrinsics, 
\begin{equation}
\mathbf{x}_{c,k} = (f_{k}, F_{k}, A_{k}).
\end{equation}
The relationship of these parameters with the extrinsics to determine the images acquired by the camera is detailed in Section~\ref{sec:MPC}. In this section, we describe how we model their dynamic behavior.
The intrinsic parameters can be set to any value within the physical camera range. 
We prevent large variations of the intrinsics in a short time, which can lead to aggressive image changes, not artistically pleasant in cinematography, by controlling their velocities instead of acting on their values directly, 
\begin{equation}
\mathbf{u}_{c,k}=(v_{f,k}, v_{F,k}, v_{A,k}),
\end{equation}
where $v_{f,k} \in \mathbb{R}$ denotes the velocity of the focal length, expressed in $mm/s;$ $v_{F,k} \in \mathbb{R}$ denotes the velocity of the focus distance, in $m/s;$ $v_{A,k} \in \mathbb{R}$ is the velocity of the aperture, in $f\_stop/s$ all of them measured in the discrete-time step $k$.

This way, the intrinsic parameters evolve according to a single integrator model,

\begin{equation}
\mathbf{x}_{c,k+1} = \mathbf{x}_{c,k} + \Delta_T \mathbf{u}_{c,k}.
\label{eq:dynamics_camera1}
\end{equation}

\vspace{-15pt}
\subsection{Scene}
The scene is the part of the environment captured by the camera, where many complex elements participate, e.g., foreground, background, people, and objects.

We model the scene as a set of $n$ targets, represented by points of interest to be recorded. 
Similar to the drone, the state, $\mathbf{x}_{t,k}$, of each target is described by its position, $\mathbf{p}_{t,k} \in \mathbb{R}^3,$ velocity $\mathbf{v}_{t,k} \in \mathbb{R}^{3}$ and rotation in the world, $\mathbf{R}_{t,k} \in \mathbb{R}^3$,  
\begin{equation}
\label{Eq:targetState}
\mathbf{x}_{t,k} = \left(\mathbf{p}_{t,k},\mathbf{v}_{t,k}, \mathbf{R}_{t,k} \right).
\end{equation}
Besides, we include additional information about the targets to describe their nature,
$t_{nature}$, e.g., person, plane, etc., their estimated sizes in meters, i.e., width, $t_{w}$ and height, $t_{h}$, and a preliminary orientation,  $t_{R} \in SO(3)$,
\begin{equation}
\label{Eq:TargetInfo}
    \bm{\mu}_{t} = (t_{nature}, t_{h}, t_{w}, t_{R}).
\end{equation} 
This information is used in the control module to handle scene constraints and in the perception module to help in the estimation of the target state (Eq.~\ref{Eq:targetState}). We provide more details in the next sections.

\subsection{User}
\label{sec:user}
The user gives the artistic and technical instructions to record the footage. Examples of individuals in this role include a movie director, a photographer, or an amateur user.
In CineMPC, the user specifies the recording instructions and constraints, which are grouped into the sets $\bm{\mu}$ and $\bm{\mathcal{C}}$, respectively. Vector  $\bm{\mu}$ contains the recording instructions, namely the instructions on the nature of the targets, $\bm{\mu}_{t}$, the composition, $\bm{\mu}_{im}$, and  depth of field of the image, $\bm{\mu}_{DoF}$, the desired values of the intrinsics, $\bm{\mu}_{f}$, and the relative pose where the camera should be placed to record the targets, $\bm{\mu}_{p}$,
\begin{equation}
    \bm{\mu} = (\bm{\mu}_{t}, \bm{\mu}_{DoF}, \bm{\mu}_{im}, \bm{\mu}_{f}, \bm{\mu}_{p}).
\end{equation} 
All these parameters are described in the next sections of the paper.
The content and specification of the set of constraints $\bm{\mathcal{C}}$  are detailed in Sec. \ref{sec_constraints} of the paper.

%% file: 05_Control_Problem.tex
\begin{figure}[h!]
\centering    
\includegraphics[width=\columnwidth]{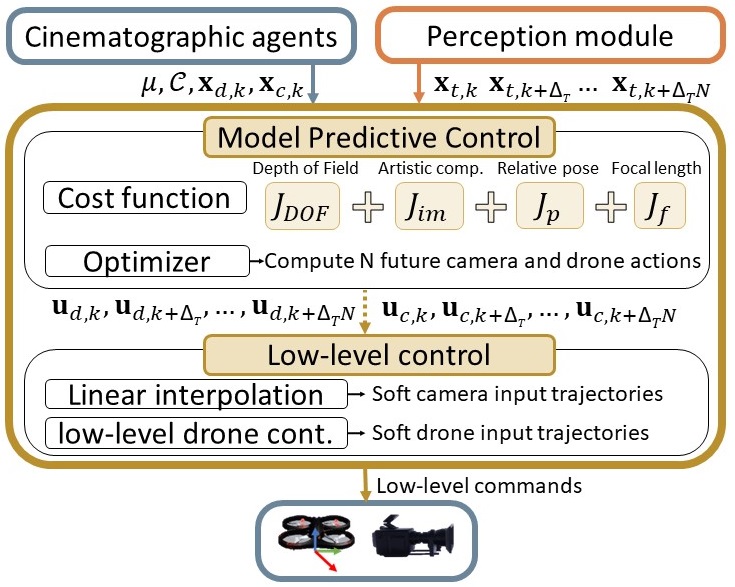} 
\caption{\footnotesize \textbf{Control module diagram.} Components of the control module and their interaction. The module consists of the MPC framework, comprising the cost function, optimizer, and the low-level control submodule. The diagram incorporates inputs from other modules and the module's outputs.
}
\label{fig:control_graph}
\end{figure}

CineMPC solves a non-linear optimization problem inside an MPC framework~\cite{camacho2013model}. Then, a low-level controller transforms the output of the MPC framework into commands to be sent to the drone and the camera. Figure ~\ref{fig:control_graph} shows a graphical explanation of the control module.
\subsection{Model Predictive Control}
\label{sec:MPC}

At a given time $k_0$, CineMPC solves the following problem over a time horizon $N$,
\begin{equation}
\begin{aligned}
\min_{\substack{\mathbf{u}_{d,k_0}..\mathbf{u}_{d,k_0+N}\\ \mathbf{u}_{c,k_0}..\mathbf{u}_{c,k_0+N}}} \quad & \sum_{k=k_0}^{k_{0}+N}{J(\bm\mu,\mathbf{x}_{d,k},\mathbf{x}_{c,k})} \\
\textrm{s.t.} \quad & \eqref{eq:dynamics_drone1}, \eqref{eq:dynamics_drone2}\hbox{ and }\ \eqref{eq:dynamics_camera1}\\
  & g(\bm{\mathcal{C}},\mathbf{u}_{d,k},\mathbf{u}_{c,k},\mathbf{x}_{d,k},\mathbf{x}_{c,k}) \ge 0 \\
\end{aligned},
\label{eq:main_cost}
\end{equation}
where $J(\bm\mu,\mathbf{x}_{d,k},\mathbf{x}_{c,k})$ is the cost function, which considers the user instructions and $g(\bm{\mathcal{C}},\mathbf{u}_{d,k},\mathbf{u}_{c,k},\mathbf{x}_{d,k},\mathbf{x}_{c,k})$ encodes the set of constraints.
Note how the optimization problem computes all intrinsic parameters alongside extrinsic factors while considering scene and recording constraints. In the following, we provide more technical details on these two functions.

\subsubsection{Cost function}
The cost function is composed of four terms,
\begin{equation}
J(\bm\mu,\mathbf{x}_{d,k},\mathbf{x}_{c,k}) \equiv J_k = J_{DoF,k} + J_{im,k} + J_{p,k} + J_{f,k}
\end{equation}
associated with the depth of field, the artistic composition, the relative position between the camera and the targets, and the desired values of the intrinsics, respectively.

\paragraph{Focus of the image - Depth of Field} 

Our solution autonomously controls the camera depth of field, which represents the space of the scene that appears acceptably in focus in the image. 
According to specialized literature in cinematography and optics~\cite{bass2010handbook}, the image depth of field is delimited by two points, the near, $D_{n}$ and 
the far, $D_{f}$ distances. 
The region in the scene between $D_{n}$ and $D_{f}$ is in focus, while the rest appears blurry in the image.

To relate these distances to the camera intrinsics it is convenient to describe first the HyperFocal Distance, $H_{k}$, 

calculated as follows: 
 \begin{equation}
 H_{k} = \frac{f_k^{2}}{A_k c} + f_k,
 \end{equation}
 where $c$ is the circle of confusion, a constant parameter that depends on the model of the camera and expresses the limit of acceptable sharpness.
The near distance, $D_{n,k}$, represents the closest distance to the camera where the focus of the  projected points is acceptable, 
\begin{equation}
D_{n,k} = \frac{F_k (H_{k} - f_k)}{H_{k} + F_k - 2 f_k}.
\end{equation}
Analogously, the far distance, $D_{f,k},$ is the farthest distance to the camera where projected points are acceptably in focus, 
\begin{equation}
D_{f,k} = \frac{F_k (H_{k} - f_k)}{H_{k} - F_k}.
\end{equation}

To determine the desired part of the scene to be in focus, the set of instructions $\bm{\mu}_{DoF}$ 
includes the desired near, $D_{n,k}^{*}$, and far distances, $D_{f,k}^{*}$, expressed in meters from the camera. Additionally, $w_{D_{n}}$ and $w_{D_{f}}$ represent the weights associated with the cost terms of the near and far distances.
The cost term of the depth of field in the time step $k$ penalizes intrinsic values that make actual distances depart from the desired values,
\begin{equation}
J_{DoF,k} = w_{D_{n}}\left(D_{n,k} - D_{n,k}^{*}\right)^2 + w_{D_{f}}\left(D_{f,k} - D_{f,k}^{*}\right)^2.
    \label{eq:dof-cost}
\end{equation}
It is important to note that as $f_k$ approaches low values, $D_f$ tends towards infinity, implying that the image background is in focus. In contrast, $D_n$ is always controllable. Therefore, it is common for $w_{D_{n}}$ to be higher than $w_{D_{f}}$, or alternatively, setting $w_{D_{f}} = 0$ when $f_k$ has a reasonably low value.

\paragraph{Artistic composition - Position of elements in image}
\label{sec_control_artistic}
The objective of this term is to show the targets placed in particular regions of the image. This term makes the elements appear in the final image so that they satisfy some cinematographic composition rules, e.g., the rule of thirds. 
Using the camera projection model, we define a cost term that penalizes deviations from the desired image composition.
Let $\mathbf{K}$ be the calibration matrix of the camera~\cite{szeliski2010computer},
\begin{equation*}
\mathbf{K}=
\begin{bmatrix}
\beta_x f_k && s && c_u
\\
0 && \beta_y f_k && c_v
\\
0 && 0 && 1
\end{bmatrix},      
\end{equation*}
with $c_u$ and $c_v$ the image optical center coordinates and $s$ the skew. 
The focal length affects the projection, thus coupling the depth of field and artistic composition objectives.
The parameters $\beta_x$ and $\beta_y$ are other constants necessary to transform the units of the focal length, given in millimeters, to pixels.
Specifically, the ratios, $\beta_x =  \frac{W_{px}}{W_{mm}}$ and $\beta_y = \frac{H_{px}}{H_{mm}}$, relates the width, $W_{mm}$, and the height, $H_{mm}$, of the camera sensor in millimeters with the width, $W_{px}$, and the height, $H_{px}$, of the image in pixels.
The projection also requires the relative position between the camera and the target $t$, denoted by $\mathbf{p}_{dt,k}=\mathbf{R}_{d,k}^T\left(\mathbf{p}_{t,k} - \mathbf{p}_{d,k}\right)$. 

The target position in the image, $\textbf{im}_{t,k} \in \mathbb{R}^2,$ is
\begin{equation}
\textbf{im}_{t,k}=
\lambda\mathbf{K}\
\mathbf{p}_{dt,k},
\label{eq:im_position_image}
\end{equation}
where $\lambda$ is the normalization factor to remove the scale component in the projection.

The subset $\bm{\mu}_{im}$ stores the desired image composition for the target $t$, denoted as $\textbf{im}_{t,k}^*,$ along with the associated weight, represented by $w_{im,t}$, for all scene targets. The cost term penalizes deviations from this composition,
\begin{equation}
J_{im,k} = \sum_{t=1}^{n}{w_{im,t}\|\textbf{im}_{t,k} - \textbf{im}_{t,k}^{*}\|^2},
    \label{eq:im-cost}
\end{equation}

A target can be defined by multiple image coordinates, such as a person's face and body. Our solution also considers the option to control the position of various parts of a target within the image. For instance, a person's face could be positioned in the upper right third, while the knees are aligned with the bottom right third.

\paragraph{Relative position camera-target - Canonical shots} 

The target's depth, $d_{dt,k},$ is the distance between the drone and the target, usually calculated using the Euclidean Distance,  $d_{dt,k} = \lVert\mathbf{p}_{dt,k}\rVert$. It is the only position-related value that cannot be controlled through $J_{im}.$ When combined with a certain value of the focal length, $d_{dt,k}$ affects the amount of effective background visible and the image focus level.

The relative rotation between the camera and target,
$\mathbf{R}_{dt,k}=\mathbf{R}_{d,k}^T\mathbf{R}_{t,k},$
determines the filming perspective.
In the control problem, this is required to enable wide-angle shots and other types of aerial shots. 
This cost term is defined in terms of the subset  $\bm{\mu}_{p}$, that contains the desired values of these two parameters, $d_{dt,k}^{*}$ and $\mathbf{R}_{dt,k}^{*}$ for each target $t$, and their corresponding weights, $w_{R}$ and $w_{d}$,
\begin{equation}
J_{p,k} = \sum_{t=1}^{n}w_{R}
\left\|\mathbf{R}_{dt,k}^T - \mathbf{R}_{dt,k}^{*}\right\|_F  + w_{d}\left(d_{dt,k} - d_{dt,k}^{*}\right)^2,
\label{eq:kind-cost}
\end{equation}
and $\left\|x\right\|_F$ calculates the Frobenius norm of $x$.

\paragraph{Control of the focal length}
To achieve some cinematographic effects, we need to adjust the focal length ($f_k$) of the camera, e.g. zooming in or out, \textit{Dolly Zoom } \cite{liang2020vertigo}, perspective distortion. 
The control of this term drives the focal length to the desired value, $f_k^*$, which is stored in the vector $\bm{\mu}_{f} \in \bm{\mu}$ along with its weight $w_f$,
\begin{equation}
J_{f,k} = w_{f}\left(f_{k} - f_{k}^{*}\right)^2,
    \label{eq:jf-cost}
\end{equation}

\subsubsection{Optimizer}
The MPC problem needs an optimizer that iterates the cost function and calculates the next control actuators that minimize that function in the future. The optimizer is a decision of implementation.
In CineMPC, as the proposed optimization problem is non-linear, we use Ipopt (Interior Point OPTimizer) \cite{wachter2006implementation}, but this does not preclude the use of any other existing optimization libraries.

\subsubsection{Constraints}
\label{sec_constraints}
The constraints are defined as a set of inequalities $g(\bm{\mathcal{C}},\mathbf{u}_{d,k},\mathbf{u}_{c,k},\mathbf{x}_{d,k},\mathbf{x}_{c,k}) \ge 0$ that depend on the states, the inputs and the set $\bm{\mathcal{C}},$ specified by the user.

First, we consider upper and lower bounds on the control inputs,
$\mathbf{u}_{\min}$ and $\mathbf{u}_{\max}$ in $\bm{\mathcal{C}},$
that depend on the cinematographic platform in which CineMPC is used, and are used to guarantee that the commands are physically feasible,
\begin{equation}
\label{Eq:inputConstraints}
\begin{aligned}
    &\mathbf{u}_{d,k} - \mathbf{u}_{d,\min} \ge 0,\quad
    \mathbf{u}_{d,\max} - \mathbf{u}_{d,k} \ge 0,\\
    &\mathbf{u}_{c,k} - \mathbf{u}_{c,\min} \ge 0,\quad
    \mathbf{u}_{c,\max} - \mathbf{u}_{c,k} \ge 0.
\end{aligned}
\end{equation}

Similarly, we consider the possibility of adding upper and lower bounds and state constraints, e.g., to maintain the gimbal rotation in the allowed range,
\begin{equation}
\label{Eq:stateConstraints}
\begin{aligned}
    &\mathbf{x}_{d,k} - \mathbf{x}_{d,\min} \ge 0,\quad
    \mathbf{x}_{d,\max} - \mathbf{x}_{d,k} \ge 0,\\
    &\mathbf{x}_{c,k} - \mathbf{x}_{c,\min} \ge 0,\quad
    \mathbf{x}_{c,\max} - \mathbf{x}_{c,k} \ge 0.
\end{aligned}
\end{equation}

The next set of constraints is used to prevent collisions of the drone with the targets,
\begin{equation}
\label{Eq:collisionConstraints}
  d_{dt,k} - d_{\min} \ge 0,
\end{equation}
where $d_{\min}$ is the desired safety distance, introduced in $\bm{\mathcal{C}}$. 

Finally, we consider a last set of constraints to handle potential occlusions of the targets.
Using $\mathbf{x}_{t,k}$, the target height and width, $t_h$ and $t_w$ of Eq.~\eqref{Eq:TargetInfo}, we can predict the bounding box of each target in the image using Eq.~\eqref{eq:im_position_image}, which is represented by the left-top and right-bottom pixels in the image,
$\textbf{im}_{t,k}^{lt}$ and $\textbf{im}_{t,k}^{rb}.$

Ideally, to guarantee occlusion-free trajectories, the bounding boxes of two targets should not intersect at any time. This can be formally introduced in the problem with a strong increase of the computational load, transforming it into a mixed-integer linear program. We consider instead a simplification that seems to work well in our experiments without increasing the complexity.

Before solving an instance of the optimization problem, we check the relative location of each pair of bounding boxes and analyze the potential risk of occlusion. We describe the process for the left-top horizontal coordinate of the bounding box, noting that the process can be done analogously for the other three coordinates of interest.
Let $x_{ti,k}^{lt}$ represent the horizontal coordinate of $\textbf{im}_{ti,k}^{lt}$ -the left-top of the target $i$-. Then we include the following constraint,
\begin{equation}
\label{eq:occ_constraints}
x_{t1,k}^{lt} - x_{t2,k}^{br} \ge 0,
\end{equation}
if and only if the next condition holds for the initial configuration,
\begin{equation}
\label{eq:occ_constraints2}
(y_{t1,k_0}^{lt} > y_{t2,k_0}^{br}) \wedge (y_{t2,k_0}^{lt} > y_{t1,k_0}^{br})  \wedge  (x_{t2,k_0}^{lt} > x_{t1,k_0}^{br}), 
\end{equation}
using the same notation as Eq. \ref{eq:occ_constraints}.
Otherwise, we neglect the chance of occlusion and do not include this constraint.

\subsection{Low-Level control}
\label{Sec:lowlevel}
The MPC calculates $N$ high-level control actions that the drone should execute every $\Delta_T$ seconds.
Since these actions are computed considering a simplified motion model,
we include a low-level controller that transforms the MPC commands into actuator commands to achieve smooth trajectories that ensure suitable footage.

For the rotation and the intrinsics of the camera, we use linear interpolation to split each command into $m$ smaller portions that are sent to the drone with a higher frequency, which corresponds to $\Delta_T/m$. 
The choice of linear interpolation is made to prevent the overstepping of the commanded values, which would imply shaky images. Similarly, to smooth the position of the drone, we use a standard low-level drone controller that receives position key-points or velocity commands and transforms them into commands that the drone executes following smoother high-level trajectories.

%% file: 06_Perception.tex
\begin{figure}[h!]
\centering    
\includegraphics[width=\columnwidth]{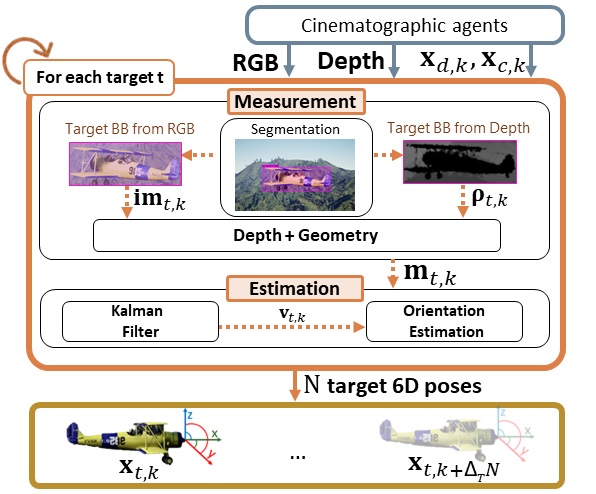}
\caption{\footnotesize \textbf{Perception module diagram.} Components of the perception module and their interaction. The module consists of depth and position measurements, along with the estimation of the targets' next poses. The diagram displays inputs from other modules and the outputs of this module.
}
\label{fig:perception_graph}
\end{figure}

The perception module estimates the poses of all targets from step $k$ until $k+\Delta_T N$ from the RGB images and depth data recorded by the drone. Figure \ref{fig:perception_graph} summarizes the process. As detailed next, the perception process is done in two steps, measurement and estimation.
\subsection{Measurement}
The measurement process receives RGB-D images from the camera, extracting the relative position of the present targets. 
\subsubsection{Detection of targets position in the thin-lens image}
In this step, the system extracts the position in pixels of the targets present in the image. Our implementation uses Yolo\cite{redmon2018yolov3}, an off-the-shelf deep-learning approach for image segmentation. This choice is motivated by its low computational demand and the capacity to detect the targets even if they appear out of focus, as required by filming instructions \cite{zheng2021deblur}.

We only consider the detections that belong to $t_{nature}$. For the remaining steps in the perception module methodology, we use the bounding box provided by the detector to select a pixel that identifies the target, $\mathbf{im}_{t,k}$. 
The decision on this pixel varies depending on the target nature, and $t_w$ and $t_h$.

\subsubsection{Detection of targets depth}
To obtain the 3D position, we extract the relative distance between the drone and the target from the depth image.
To make the measurement robust to noise, the depth value, $\rho_{t,k}$, is defined by the median of the minimum depth of each row of the bounding box, 
\begin{equation}
\begin{aligned}
\rho_{t,k}= \underset{i}{\mathrm{median}}
    \left( \min_{j}\left(\left[\mathbf{D}_{t,k}\right]_{i,j}\right)
    \right)
\end{aligned},
\label{eq:depth}
\end{equation}
where $\left[\mathbf{D}_{t,k}\right]_{i,j}$ represents the pixel in row $i$ and column $j$ of the bounding box in the depth image.
This method helps to filter depth values from the background or foreground, as well as noisy readings.
Alternatively, for distant targets where an RGB-D camera may lack sufficient resolution, the issue could be mitigated by employing a LiDAR or a Laser sensor.

\subsubsection{Relative position of the target}
We use the image coordinates of each target, $\mathbf{im}_{t,k}$, its depth $\rho_{t,k}$, and the camera calibration matrix $\mathbf{K}$—which is dependent on the camera intrinsics, i.e., the focal length—to compute the relative position of the target with respect to the drone at time step $k$,
\begin{equation}
\mathbf{p}_{dt,k}=
\rho_{t,k}\mathbf{K}^{-1}\
\mathbf{im}_{t,k}.
\label{eq:im_calc}
\end{equation}
\label{sec:perception_main}
Since the estimation step uses absolute positions, we convert the relative position to world coordinates using geometry,
\begin{equation}
\mathbf{m}_{t,k}=\mathbf{p}_{d,k}+\mathbf{R}_{d,k}\mathbf{p}_{dt,k},
\label{eq:im_calc1}
\end{equation}
and the drone pose in the world, which is assumed to be available.
The reason why we use absolute instead of relative positions is detailed in the next subsection. 

\subsection{Estimation}
The estimation process receives 3D absolute position measurements of the targets. With this information, it estimates the position, velocity, and orientation of the targets for the next $N$ time steps, which are incorporated into the control module.
\subsubsection{Kalman Filter (KF)}
The central element of this process is a Kalman Filter.
The filter's state is defined by the position of the target in the world, $\mathbf{p}_{t,k}$, and its velocity, $\mathbf{v}_{t,k}$.
The motion model for the prediction stage considers a double integrator with noise defined as small accelerations. The measurement used in the correction is the absolute targets' position in the world, $\mathbf{m}_{t,k}$ provided by the Measurement module.

\subsubsection{Estimation of targets' orientation}
\label{sec:estimation_orientation}
Extracting the targets' orientation from an RGB-D is not trivial. In cinematographic applications, the view directions are typically aligned with the movement direction of a target, i.e., recording a car from its front. Thus, it is reasonable to associate the rotation of a target in terms of its frontal plane, which matches the movement plane of the target. We use the targets' velocity from the Kalman Filter and vector algebra to construct the rotation matrix associated with each target.

The three velocity vectors that form the targets' rotation matrix in the world, $\mathbf{R}_{t,k}$, are normalized and orthogonal to each other. The first vector, $\mathbf{r}_{1,k}$, represents the estimated velocity of the target in the world and is a component of the Kalman Filter state: $\mathbf{r}_{1,k} = \mathbf{v}_{t,k} \in \mathbb{R}^{3}$. The remaining two vectors are calculated using vector algebra as follows. First, we associate the gravity vector, which is always pointing to the ground, to the target, $\mathbf{g}_{k} = [0,0,-1]$. We obtain the second vector by taking the cross product of  $\mathbf{g}_{k}$ and $\mathbf{r}_{1,k}$, $\mathbf{r}_{2,k} = \mathbf{v}_{t,k}  \times \mathbf{g}_{k} $, which is orthogonal to  $\mathbf{r}_{1,k}$. 
 Finally, the cross product of the previous vector $\mathbf{r}_{2,k}$ and the velocity vector $\mathbf{r}_{1,k}$, returns the third vector, $\mathbf{r}_{3,k} = \mathbf{r}_{2,k} \times \mathbf{r}_{1,k}$, which is also orthogonal to $\mathbf{r}_{1,k}$ and $\mathbf{r}_{2,k}$.
The rotation matrix of the target is composed of the three orthogonal vectors, 
\begin{equation}
\begin{aligned}
\mathbf{R}_{t,k}=[\mathbf{r}_{1,k}, \mathbf{r}_{2,k}, \mathbf{r}_{3,k}]
\end{aligned}.
\end{equation}

%% file: 07_Implementation.tex
\begin{table*}[bh!]
\begin{center}
\caption{\textbf{Experiment index}} 
\scriptsize{
\begin{tabular}{| c | c | c | c | c | c | c | c | c | c | c | c | c | c | c | c | c |}
\hline 
\multirowcell{2}{\textbf{Experiment} \\ \textbf{(Scenario)}}
   & \head{0.8cm}{Real / Sim}  
   & \head{1.1cm}{No. Sequences}  
   & \head{1.1cm}{Control Extrinsics}  
   &  \head{1.0cm}{Control Intrinsics} 
   & \head{0.75cm}{Control $J_{p}$} 
   & \head{0.75cm}{Control $J_{DoF}$} 
   & \head{0.75cm}{Control $J_{im}$} 
   & \head{0.75cm}{Control $J_{f}$}
   & \head{0.95cm}{Dynamic targets} 
   & \head{1.15cm}{Multitarget} 
   & \head{0.8cm}{Blurred effects} 
   & \head{1.1cm}{Obs/Occ avoidance}   \vline \\ \hline
  
\addstackgap[3.1pt]{Experiment 1 (A)} 
& \cellcolor{olive!25}\bf{S}
& \bf{4}
& \cellcolor{green!15}\faCheckSquare
& \cellcolor{green!15}\faCheckSquare
& \cellcolor{green!15}\faCheckSquare
& \cellcolor{white} 
& \cellcolor{green!15}\faCheckSquare 
& \cellcolor{white}
& \cellcolor{green!15}\faCheckSquare 
& \cellcolor{white} 
& \cellcolor{white} 
& \cellcolor{white} \\ \hline  

\addstackgap[3.1pt]{Experiment 2 (A)} 
& \cellcolor{olive!25}\bf{S}
& \bf{4}
& \cellcolor{green!15}\faCheckSquare
& \cellcolor{green!15}\faCheckSquare
& \cellcolor{green!15}\faCheckSquare
& \cellcolor{white} 
& \cellcolor{green!15}\faCheckSquare 
& \cellcolor{white} 
& \cellcolor{green!15}\faCheckSquare 
& \cellcolor{white} 
& \cellcolor{green!15}\faCheckSquare 
& \cellcolor{white} \\ \hline  

\addstackgap[3.1pt]{Experiment 3 (B)}
& \cellcolor{olive!25}\bf{S}
& \bf{2}
& \cellcolor{green!15}\faCheckSquare 
& \cellcolor{green!15}\faCheckSquare
& \cellcolor{green!15}\faCheckSquare  
& \cellcolor{green!15}\faCheckSquare 
& \cellcolor{green!15}\faCheckSquare  
& \cellcolor{green!15}\faCheckSquare 
& \cellcolor{white}
& \cellcolor{white}
& \cellcolor{green!15}\faCheckSquare 
& \cellcolor{white} \\ \hline 

\addstackgap[3.1pt]{Experiment 4 (B)}
& \cellcolor{olive!25}\bf{S}
& \bf{2}
& \cellcolor{green!15}\faCheckSquare 
& \cellcolor{green!15}\faCheckSquare
& \cellcolor{green!15}\faCheckSquare  
& \cellcolor{green!15}\faCheckSquare 
& \cellcolor{green!15}\faCheckSquare  
& \cellcolor{green!15}\faCheckSquare 
& \cellcolor{white}
& \cellcolor{white}
& \cellcolor{green!15}\faCheckSquare 
& \cellcolor{green!15}\faCheckSquare  \\ \hline 

\addstackgap[3.1pt]{Experiment 5 (C)} 
& \cellcolor{green!15}\bf{R}
& \bf{4}
& \cellcolor{white} 
& \cellcolor{green!15}\faCheckSquare
& \cellcolor{white}
& \cellcolor{white} 
& \cellcolor{green!15}\faCheckSquare 
& \cellcolor{green!15}\faCheckSquare 
& \cellcolor{white}
& \cellcolor{green!15}\faCheckSquare 
& \cellcolor{green!15}\faCheckSquare 
& \cellcolor{white} \\ \hline 

\addstackgap[3.1pt]{Experiment 6 (D)} 
& \cellcolor{green!15}\bf{R}
& \bf{2}
& \cellcolor{green!15}\faCheckSquare
& \cellcolor{green!15}\faCheckSquare
& \cellcolor{green!15}\faCheckSquare
 & \cellcolor{green!15}\faCheckSquare  
 & \cellcolor{green!15}\faCheckSquare  
 & \cellcolor{green!15}\faCheckSquare  
 & \cellcolor{white} 
 & \cellcolor{white} 
& \cellcolor{white} 
 & \cellcolor{white} \\ \hline 

\addstackgap[3.1pt]{Experiment 7 (D)} 
& \cellcolor{green!15}\bf{R}
& \bf{2}
& \cellcolor{green!15}\faCheckSquare
& \cellcolor{green!15}\faCheckSquare
& \cellcolor{green!15}\faCheckSquare
 & \cellcolor{green!15}\faCheckSquare  
 & \cellcolor{green!15}\faCheckSquare  
 & \cellcolor{green!15}\faCheckSquare  
 & \cellcolor{white} 
 & \cellcolor{white} 
& \cellcolor{white} 
 & \cellcolor{white}  \\ \hline 
 \end{tabular}
 }
\label{table:experiments}
\end{center}
\end{table*}

\begin{figure}[!bt]
         \includegraphics[width=0.99\columnwidth]{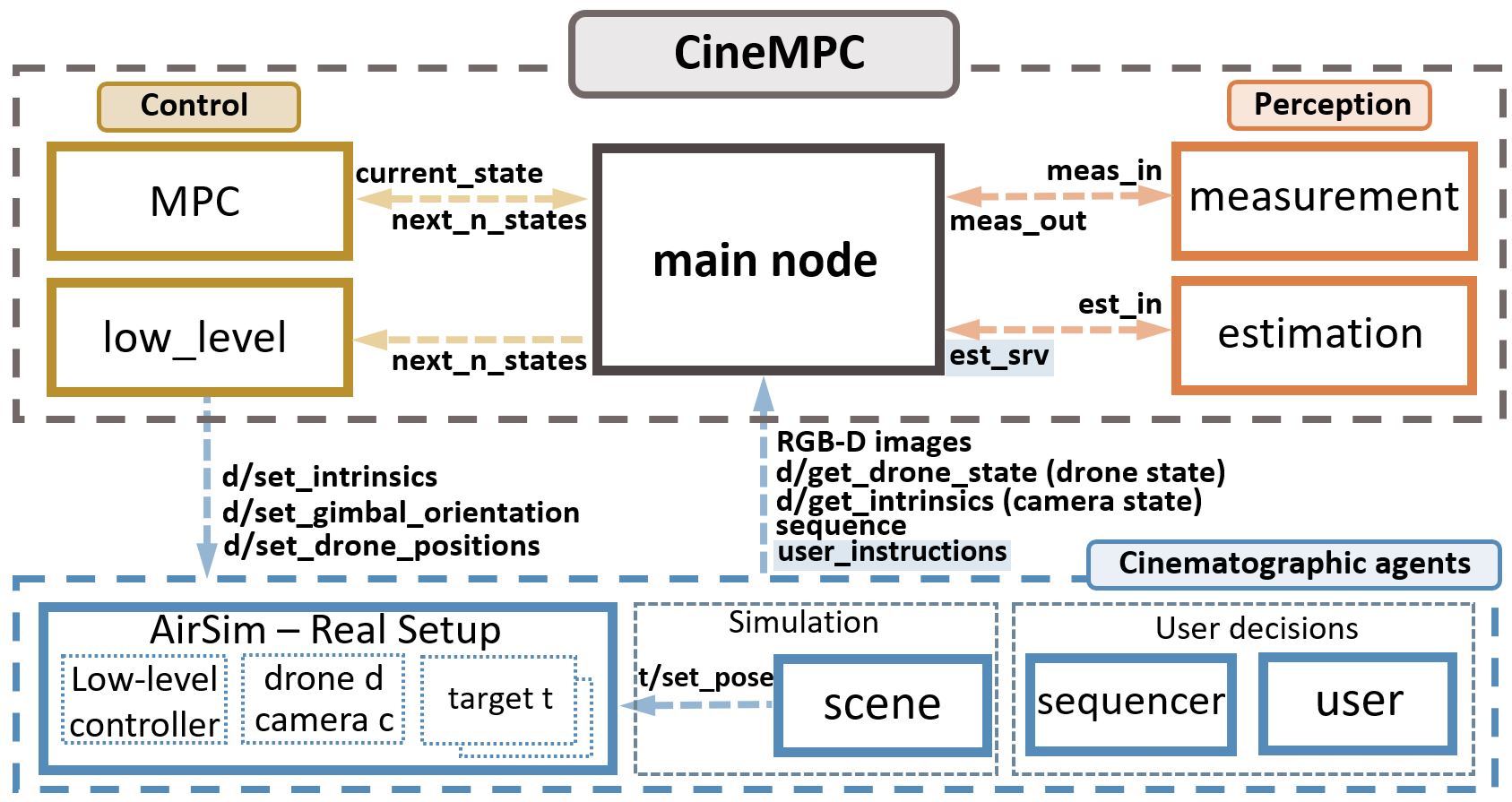}
	\caption{\footnotesize \textbf{ROS software architecture of CineMPC}. Implementation components of CineMPC. The square solid lines represent the nodes of each module. The nodes communicate through ROS topics and services, depicted using lines. Similarly to Fig. \ref{fig:system_overview}, colors denote the system module of each ROS component: control module (yellow), perception module (orange), and cinematographic agents (blue). The main node has parts of every module. }
	\label{fig:graph_ROS}
\end{figure}

To promote the widespread use of CineMPC, we release a publicly available implementation that is integrated with the ROS (Robotic Operating System)~\cite{koubaa2017ros} environment.
Following standard ROS design principles, a simplified version of the system architecture is depicted in Fig. ~\ref{fig:graph_ROS}, showcasing key ROS nodes, topics, and services. Communication arrows represent topics, which facilitate information sharing between nodes, while services, enabling information exchange in a server-client system, are denoted with a squared form.

The code also contains all the necessary elements to be run inside the photo-realistic and popular  AirSim simulator~\cite{shah2018airsim}.

The next subsections include a description of the main nodes together with their related topics and services.  

\subsection{CineMPC}
CineMPC receives the RGB-D images from the camera, the drone and camera state, and the user instructions. It returns the next inputs for the drone and the camera to record the scene and targets according to the instructions and constraints. 
Using this input/output structure, CineMPC is transparent to the platform where it is used, i.e., simulation or real drones.

\subsubsection{\textbf{Control module}}
These two nodes, shown in yellow in Fig. \ref{fig:graph_ROS}, implement the control module described in Section~\ref{sec_control_problem}. 
The {\tt \small \textbf{/MPC}} node implements the MPC solver, and the {\tt \small \textbf{/low\_level}} node generates low-level commands to ensure a smooth trajectory and recording.
The first node reads the estimation of the $N$ future target states from the topic {\tt \small /current\_state} and calculates the $N$ high-level trajectory commands. These are sent to the drone and camera through the {\tt \small{/next\_n\_states}} topic every $\Delta_T$ seconds. This is then used in the second node in a linear or spline interpolation to split each command into several, sent to the drone and the camera at a higher frequency. 
The specific topics to which these commands are sent will depend on the platform used.

\subsubsection{\textbf{Perception module}}

This module implements the extraction of the pose of the targets from camera images, as detailed in Section \ref{sec_perception}. The two nodes are highlighted in orange in Fig. \ref{fig:graph_ROS}. The {\tt \small \textbf{/measurement}} node implements the measurement process of the perception module, while the {\tt \small \textbf{/estimation}} node is responsible for filtering noise and predicting the future steps of the targets.

The first node receives input from a message containing RGB and depth images and the drone state through the topic  {\tt \small meas\_in}. 
To detect the bounding boxes of the targets, the node uses Darknet ~\cite{darknet13}, a C++ implementation of Yolo~\cite{redmon2018yolov3}. 
The node outputs a message containing the list of absolute positions for each target, published in the topic  {\tt \small meas\_out}.

The second node runs a Kalman Filter for each target, estimating their next $N$ poses, as detailed in Sec.\ref{sec:estimation_orientation}. The input of this node is the target's absolute position in the message {\tt \small /est\_in}. This node outputs the next $N$ absolute poses of the target, through service {\tt \small /est\_srv}.

\subsubsection{\textbf{System synchronization and frequency - Main node}}
 Each topic is published at a different frequency. Thus, ROS is in charge of the synchronization. The {\tt \small \textbf{/main node}} acts as a coordinator between the 
 nodes, dispatching the required information to each one at appropriate frequencies. Users should make sure to set $\Delta_T$ to a value higher than the solver's processing time,  dependent on the computer's capabilities. 
\subsection{Cinematographic agents}
The source includes a {\tt \small \textbf{/user}} node to provide the instructions to the rest of the system through the service {\tt \small{/user\_instructions}}, facilitating the introduction of different control objectives. The implementation features a user interface to simplify the definition of $\bm{\mu}$ (Section \ref{sec:user}). This program provides a JSON file to the {\tt \small \textbf{/user}} node containing the instructions.
The {\tt \small \textbf{/sequencer}} node counts the delayed time since the beginning of the execution and splits the recording into {\tt \small sequences}. 
Finally, in the case of the simulated experiments, the {\tt \small \textbf{/scene}} node automatically controls the AirSim scene to enable the recording of dynamic elements.

%% file: 08_Experiments.tex
 This section validates and demonstrates our system's capabilities with several experiments run in simulated (Sec.~\ref{sec_sim_exp}) and real (Sec.~\ref{sec_real_exp}) setups. Table \ref{table:experiments} shows an index of the conducted experiments, including the scenarios in which they were performed (in parentheses), along with the corresponding requisites they cover for clarification.
 We refer the reader to the supplementary video for further visualization of the results of the experiments in simulation and the real world. 

\subsection{Simulation}
\label{sec_sim_exp}
\input{08A_Sim_Experiments}

\subsection{Real Experiments}
\label{sec_real_exp}
\input{08B_Real_Experiments_Real}

%% file: 08A_Sim_Experiments.tex
Experiments in simulation enable more aggressive trajectories for both the cinematographic platform and targets compared to real setups. This allows for a comprehensive analysis of the platform under various situations and constraints. The experiments are conducted in two distinct scenarios. Scenario A focuses on evaluating the performance of the perception module, while scenario B involves testing the control module and conducting a simple user study to validate the system.
\subsubsection{Experimental Setup}
\label{sec:experimental_desktop}

The experiments in simulation are run in Ubuntu 20 in an Intel® Core™ i7-9700 8-Core CPU equipped with 64 GB of RAM and an NVidia GeForce GTX 1070. The experiments of scenario A are simulated at 0.5x speed, with a sample period of $\Delta_T = 0.2 s.$ and time-horizon of $N = 5$ time-steps. In scenario B, the experiments are simulated at 1x speed, with a sample period of $\Delta_T = 0.3 s.$ and time-horizon is $N = 5$ time-steps.
 Depth measurements are perturbed with Gaussian noise of zero mean and standard deviation of $\sigma = 0.04m^2,$ and filtered with a Kalman Filter.
Table \ref{table:camera-params} contains the set of constants described in Sec. \ref{sec_agents} that describe the camera of the simulation environment.
The constraints $\bm{\mathcal{C}}$, (Sec. \ref{sec_constraints}), vary for each platform. For instance, constraints on drone and camera control inputs maintain hardware realism or ensure non-aggressive trajectories, producing smooth footage.
Table \ref{table:system-constraints} details the lower and upper bounds of the system constraints applied in the experiments for our platform.
 Table \ref{table:weights} depicts the cost function weights for each sequence of experiments E1 and E3. 
 Each experiment is conducted multiple times, with the drone starting from random initial positions, capturing the actor within the field of view. The plots display the mean values across all runs with a solid line, while the standard deviation is represented by a lighter-shaded area.

\subsubsection{Scenario A: plane flight over a forest}
\label{sec:exp_plane}
\paragraph{Goals on this scenario}
We designed this scenario with two goals. The first goal is to test the control of the extrinsic parameters and the focal length by requesting wide variations in recording perspective and image composition. The second goal is to test the perception module and how it integrates with the control under different conditions and perspectives.

 \begin{table}[!b]
\begin{center}
\caption{\textbf{Simulation experiments}. Camera Parameters}
\begin{tabular}{| c | c | c | c | c | c | c | c | c | }
\hline
\multicolumn{2}{|c|}{$px$} & \multicolumn{2}{|c|}{$mm$} & $\beta_x$, $\beta_y$& $c_u$ & $c_v$& $s$& $c$
\\
\hline 

W & H & W & H & \multirow{ 2}{*}{40.40}    &\multirow{ 2}{*}{480}   & \multirow{ 2}{*}{270}  &\multirow{ 2}{*}{0} &\multirow{ 2}{*}{0.03}

 \\
 \cline{0-3}

960 & 540 & 23.76 & 13.365 &  &  &  & & \\ 
\hline
\end{tabular}
\label{table:camera-params}
\end{center}
\end{table}
\begin{table}[!b]
\begin{center}
\caption{\textbf{Simulation experiments}. System Constraints}
\scriptsize{
\begin{tabular}{| c | c | c | c | c | c | c | c | c | c | c | }
\hline
&$\mathbf{p}_d, \mathbf{v}_d, \mathbf{R}_d$ &  $\mathbf{\Omega}_d, \mathbf{a}_d$ & $f, F, A$ & $v_f, v_F, v_A$\\
\hline 
min & $-30, -40, -0.25$& $-0.25, -1$ & $15, 4, 1.2$ & $-7, -15, -3$ \\
\hline 
max & $30, 40, 0.25$& $0.25, 1$ & $500, 2000, 22$ & $7, 15, 3$ \\
\hline
\end{tabular}
}
\label{table:system-constraints}
\end{center}
\end{table}

\begin{table}[!tb]
\begin{center}
\caption{\textbf{Simulation experiments}. Weights of cost function terms}
\begin{tabular}{| c | c | c | c | c | c | c | c |}
\hline
   & $w_{D_{n}}$ & $w_{D_{f}}$ & ${w_{im.x}}$ & ${w_{im.y}}$ & ${w_{d}}$& ${w_{R}}$ & ${w_{f}}$\\ \hline
units & 10 & 10 & 1 & 1 & 10 & 100 & 1 \\
 \hline
E1-seq.1 & 0 & 0 & 1.25 & 0.5 & 20 & 100 & 0 \\
E1-seq.2 & 0 & 0 & 2 & 0.5 & 20 & 2000 & 0 \\
E1-seq.3 & 0 & 0 & 1.5 & 1 & 20 & 200 & 0 \\
E1-seq.4 & 0 & 0 & 1.5 & 0.5 & 20 & 350 & 0 \\
\hline 
E3-seq.1 & 10 & 0 &  0.5  & 1 & 0 & 500 & 10 \\
E3-seq.2 & 10 & 10 & 0.5 & 1.5 & 0 & 500 & 0.75 \\ 

 \hline
\end{tabular}
\label{table:weights}
\end{center}
\end{table}

\paragraph{Experiment 1 (E1) - Filming a moving target from different perspectives}
In this scenario, a plane moves with unknown and varying direction, orientation, and velocity over time, reaching a maximum speed of 10 m/s. 
The experiment consists of four sequences in which recording instructions are modified to capture the plane from different perspectives, e.g., \textit{`High Angle Shot'}, and different image compositions e.g., \textit{`Rule of thirds'}. Consequently, we adjust the desired values associated with the cost terms $J_p$ and $J_{im}$ in each sequence. The weight of $J_p$, $w_{p}$, is higher than $w_{im}$ to highlight its effect on the final recording, as shown in Table \ref{table:weights}. The cost term  $J_p$ controls the desired relative position between the camera and the plane.  Different values of  $\mathbf{R}_{p,k}^{*}$ and $d_{dp}^*$ are considered for each sequence to force the drone to record the plane from different perspectives. 
The term $J_{im}$ controls the position of the plane in the image. We define two targets inside the plane: the middle-top and middle-bottom coordinates of its bounding box. The desired image positions for these targets change in each sequence, varying the vertical composition around the rule of thirds, depending on the portion of the frame that the plane should occupy.

Finally, concerning the intrinsics, this experiment just focuses on the control of the focal length for $J_{im}$. Therefore, the weights associated with $J_{DoF}$ and $J_f$ are set to zero.

\begin{figure}[!bh]
\centering
\begin{tabular}{cc}
    \includegraphics[width=0.46\columnwidth,height=2.3cm]{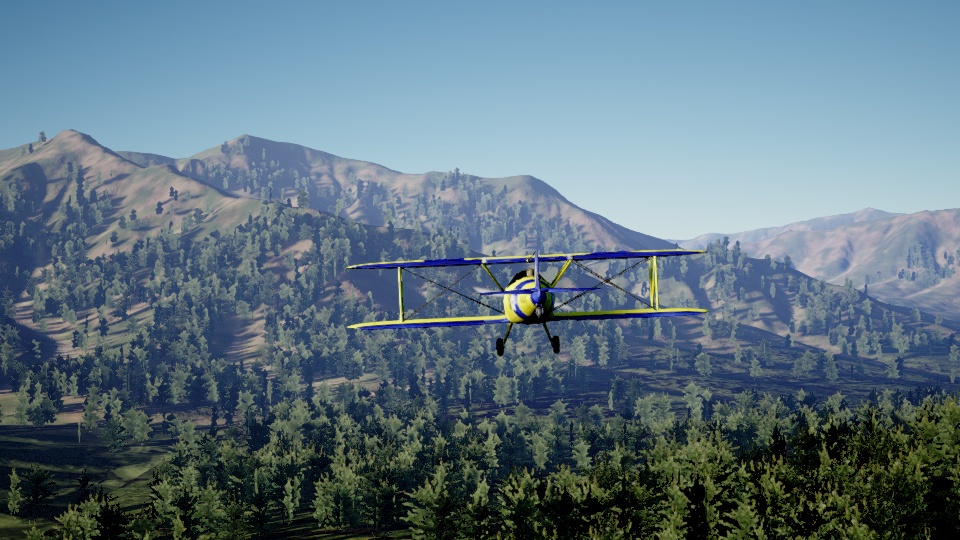}
    & 
    \includegraphics[width=0.46\columnwidth,height=2.3cm]{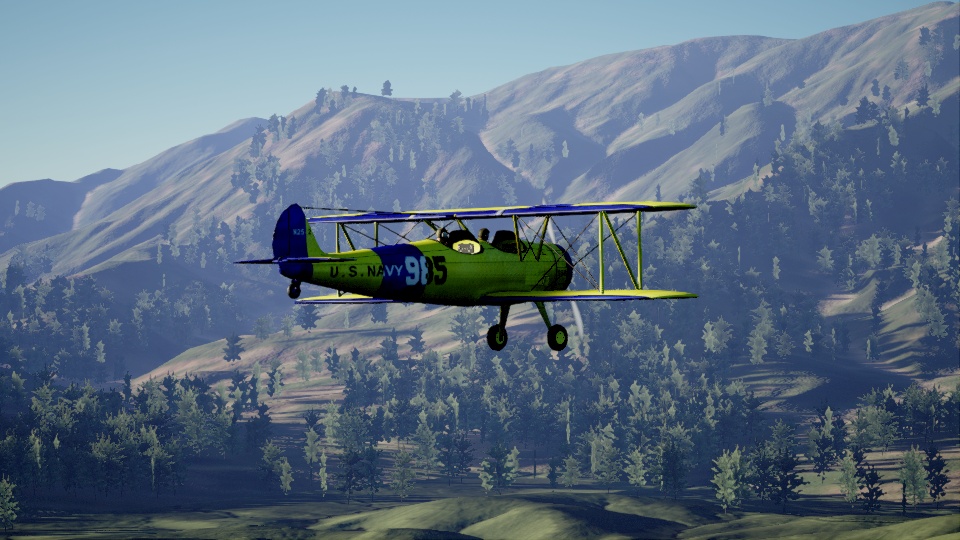} 
    \\\footnotesize (a)& \footnotesize (b) 
    \\ 
    \includegraphics[width=0.46\columnwidth,height=2.3cm]{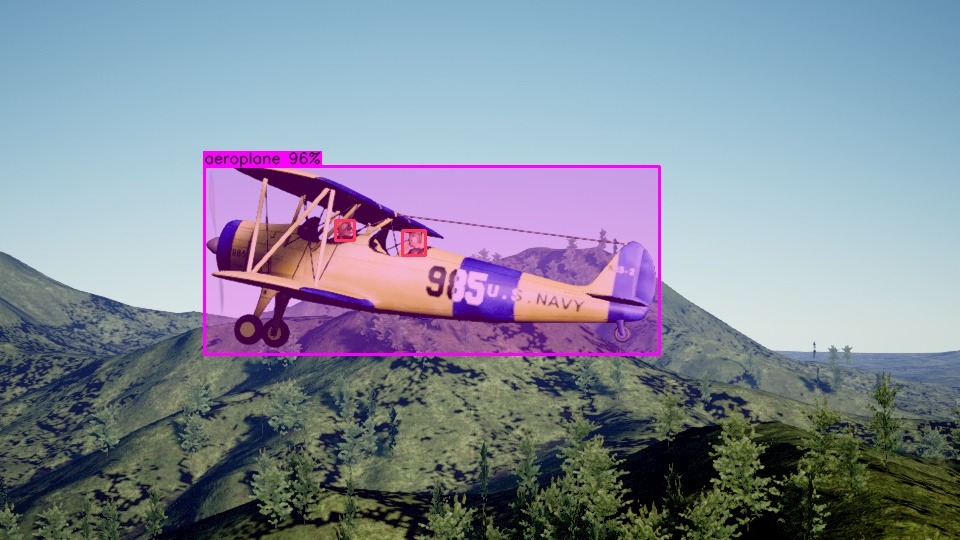}
    &
    \includegraphics[width=0.46\columnwidth,height=2.3cm]{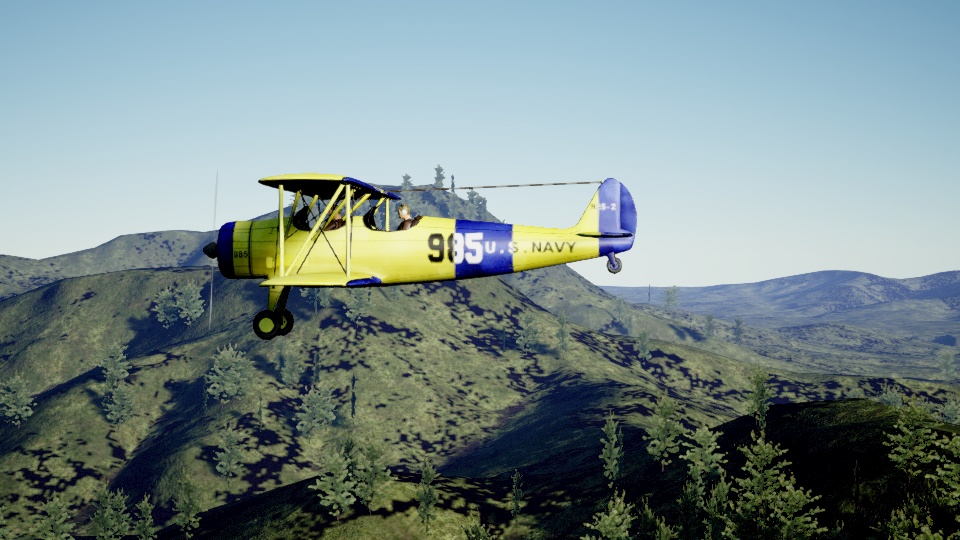}
    \\\footnotesize (c)   & \footnotesize (d) 
    \\
    \includegraphics[width=0.46\columnwidth,height=2.3cm]{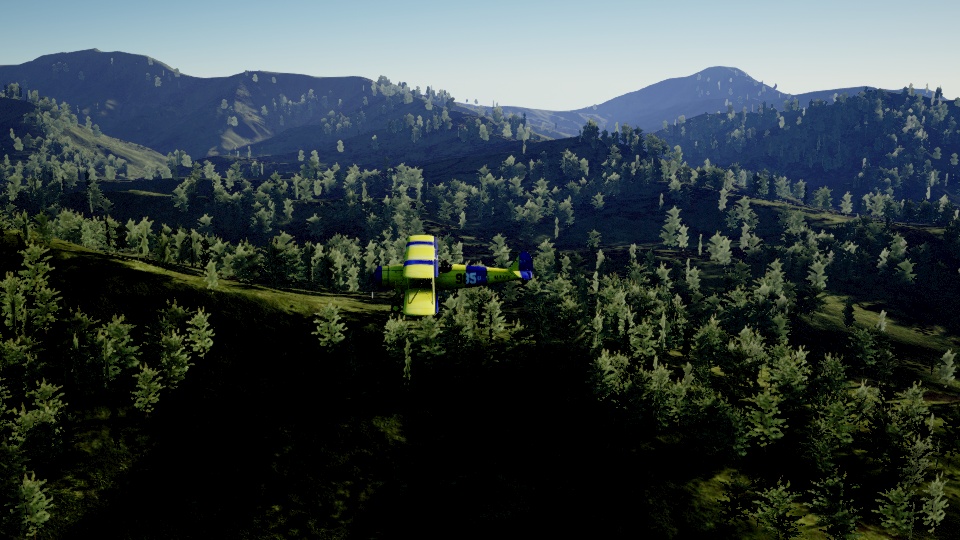}
    &
    \includegraphics[width=0.46\columnwidth,height=2.3cm]{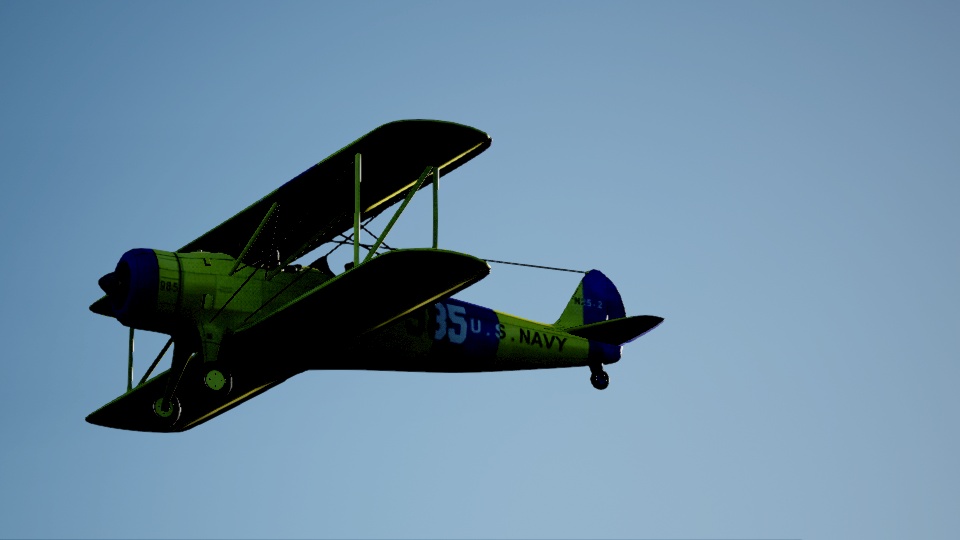}
   \\  \footnotesize (e) & (f)  \footnotesize 
 
\end{tabular}
\caption{\textbf{Experiment 1. Qualitative results:} (a)  Initial frame. (b) End frame of sequence 1. (c) An intermediate frame from sequence 2 along with the bounding box detected by the perception module. (d) End frame of sequence 2. (e) End frame of sequence 3. (f) End frame of sequence 4.
}
\label{fig:plane_quali}
\end{figure}

\begin{figure*}[!h]
\centering
\begin{tabular}{ccc}
    
    \includegraphics[width=0.3\linewidth,height=2.7cm]{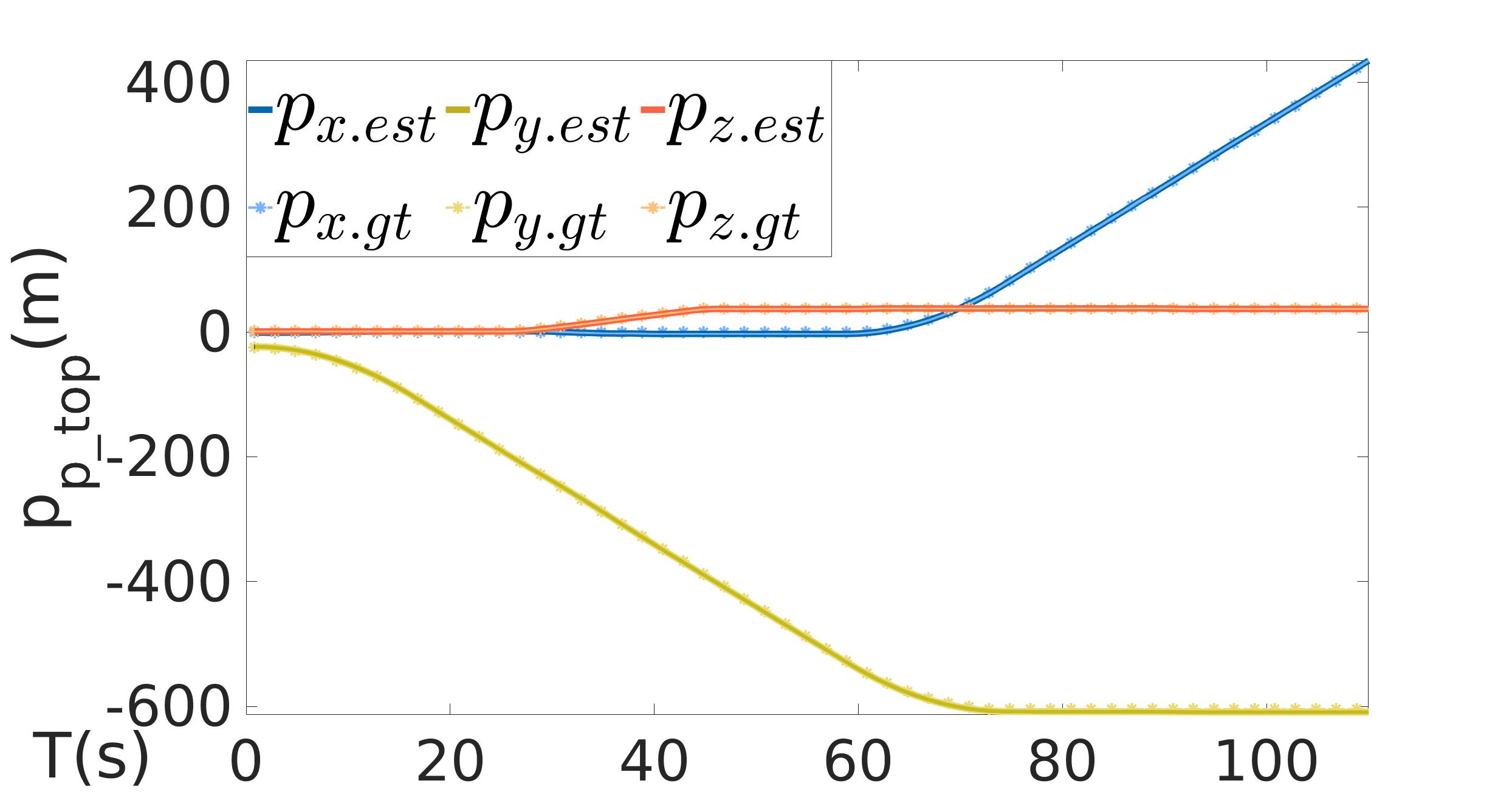} 
     &
    \includegraphics[width=0.3\linewidth,height=2.7cm]{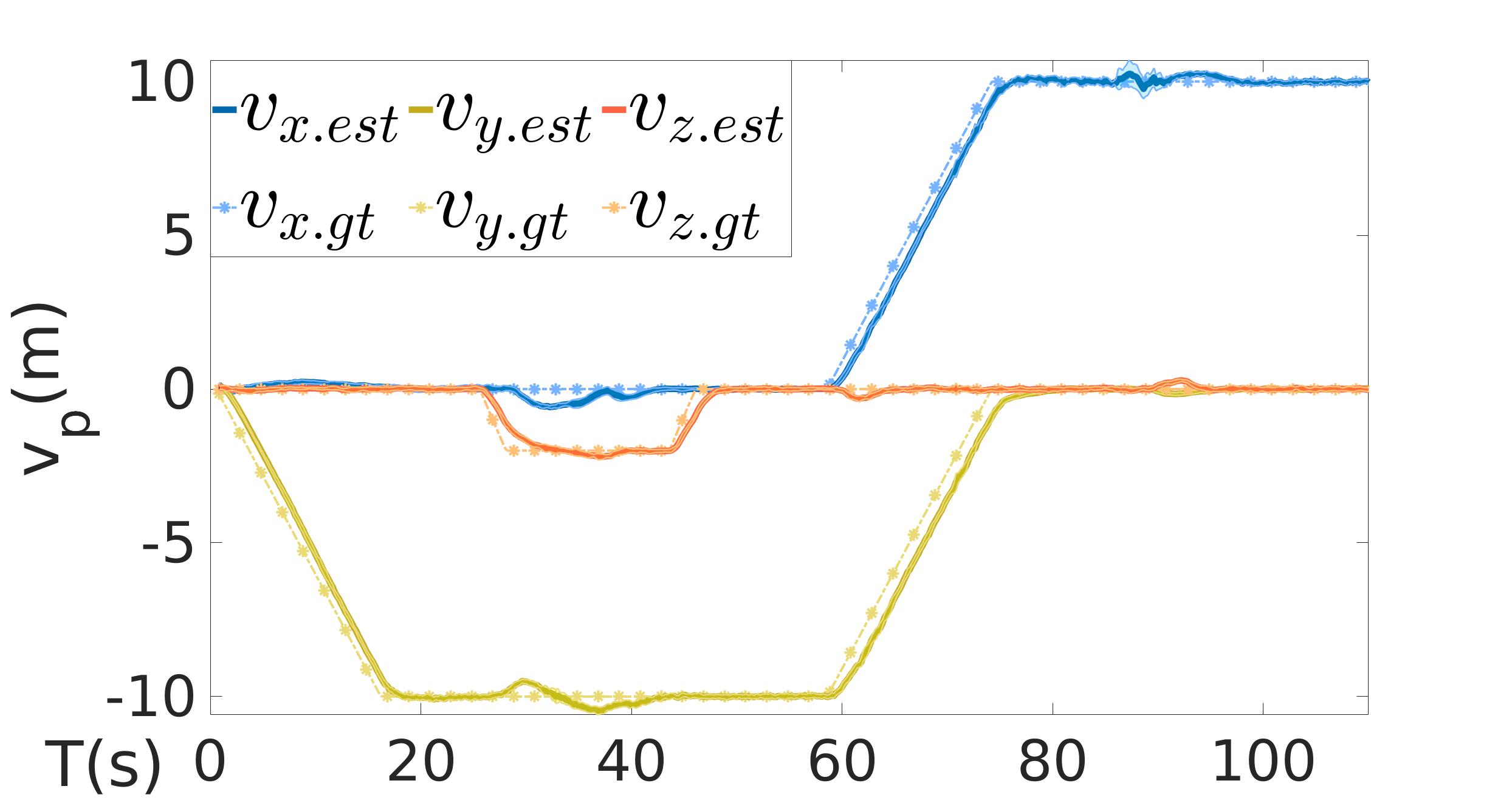}
    &
     \includegraphics[width=0.3\linewidth,height=2.7cm]{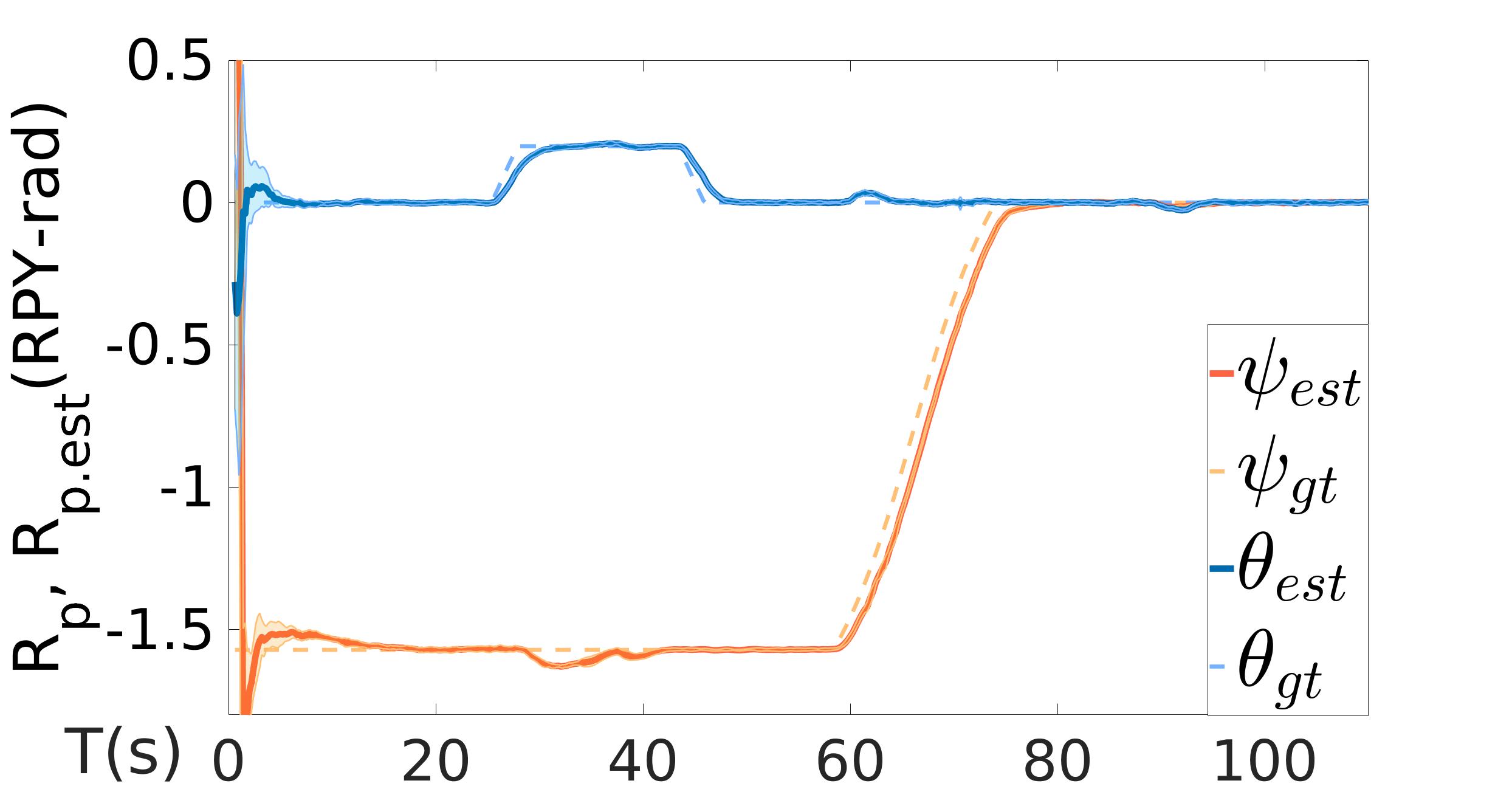}
     
    \\
    (a) & (b) & (c) \\   
    \includegraphics[width=0.3\linewidth,height=2.7cm]{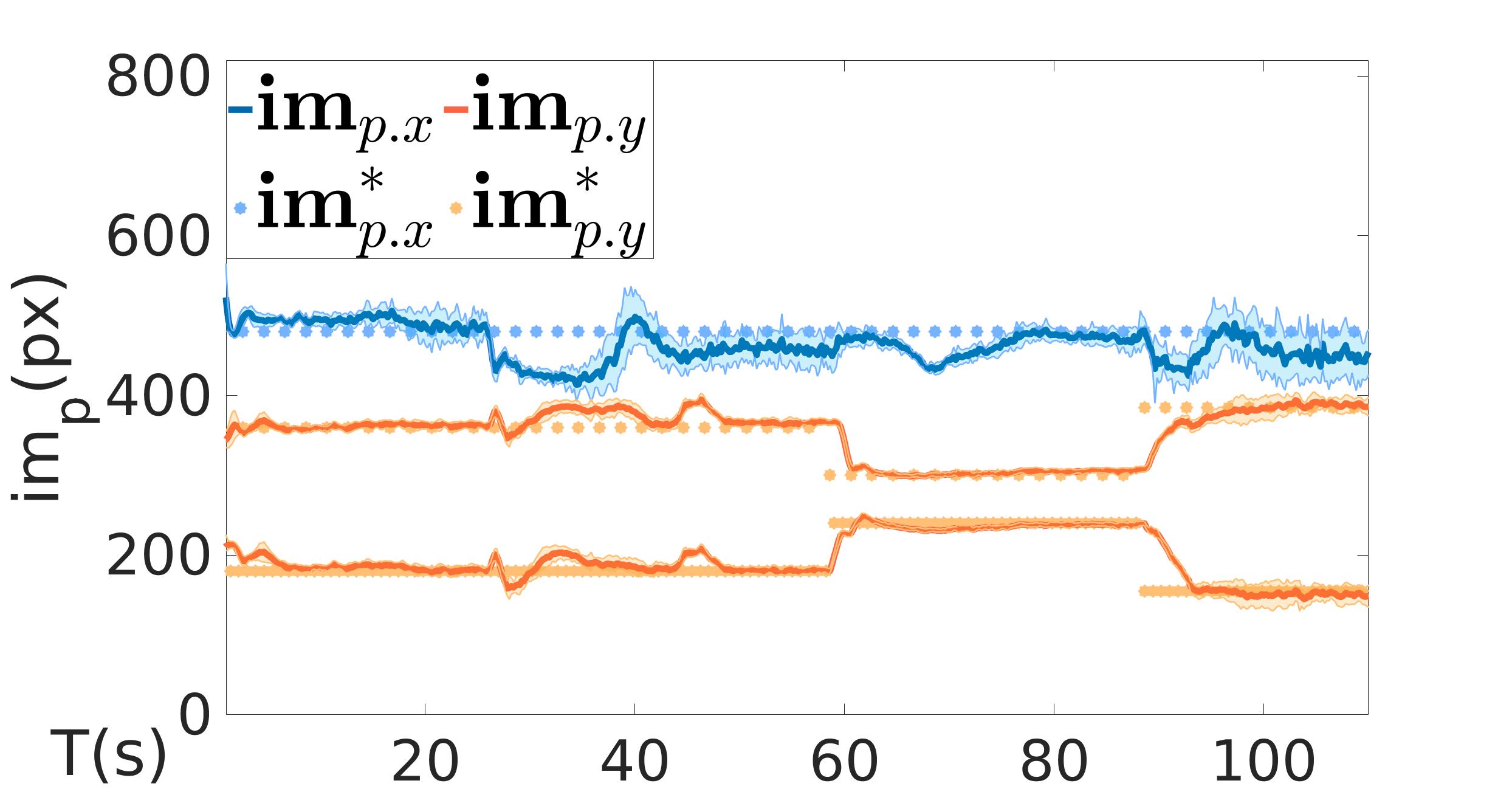}
&
    \includegraphics[width=0.3\linewidth,height=2.7cm]{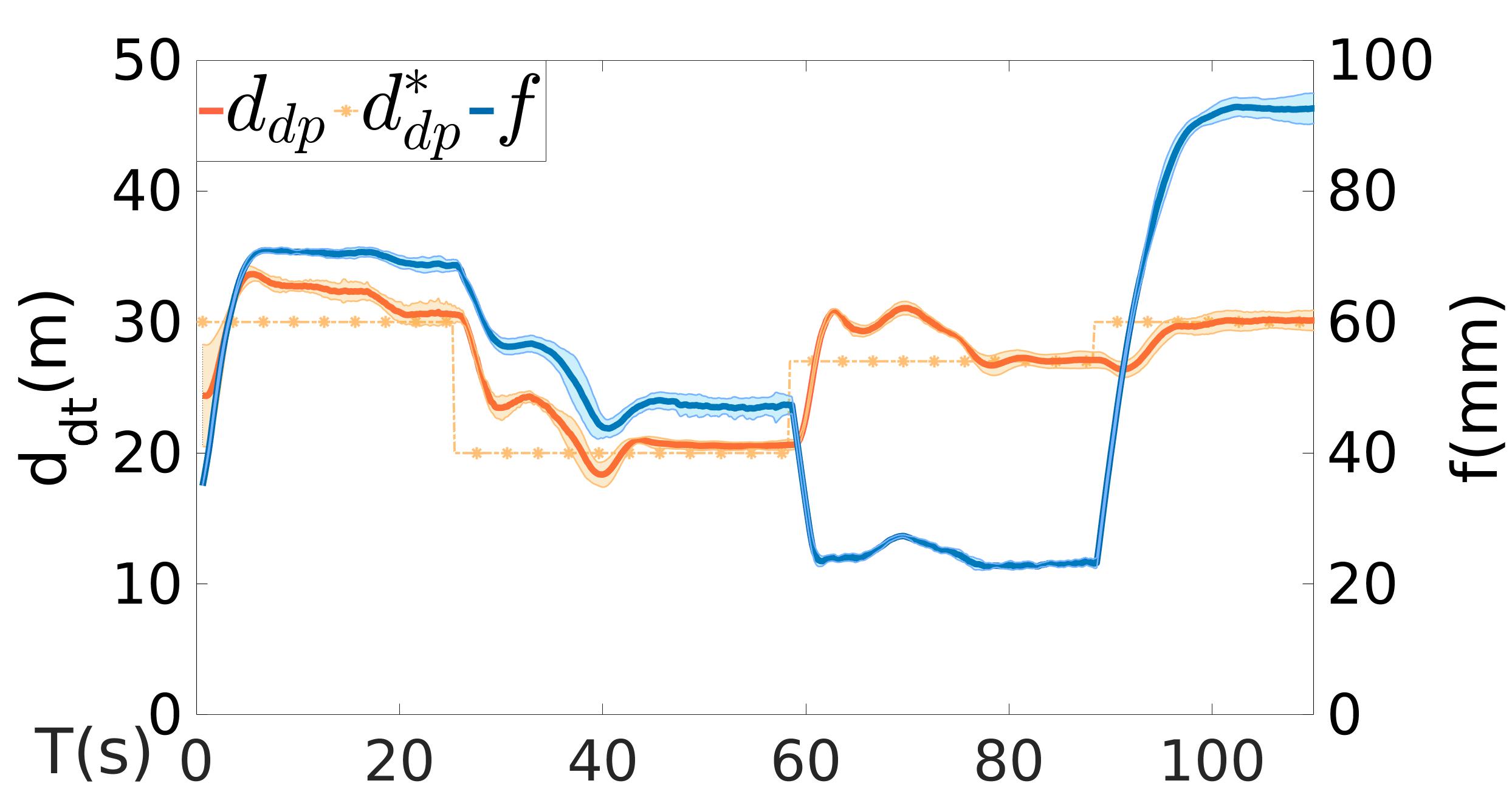}
    &
    \includegraphics[width=0.3\linewidth,height=2.7cm]{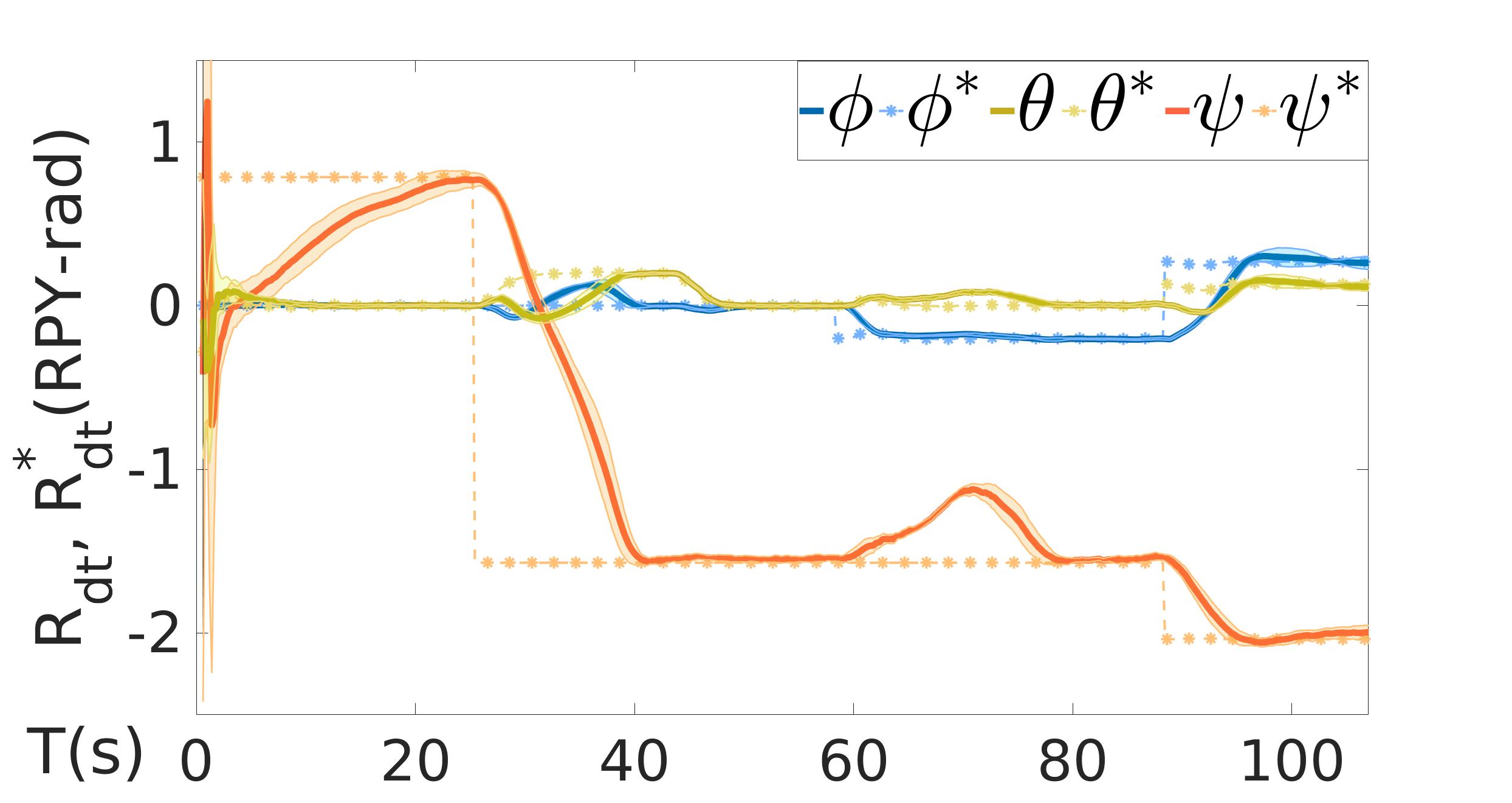}

    \\
    (d) & (e) & (f) \\   
    \includegraphics[width=0.3\linewidth,height=2.7cm]{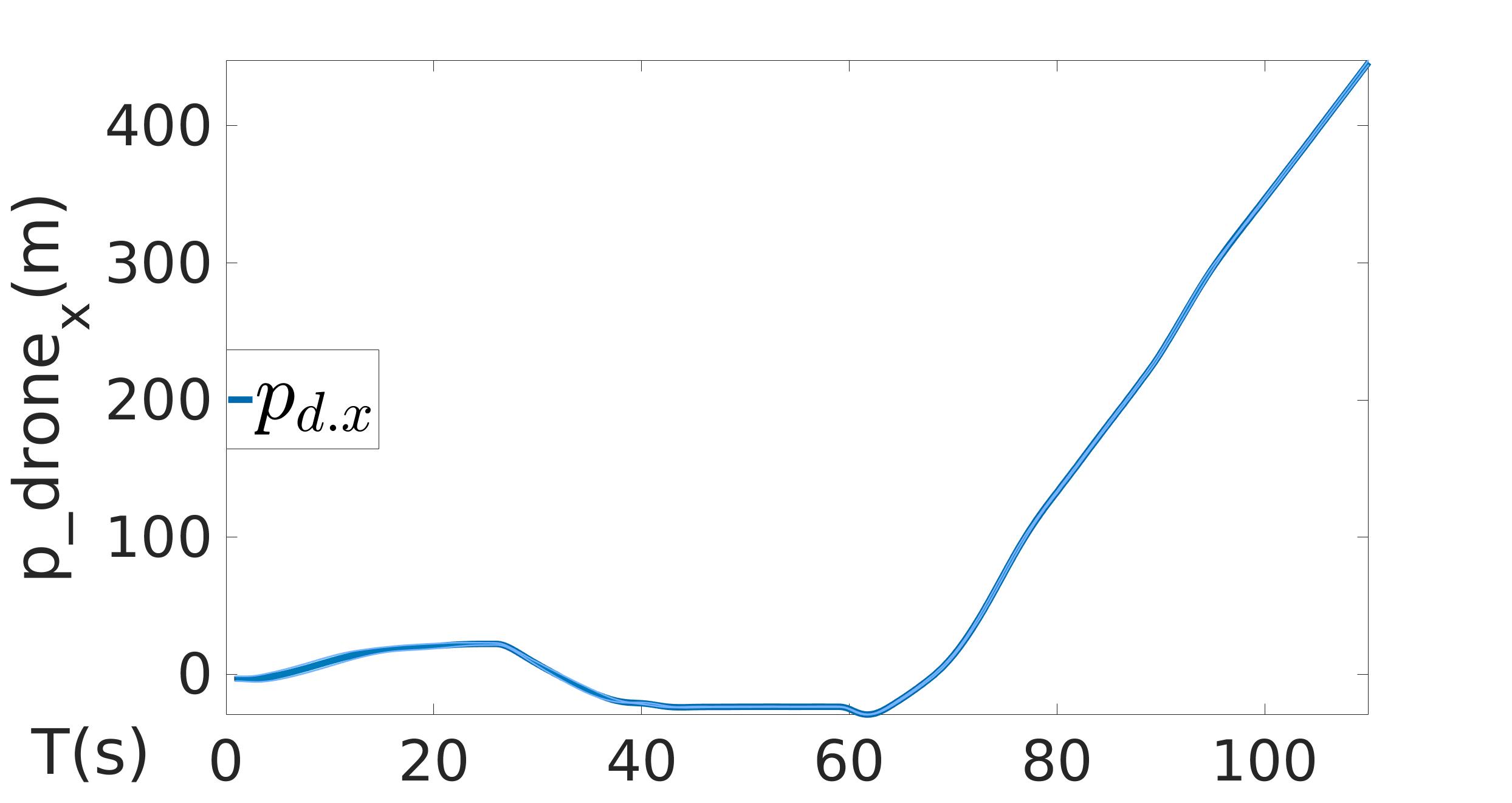}
&
    \includegraphics[width=0.3\linewidth,height=2.7cm]{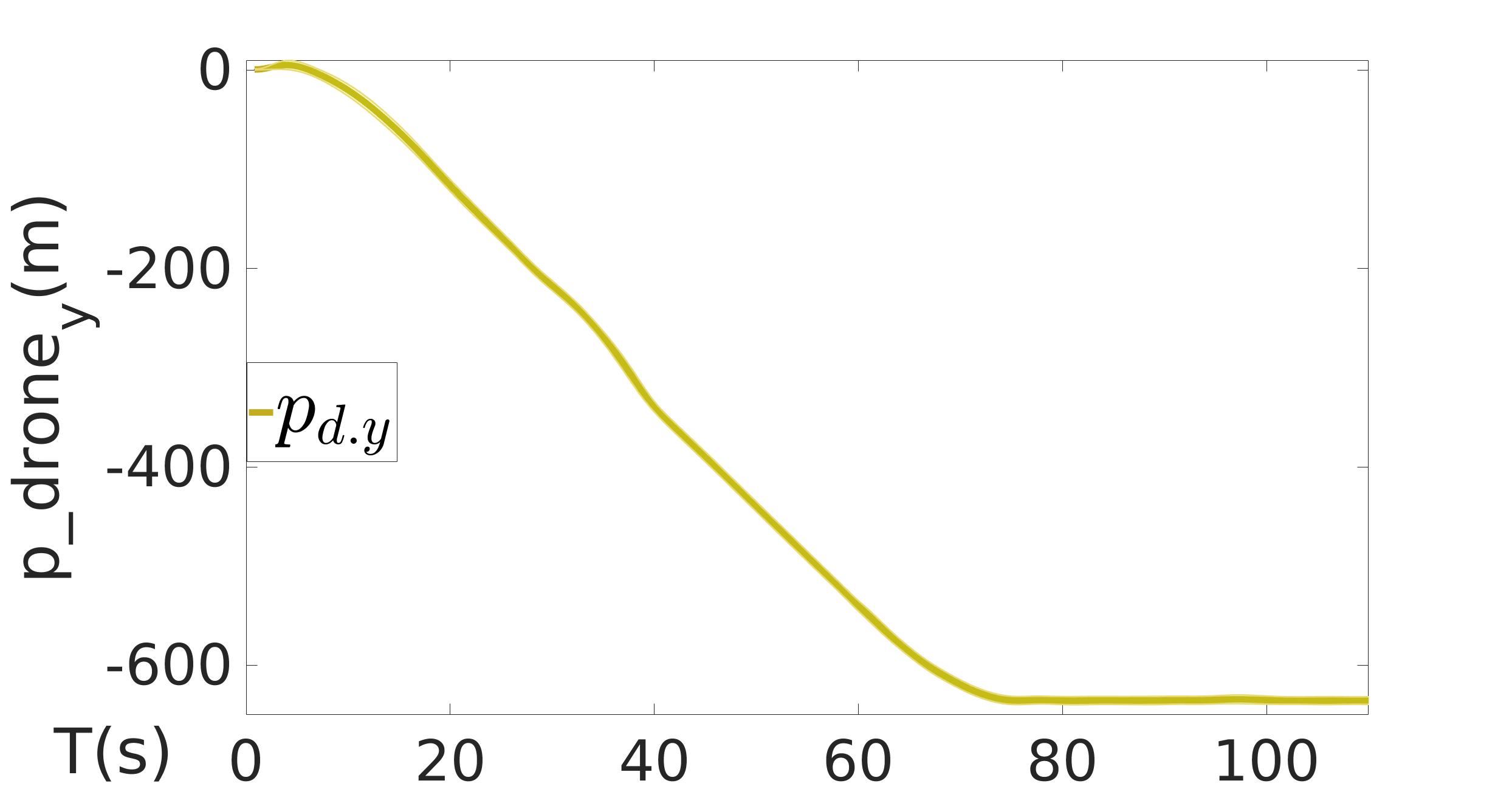}
    &
    \includegraphics[width=0.3\linewidth,height=2.7cm]{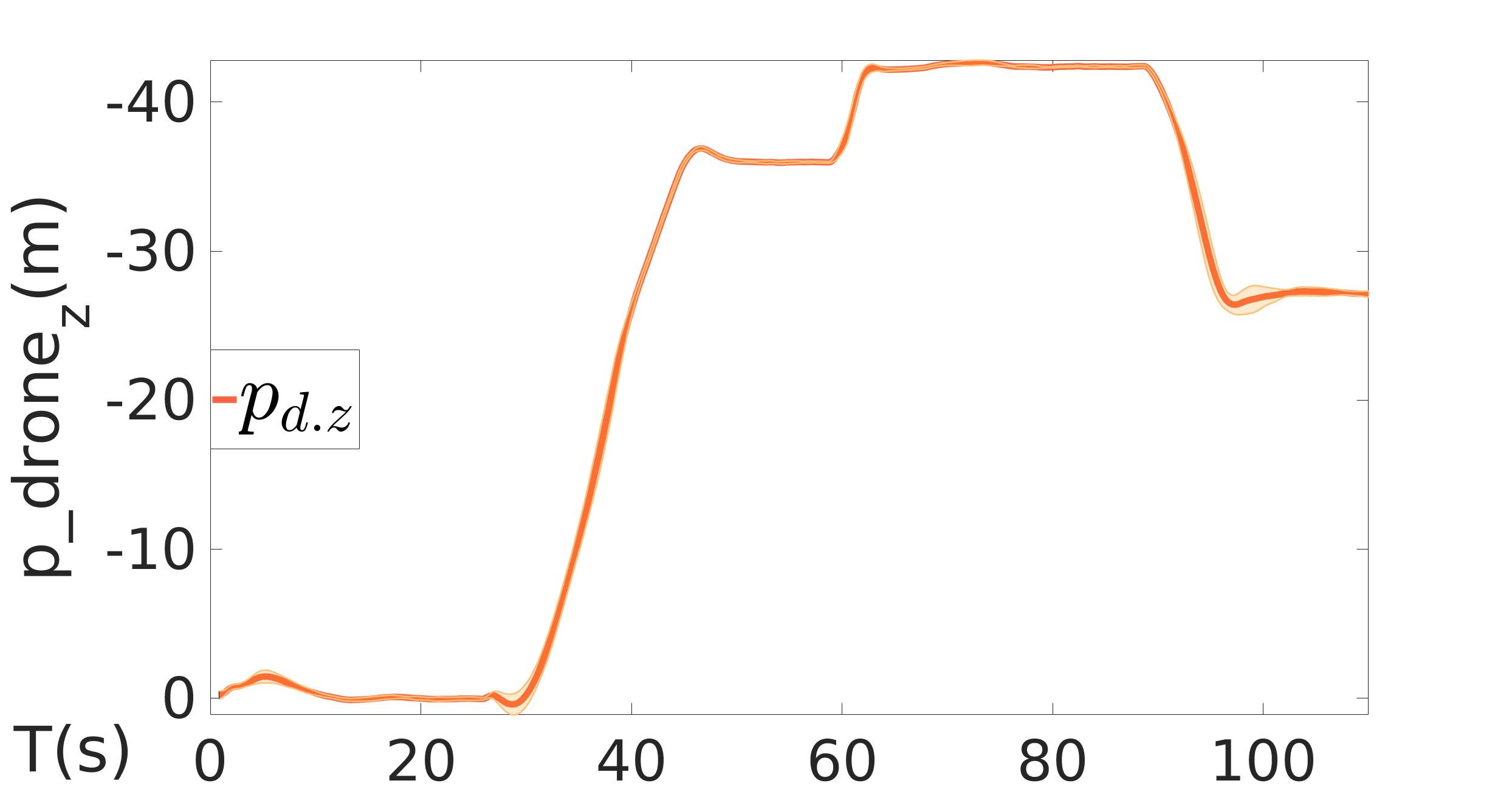}

    \\
    (h) & (i) & (j) 
\end{tabular}
\caption{\textbf{Experiment 1. Quantitative results:} The experiment was conducted 15 times from different starting points. The Z component is intentionally inverted for clarification. Solid lines represent the mean of the plotted value. Lighter areas depict the standard deviation. (a) Evolution of top position of the plane ($\mathbf{p}_{p\_top}$). Solid lines represent the estimated value from the perception module and dashed lines represent ground-truth values. (b) Evolution of the velocity of the plane ($\mathbf{v}_{plane}$). Solid lines depict estimated values extracted from the perception module. Dashed lines represent ground-truth values. (c) Evolution of the absolute rotation of the plane ($\mathbf{R}_p$) in roll, pitch, and yaw. Dashed lines are ground-truth and solid lines are the value from the perception module. (d) Evolution of position in the image of the plane ($\mathbf{im}_{p}$) (solid) and its desired values (dashed). (e) Evolution of relative distance drone-plane ($d_{dp}$) (solid), its desired values (dashed) and focal length ($f$). (f) Evolution of relative rotation of the plane ($\mathbf{R}_{dp}$) and desired value ($\mathbf{R}_{dp}^*$). Solid lines are actual values and starred lines are desired values.  (h,i,j) Evolution of the position of the drone ($\mathbf{p}_{d}$). (h) Component $x$. (i) Component $y$. (j) Component $z$.}
\label{fig:plane_quanti}
\end{figure*}

Figure \ref{fig:plane_quali} displays one frame for each sequence, providing a visual representation of all the recording perspectives.
For quantitative results, we run the experiment 15 times and show the results in Fig.\ref{fig:plane_quanti}.
 Figure \ref{fig:plane_quanti}-a shows the position of the top of the plane. The ground-truth value and the estimation of the perception module are shown for each 3D point $[x,y,z]$. 
Figure \ref{fig:plane_quanti}-b shows the estimated velocity by the perception module and the ground-truth values for comparison. Analogously, Fig. \ref{fig:plane_quanti}-c depicts the estimated and ground-truth value of the rotation of the plane in the world, $\mathbf{R}_{p}$, converted to roll (not showing, always zero), pitch and yaw notation for simplicity. 

Figure \ref{fig:plane_quanti}-d shows the position of the plane in the image. The upper line depicts the horizontal position in pixels, while the two lower lines represent the vertical position of the top and bottom parts of the plane, respectively.
As mentioned, the higher weight of $J_p$ may cause the other cost terms, like $J_{im}$ take longer to achieve their set-points. 

Figure \ref{fig:plane_quanti}-e shows the drone-plane relative distance (${d}_{dp}$) and the desired values. 
The focal length ($f$) is shown to demonstrate how the camera zoom helps to keep the plane shown in the image following the requested image composition, which changes over time (Fig. \ref{fig:plane_quanti}-d). 
The relative distance is also controlled by $J_p$, so the controller automatically controls the focal length to adjust the composition of the image keeping the drone at the same relative distance to the target, satisfying both $J_p$ and $J_{im}$.

Figure \ref{fig:plane_quanti}-f depicts the relative rotation, $\mathbf{R}_{dp}$, along with its varying desired values in each sequence, sometimes experiencing abrupt changes. After one of those, the drone's recording perspective must change significantly to keep the plane on screen and record it from the desired relative distance.

Finally, Figs. \ref{fig:plane_quanti}-(h,i,j) represent each component of the drone's position ($\mathbf{p}_d$), illustrating the smoothness of the trajectories.

\begin{figure}[!th]
\centering
\begin{tabular}{cc}
    \includegraphics[width=0.46\columnwidth,height=2.3cm]{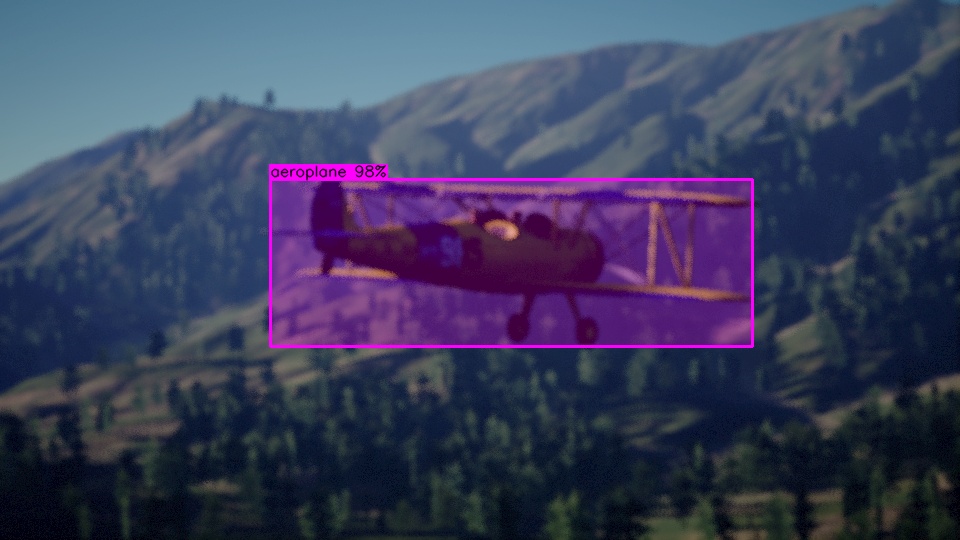}
    &
    \includegraphics[width=0.46\columnwidth,height=2.3cm]{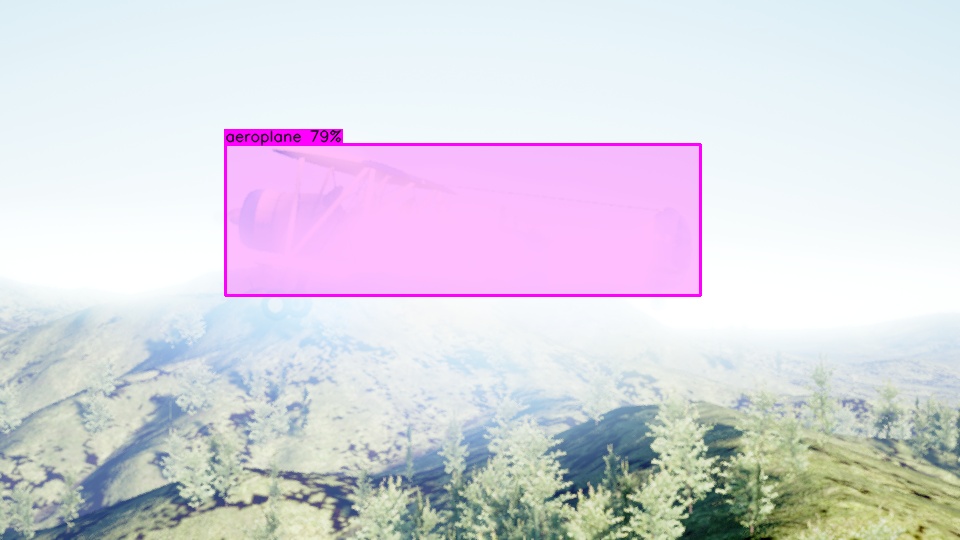}  
    \\\footnotesize (a)  & \footnotesize (b)  \\
     \includegraphics[width=0.46\columnwidth]{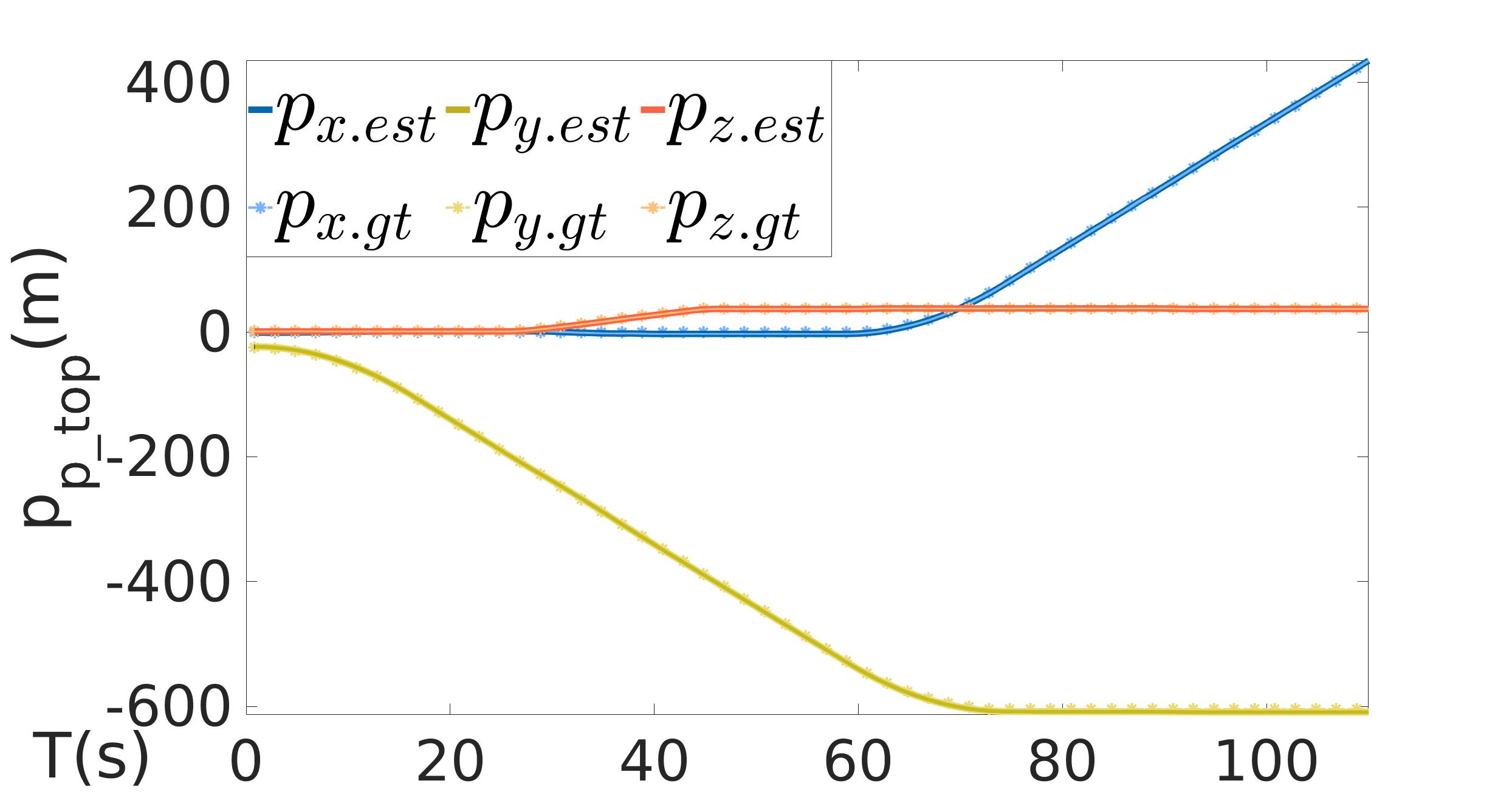}
    & \includegraphics[width=0.46\columnwidth]{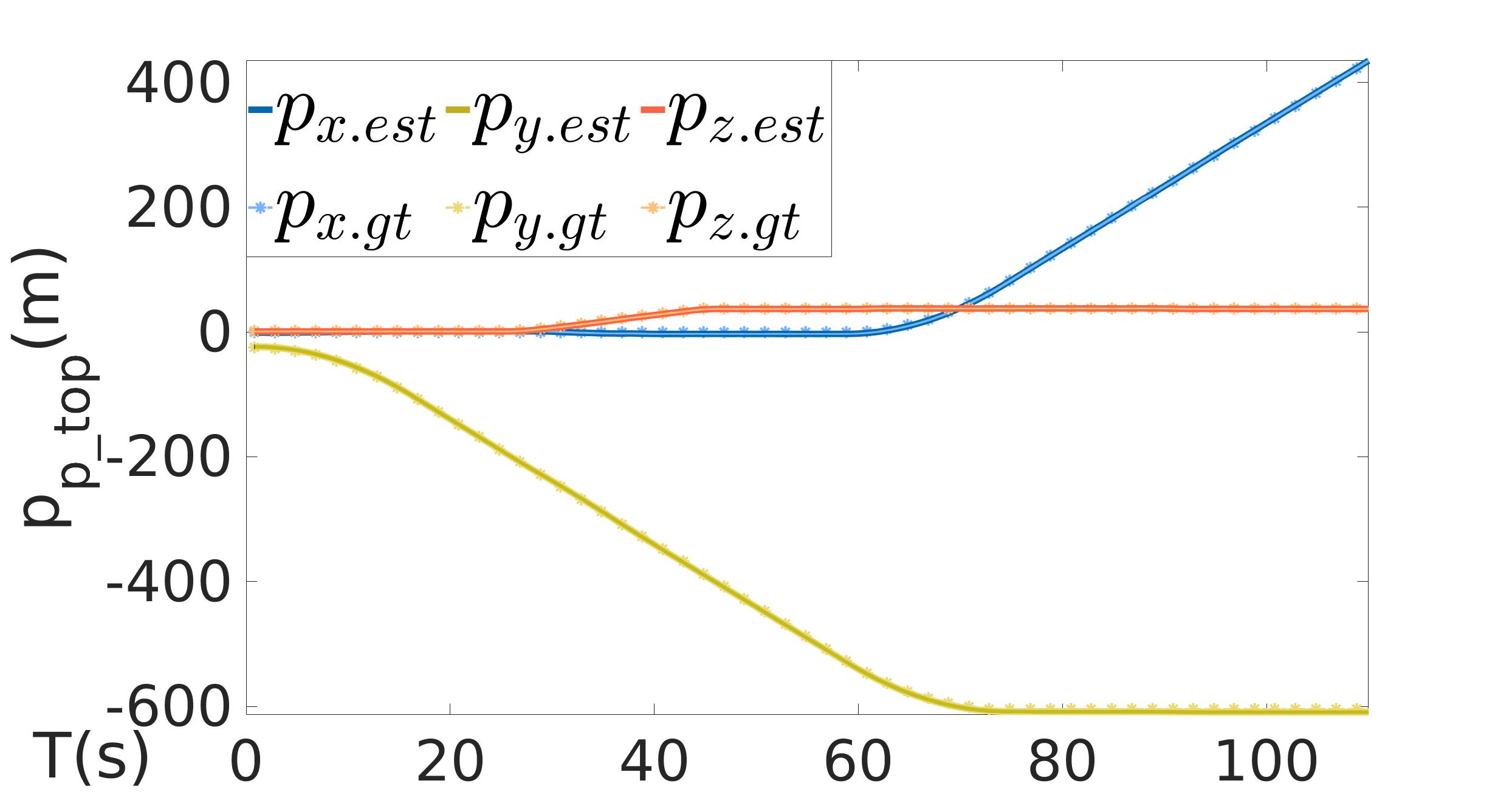}
    
    \\\footnotesize (c)  & \footnotesize (d)  \\
     \includegraphics[width=0.46\columnwidth]{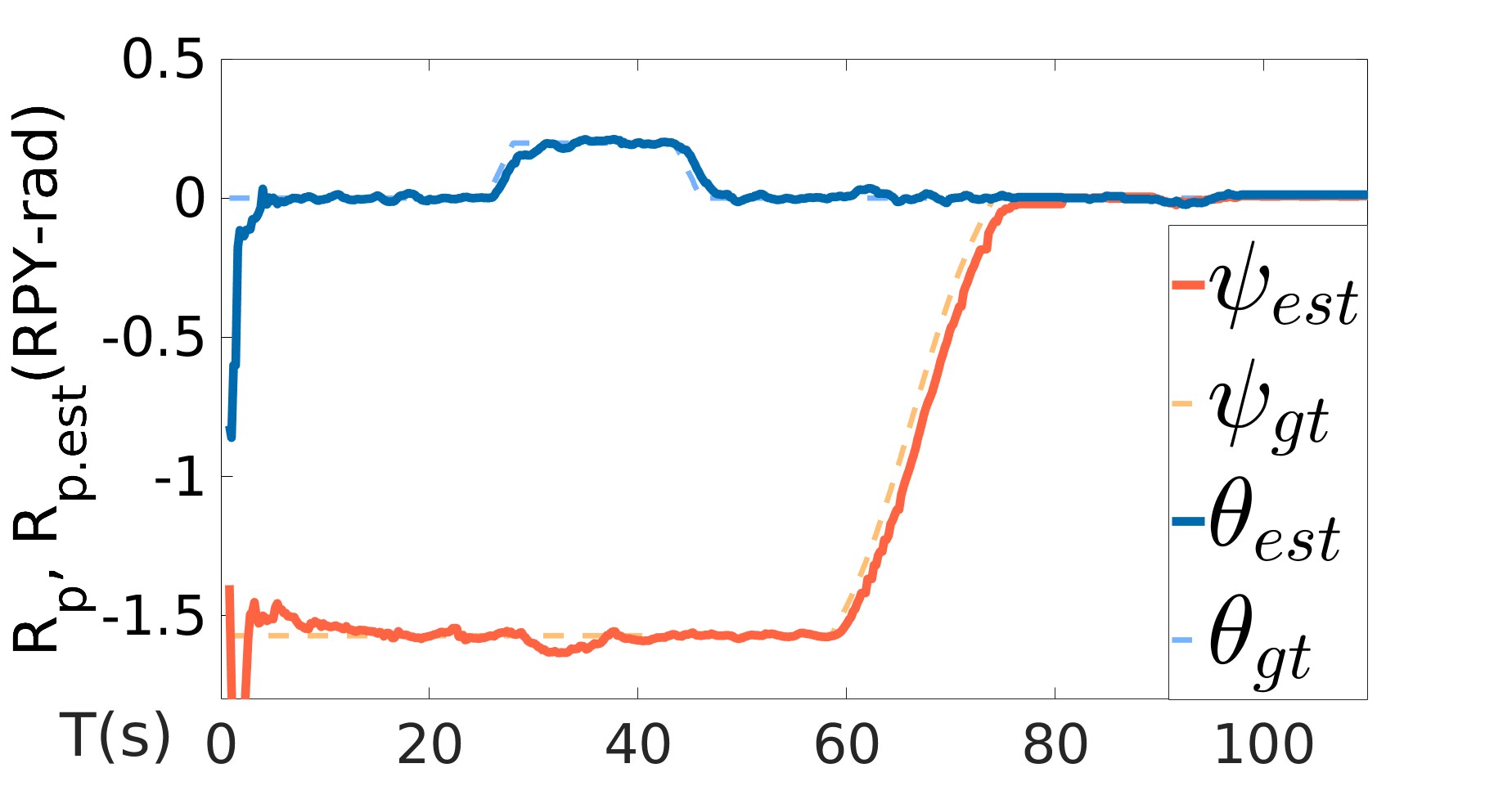}
    & 
     \includegraphics[width=0.46\columnwidth]{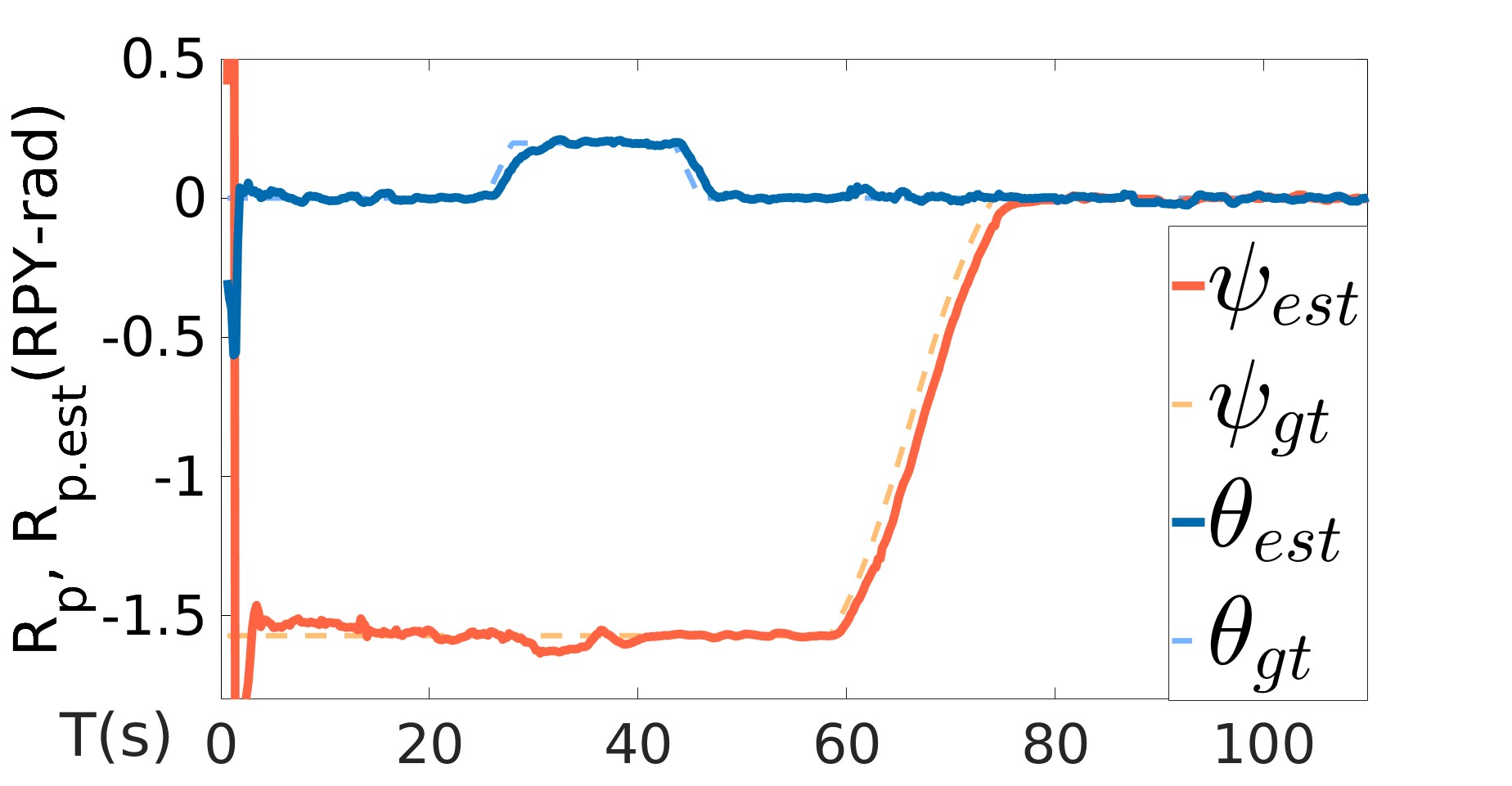}
    \\\footnotesize (e)  & \footnotesize (f) \\
 
\end{tabular}
\caption{\textbf{Experiment 2. Analysis of the perception module}. Recreation of E1 with changes in the scene focus and illumination. First column shows results when the scene is dark and blurry. Second column shows results when the scene is over-illuminated and blurry. (a,b) Frame of the experiment including the detected bounding box. (c,d) Position of the plane ($\mathbf{p}_{p\_top}$). (e,f) Absolute rotation of the plane ($\mathbf{R}_p$) in roll, pitch, and yaw. Solid lines are the estimated values by the perception module and dashed lines are ground-truth. }
\label{fig:perception_test_plane}
\end{figure}

\paragraph{Experiment 2 (E2)- Analysis of the perception module}
In this experiment, we manipulate the scene to show how the perception module is affected by changes in illumination and focus. 
First, we deliberately alter certain camera parameters to intentionally blur the scene throughout the entire experiment. Besides, we conducted illumination changes, running the experiment both in a dark scene and in an over-illuminated scene  (top row of Fig. \ref{fig:perception_test_plane}). 
Figures \ref{fig:perception_test_plane}-(b,e) and Figs. \ref{fig:perception_test_plane}-(c,f) compare the estimation of the position and orientation of the plane and their ground truth, respectively.
In both cases, CineMPC was able to carry out the filming instructions without trouble.

\subsubsection{Scenario B: An Actor standing in the desert}
\label{sec:exp_tuareg}

\paragraph{Goals on this scenario}
The \textit{Dolly Zoom Effect}, or \textit{Vertigo Effect} \cite{liang2020vertigo}, is a well-known kind of shot in cinematography. Important directors used it in awarded films\footnote{See Vertigo (1958) by Alfred Hitchcock or Jaws (1975) by Steven Spielberg}. In this kind of shot, the main target of the scene appears firstly centered vertically in the image. Then, the background suddenly appears to come closer to the viewer while keeping the target centered in the image with the same proportions, transmitting a feeling of vertigo and unreality.

The first goal of this scenario, addressed in Experiment 3, is to assess the control module. This entails showcasing the impact of each cost term in CineMPC on automatically reproducing the shot, with a focus on controlling both the extrinsic and, more importantly, the intrinsic camera parameters.
Moreover, the depth of field is varied over time and controlled automatically. 
In Experiment 4, we introduce changes to the scene to observe how the system can handle various environmental constraints.

\paragraph{Experiment 3 (E3) - Dolly zoom in the desert}
Three targets are positioned in the middle of a desert. 
The primary target is a human actor standing. We intentionally place two cacti far from him -one behind and another in front- to show the effect of the variation of the depth of field.  The experiment is divided into two sequences. In the first sequence, the actor should adhere to the rule of thirds, centered horizontally in the image, while the camera maintains a stable focal length. The \textit{Dolly Zoom Effect} and the changes in the depth of field are conducted in the second sequence. 
The two sequences and their control requirements are described in detail next.

\paragraph*{\textbf{First sequence - Placing targets on the image}}
The goal of this experiment segment is to record a so-named \textit{Cowboy Shot}\cite{thompson2009grammar}, i.e., showing the upper part of the body on the image. The actor is recorded from the front ($J_p$), with the camera maintaining a stable focal length ($J_f$), only controlling the extrinsic parameters.
To achieve the shot, the actor is treated as 'two targets'—the center of his head (nose) and hips. The goal for the actor is to appear centered horizontally, with his nose and hips aligning with the top and bottom vertical thirds.

\begin{figure}[!t]
\centering
\begin{tabular}{cc}
    \includegraphics[width=0.46\columnwidth,height=2.4cm]{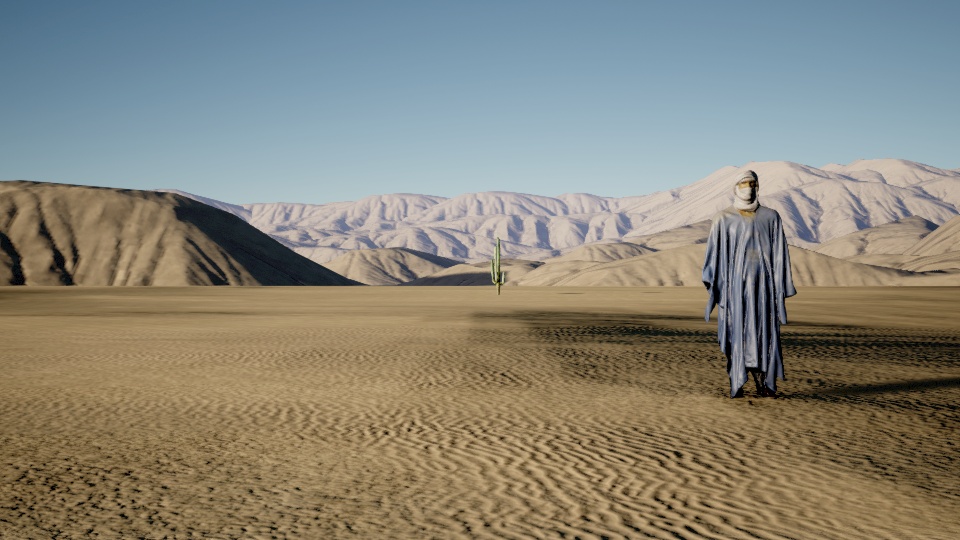}
    & 
    \includegraphics[width=0.46\columnwidth,height=2.4cm]{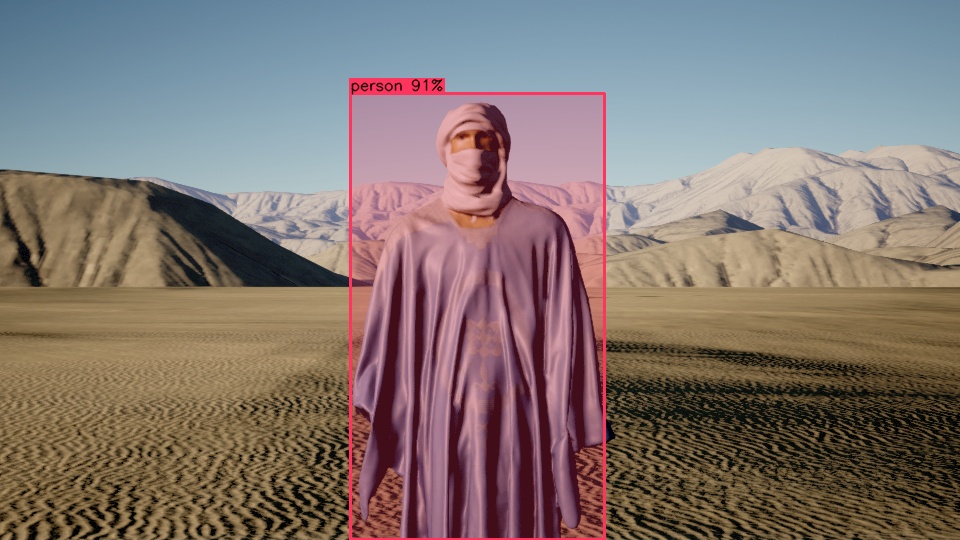}
     \\\footnotesize (a)  & \footnotesize (b) 
     \\
    \includegraphics[width=0.46\columnwidth,height=2.4cm]{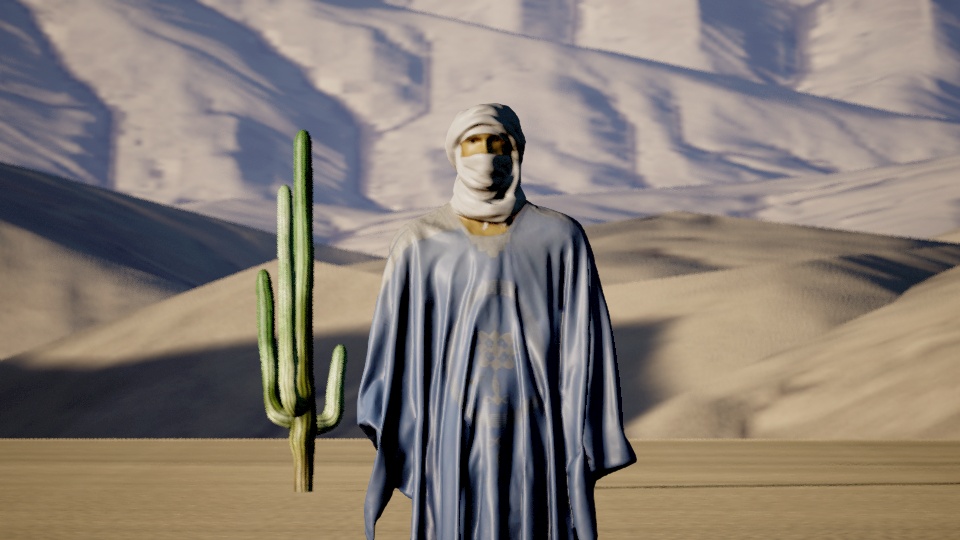}
    & 
    \includegraphics[width=0.46\columnwidth,height=2.4cm]{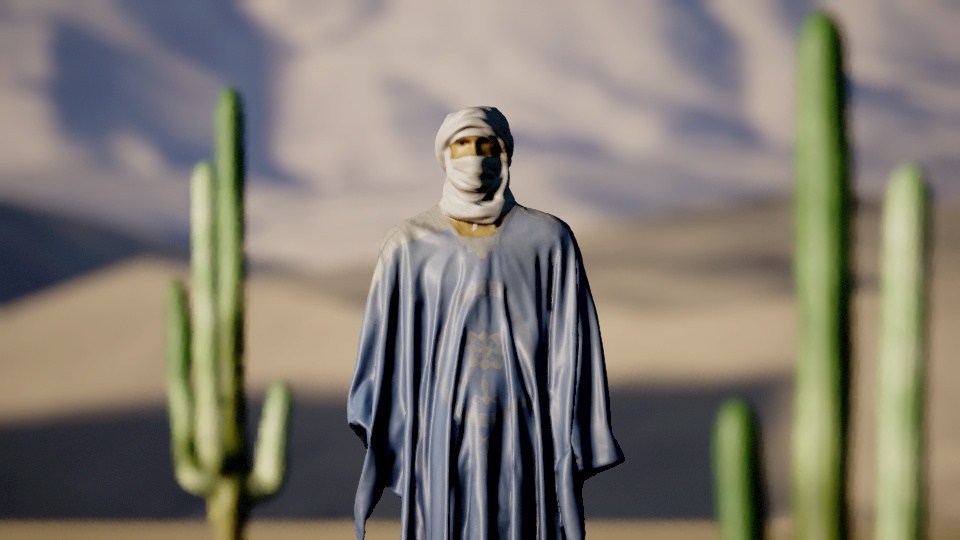}
     \\\footnotesize (c)   & \footnotesize (d) 
\end{tabular}
\caption{\textbf{Experiment 3. Dolly zoom effect - Qualitative results:} (a)  Initial frame of sequence 1. (b) End frame of sequence 1, including the bounding box detected by the perception module. (c) An intermediate frame of sequence 2. (d) End frame of sequence 2.} \label{fig:actor_quali}
\end{figure}

\paragraph*{\textbf{Second sequence - Dolly zoom effect}}
The costs associated with achieving this shot are set as follows:
\begin{itemize}
\item{\bm{$J_{DoF}$}}: This term controls the parts of the scene that appear in focus. We introduce significant variations in the desired depth of field to illustrate its effect. As the background comes closer, the depth of field widens, showing more elements of the scene in focus. The near distance is requested close to the actor, $D_{n,k}^{*} = ({d}_{da,k} - 3) m$, and the far distance, $D_{f,k}^{*}$, ranges from  $({d}_{da,k} + 5)m$ to $({d}_{da,k} + 55)m$.
When the two cacti appear in the image, the foreground and background of the scene get blurrier while keeping the actor sharp. This effect is performed by requesting a narrow depth of field, setting the near and far distances close to the distance to the actor, $D_{n,k}^{*} = ({d}_{da,k} -3)m$, $D_{f,k}^{*} = ({d}_{da,k} +1)m$. Everything that is out of this range, e.g., background, foreground, and cacti, is shown out of focus.
\item{\bm{$J_{im}$}}: This term keeps the actor's proportions in the image untouched from the first sequence.
\item{{\bm{$J_p$}}}: This cost term adjusts $\mathbf{R}_{dt,k}^{*}$ to record the actor from the front. The weight of the relative distance term  ${d}_{da,k}^{*}$ is set to zero so the solver decides about it freely.
\item{\bm{$J_f$}}: The focal length plays a crucial role in executing the \textit{Dolly Zoom Effect}. It is increased linearly at each iteration, starting from 35mm and reaching 450mm, to achieve the desired Dolly Zoom effect.
\end{itemize}

\begin{figure}[!h]
\centering
    \includegraphics[width=0.8\columnwidth]{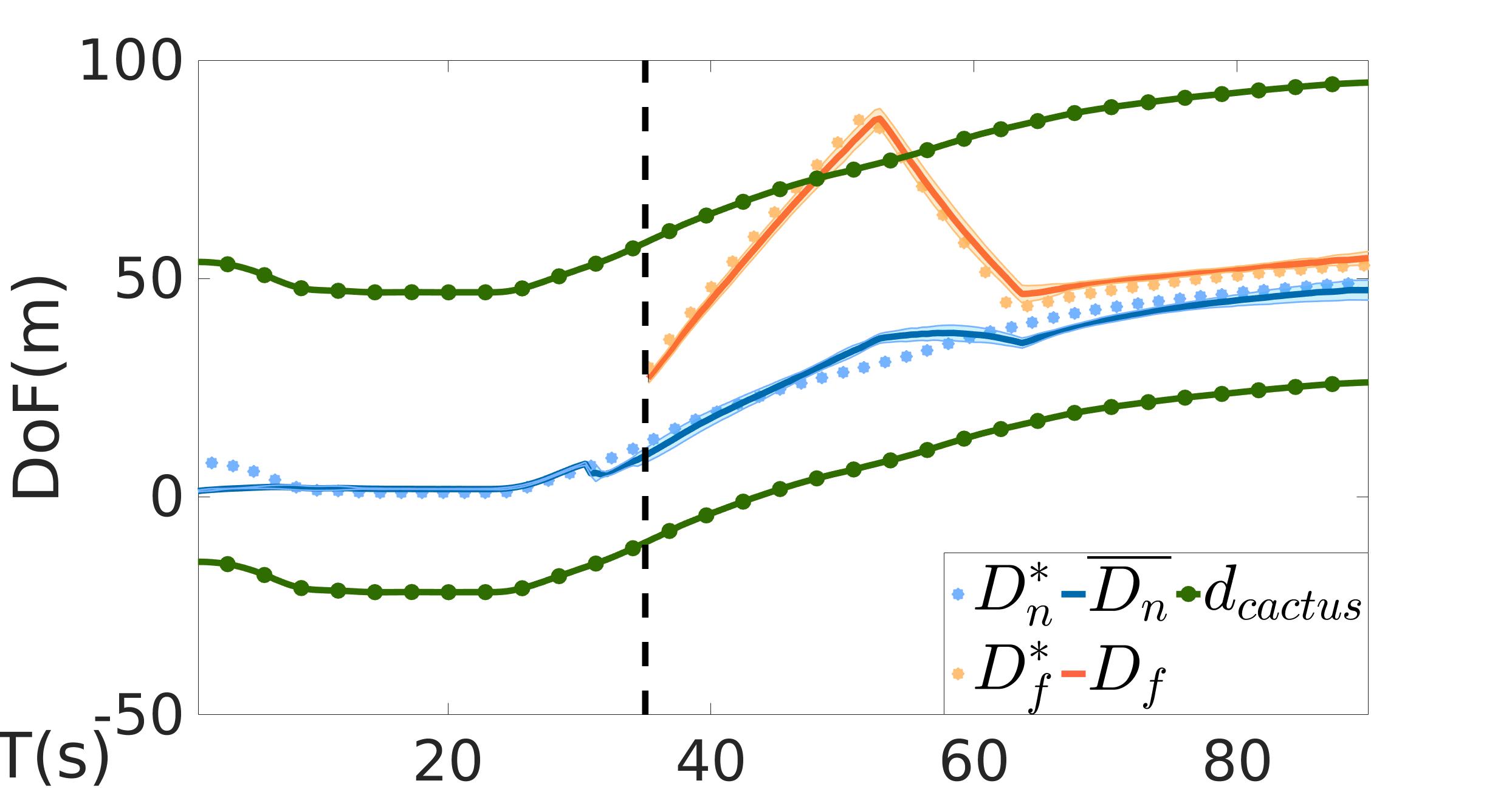}
    \caption{\textbf{Experiment 3. Dolly zoom effect - Quantitative results (DoF):} Evolution of depth of field -near and far distances- and desired values. The solid lines represent actual values and the starred lines depict desired values. The dashed line marks the initiation of the dolly-zoom effect in the video. }
    \label{fig:dolly_dof}
\end{figure}

\begin{figure*}[!t]
\centering
\begin{tabular}{ccc}
    
    \includegraphics[width=0.3\linewidth]{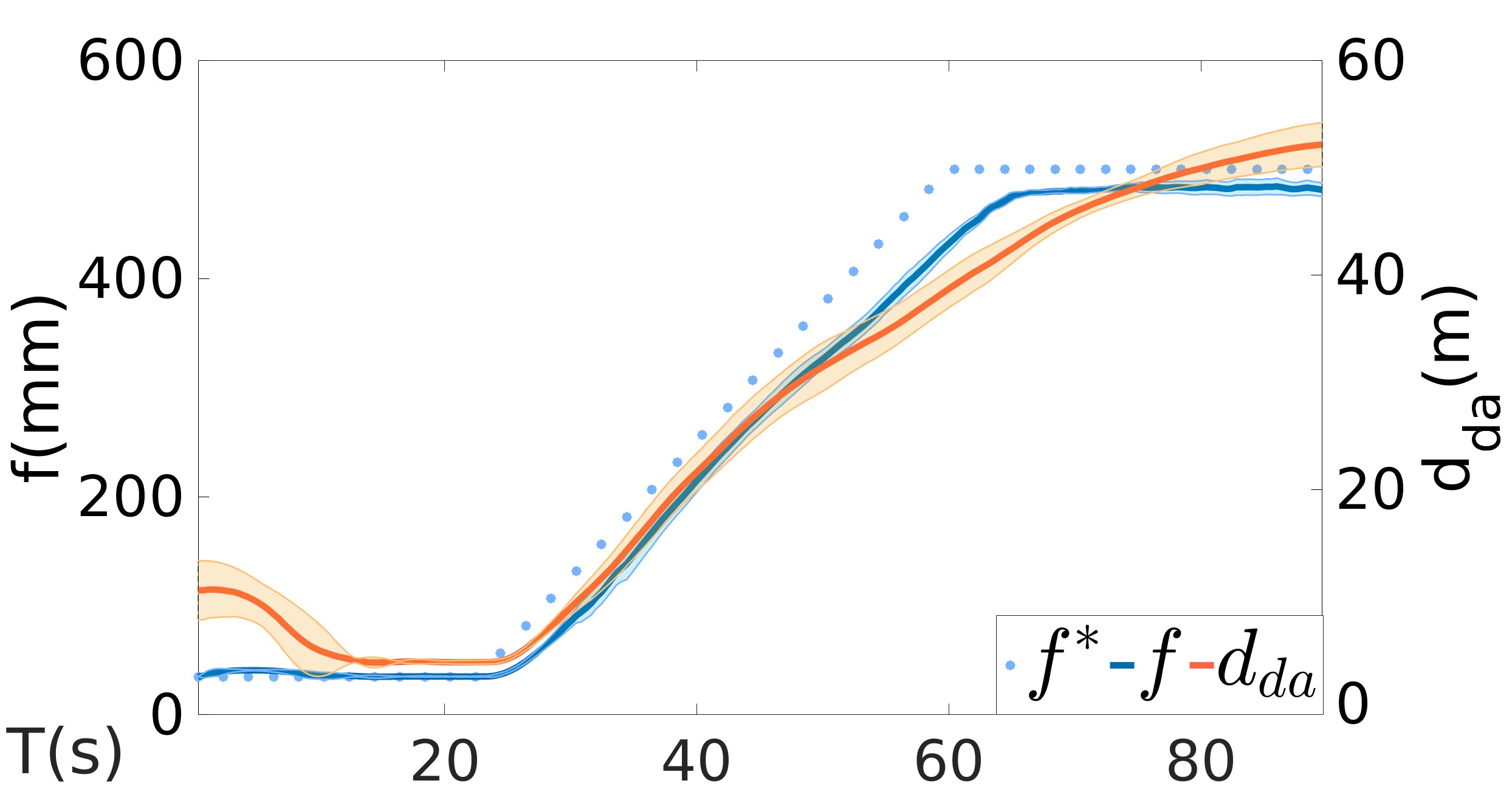} 
     &
     \includegraphics[width=0.3\linewidth]{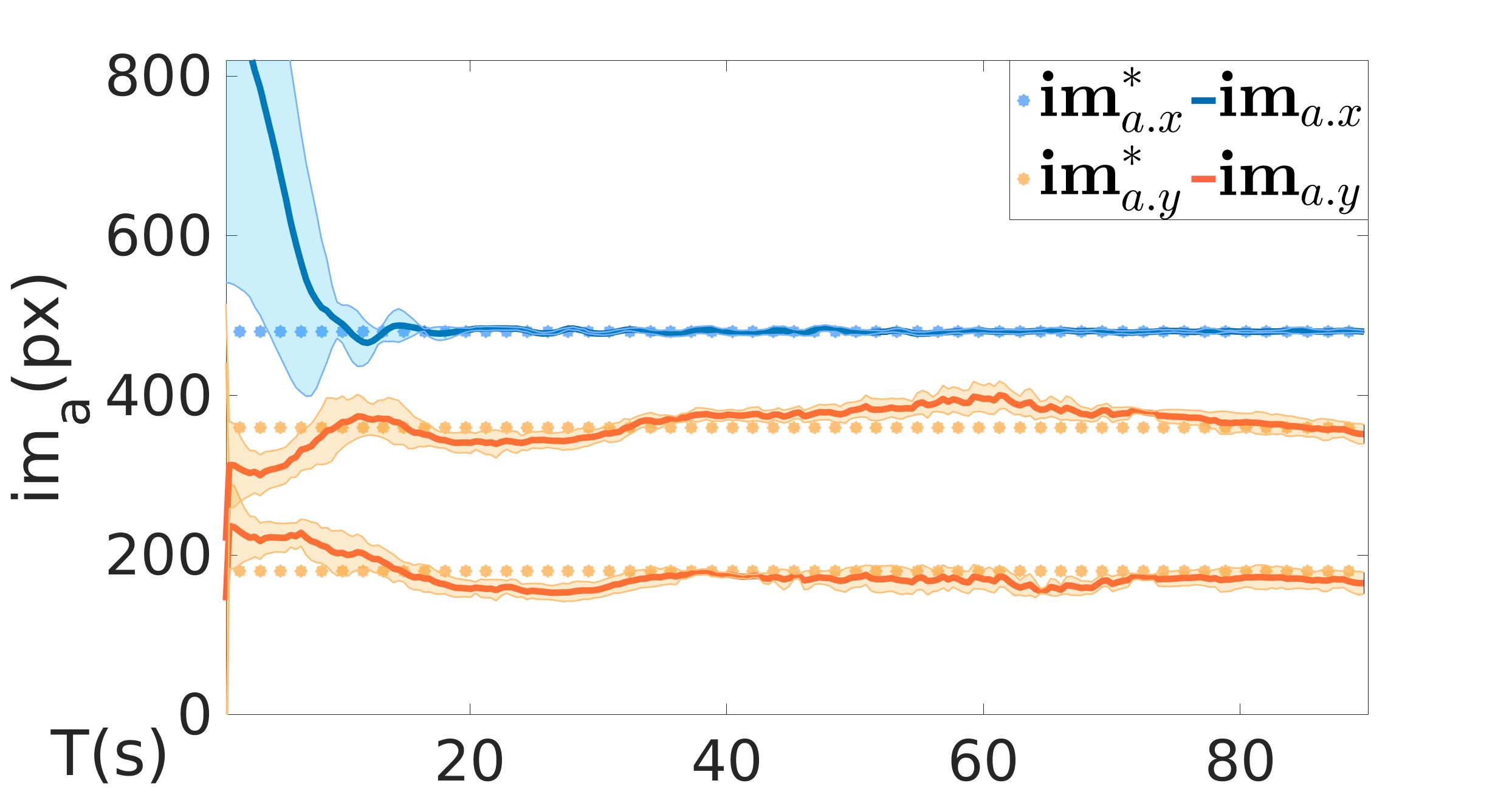}
     &
    \includegraphics[width=0.3\linewidth]{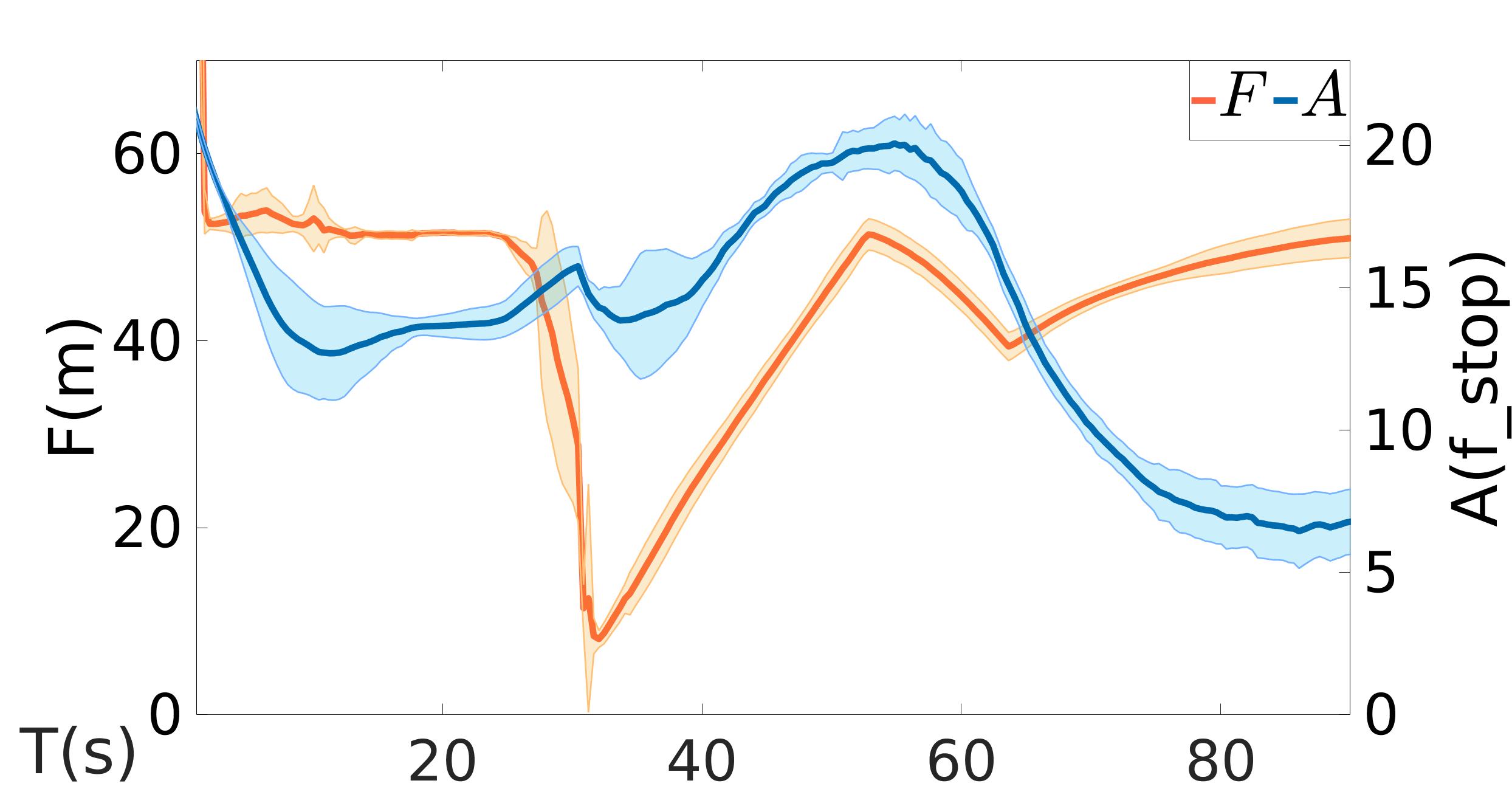}
    \\
    (a) & (b) & (c) \\
\end{tabular}
\caption{\textbf{Experiment 3. Dolly zoom effect - Quantitative results:} The experiment was conducted 15 times from different starting points. Solid lines represent the mean of the plotted value. Lighter areas depict the standard deviation. (a) Evolution of focal length ($f$) and relative distance drone-target ($d_{dt}$). (b) Evolution of the image position of the actor ($\mathbf{im}_a$) and its desired value ($\mathbf{im}_a^*$). Solid lines represent actual values while starred lines depict desired values. The top line is the horizontal pixel and the bottom two lines are the vertical pixels. (c) Evolution of intrinsics (aperture ($A$) and focus distance ($F$)).} 
\label{fig:actor_quanti}
\end{figure*}

Figure~\ref{fig:actor_quali} depicts some frames of the recording for qualitative results, while Fig. \ref{fig:dolly_dof} and Fig. \ref{fig:actor_quanti} show quantitative results of the experiment. Figure \ref{fig:dolly_dof} illustrates the controller's ability to track the desired values of the near ($D_n$) and far distances ($D_f$) of the depth of field. The actor is positioned between those distances, appearing in focus. The other two lines represent the distance to the cacti, which may fall out of the focus range when requested. 

Figure \ref{fig:actor_quanti}-a displays the changes over time in the focal length and the relative distance between the drone and the actor, $d_{da}$. The zoom of the image increases with the focal length. To maintain the image composition, the actor should remain in the same place in the image. Therefore, the drone automatically flies farther away from the actor to compensate for the effect of the higher focal length.
Figure \ref{fig:actor_quanti}-b shows the desired and real position of the actor in the image in pixels, $\mathbf{im}_{a,k}$. At the beginning of the sequence, the drone is positioned far from the actor, resulting in an initial transient period as the drone flies to a position to achieve the desired image composition. Subsequently, it maintains this position throughout the remainder of the experiment despite the zoom.
CineMPC autonomously controls the camera's intrinsic parameters, with the aperture and focus distance depicted in Fig. \ref{fig:actor_quanti}-c. As explained in Sec. \ref{sec_agents}, the aperture plays a vital role in determining the depth of field. In the initial phase of the second sequence, we request a wide depth of field, causing the aperture to reach its maximum value. Consequently, as we request a narrow depth of field towards the end of the sequence, the aperture decreases. Another parameter influencing the depth of field is the focus distance, which the controller sets close to the distance of the actor.

\begin{figure}[!b]
\centering
\begin{tabular}{cc}
    \includegraphics[width=0.46\columnwidth,height=2.4cm]{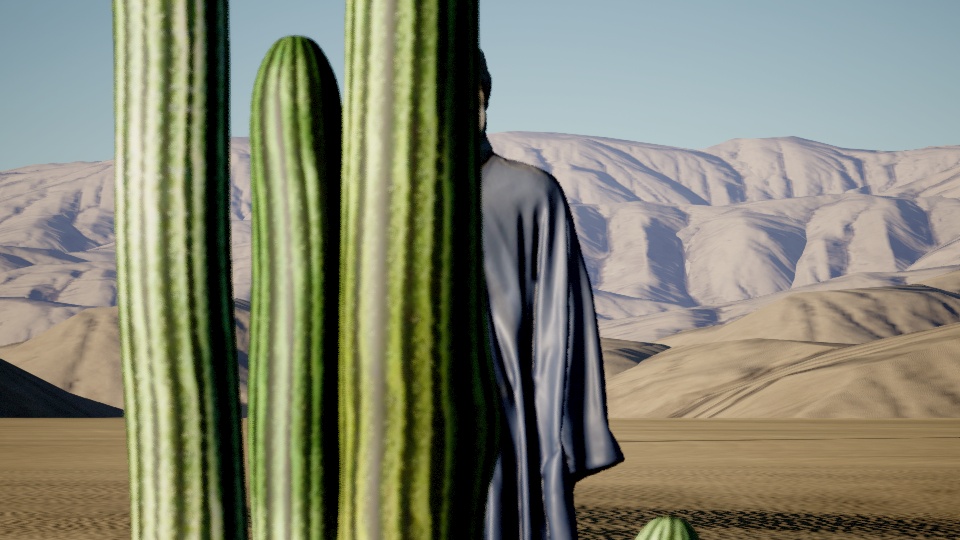}
    & 
    \includegraphics[width=0.46\columnwidth,height=2.4cm]{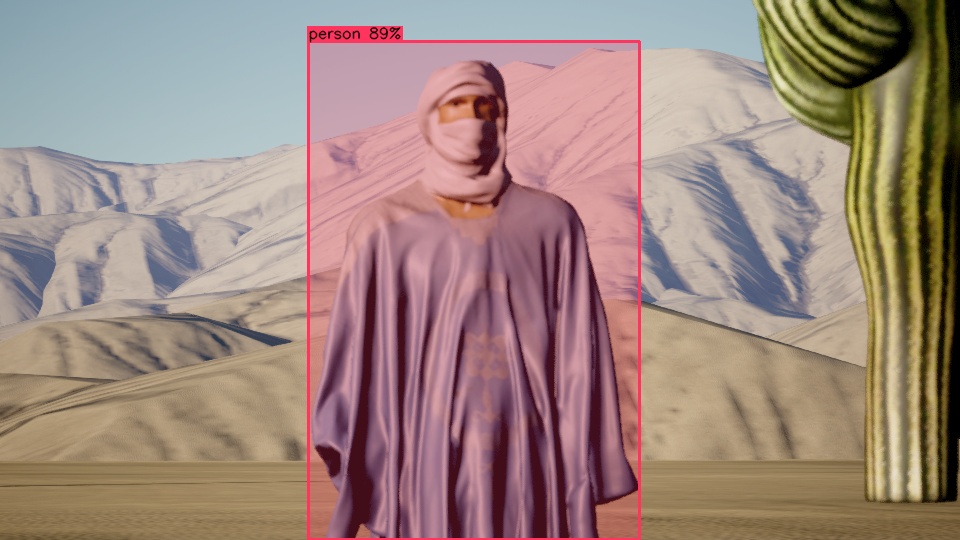} 
    \\\footnotesize (a)  & \footnotesize (b)
    \\
    \includegraphics[width=0.46\columnwidth,height=2.4cm]{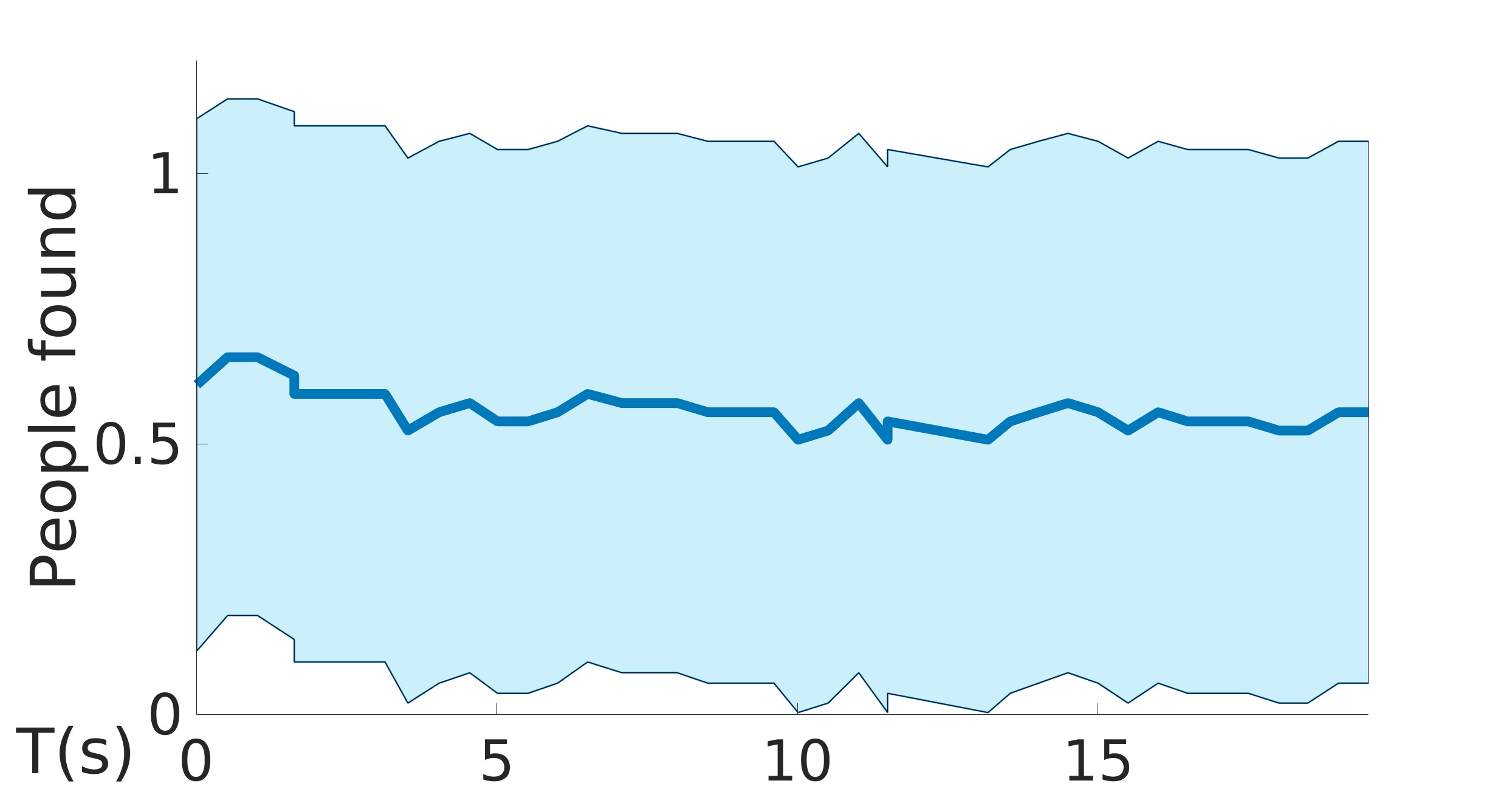}
    & 
    \includegraphics[width=0.46\columnwidth,height=2.4cm]{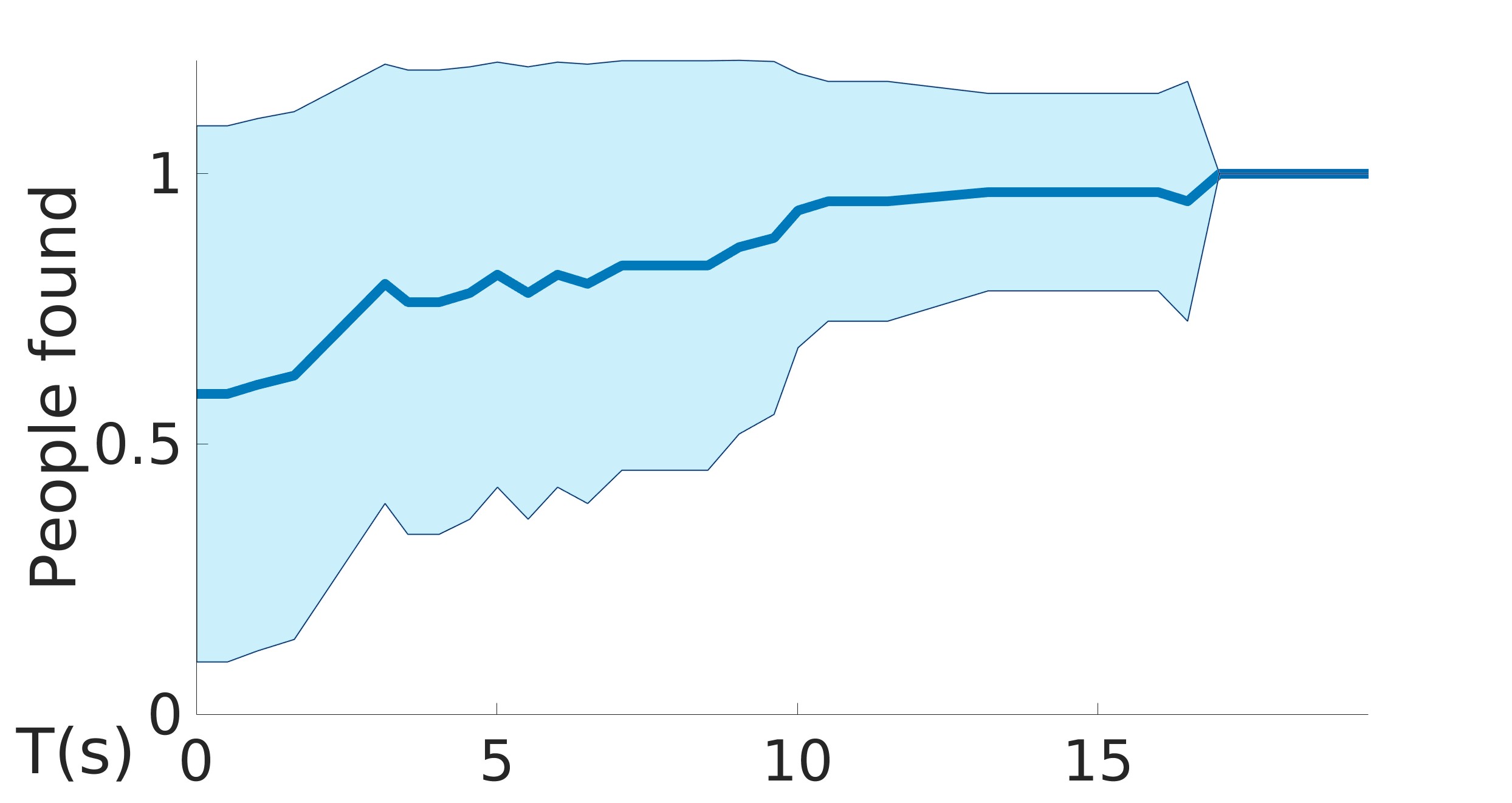} 
    \\\footnotesize (c)  & \footnotesize (d)
    \\
 
\end{tabular}
\caption{\textbf{Experiment 4, Occlusion avoidance:} The left column depicts the experiment without occlusion avoidance constraints, leading to an occlusion of the main target caused by the cactus. This leads the perception module to lose track of the target. On the right, occlusion avoidance constraints prevent the cactus from occluding the target by readjusting the drone trajectory. (a,b)  Final frames captured by the camera drone. (c,d) Plots with average (solid line) and standard deviation (light blue area) of the number of detected persons along the sequence. In (c), existing occlusions cause the perception module to lose track of the actor most of the time. In (d), occlusion avoidance constraints enable the perception module to localize the actor at the end of the sequence.
}
\label{fig:occlusion_test}
\end{figure}

\paragraph{Experiment 4 (E4) - Handling Environmental Constraints}
\label{sec:occlusion}
The goal of this experiment is to demonstrate how CineMPC properly handles different constraints relevant to cinematography and robotics, such as collisions and occlusions.
For this purpose, we add a cactus in a position that affects the recording.
While obstacle detection is not the primary focus of the paper, our modular architecture would facilitate the integration of advanced obstacle/occlusion avoidance techniques~\cite{bonatti2020autonomous, penicka2022minimum, yang2019fast}. Therefore, we assume that the cactus's position is known in advance. However, the target's position is fully determined using the perception module. 

We first test the \textbf{occlusion avoidance constraints}. We request CineMPC to record the actor from the front, following the rule of thirds, for 20 seconds. We conducted the experiment 60 times, randomly varying the initial positions of the drones and cacti (obstacles) within reasonable value intervals that ensure that the actor always remains in the field of view, although the cacti may initially occlude the actor. Figures~\ref{fig:occlusion_test}-a,b depict two frames of this experiment. In Fig.\ref{fig:occlusion_test}-a we do not request CineMPC to satisfy the occlusion avoidance constraints, causing an occlusion. In Fig.\ref{fig:occlusion_test}-b, CineMPC satisfies this hard constraint by redesigning the trajectory of intrinsic and extrinsic parameters to avoid occlusion. This is achieved, even if the primary recording objective (recording from the front) is not entirely met. To quantify the results of this experiment, we evaluate the average number of people detected by the perception module at each time. If the cactus is not occluding the actor, the module localizes one person.  Figures ~\ref{fig:occlusion_test}-c,d show the number of detected people over time without and with occlusion constraints, respectively. 

\begin{figure}[!b]
\centering
\begin{tabular}{cc}
    \includegraphics[width=0.46\columnwidth,height=2.4cm]{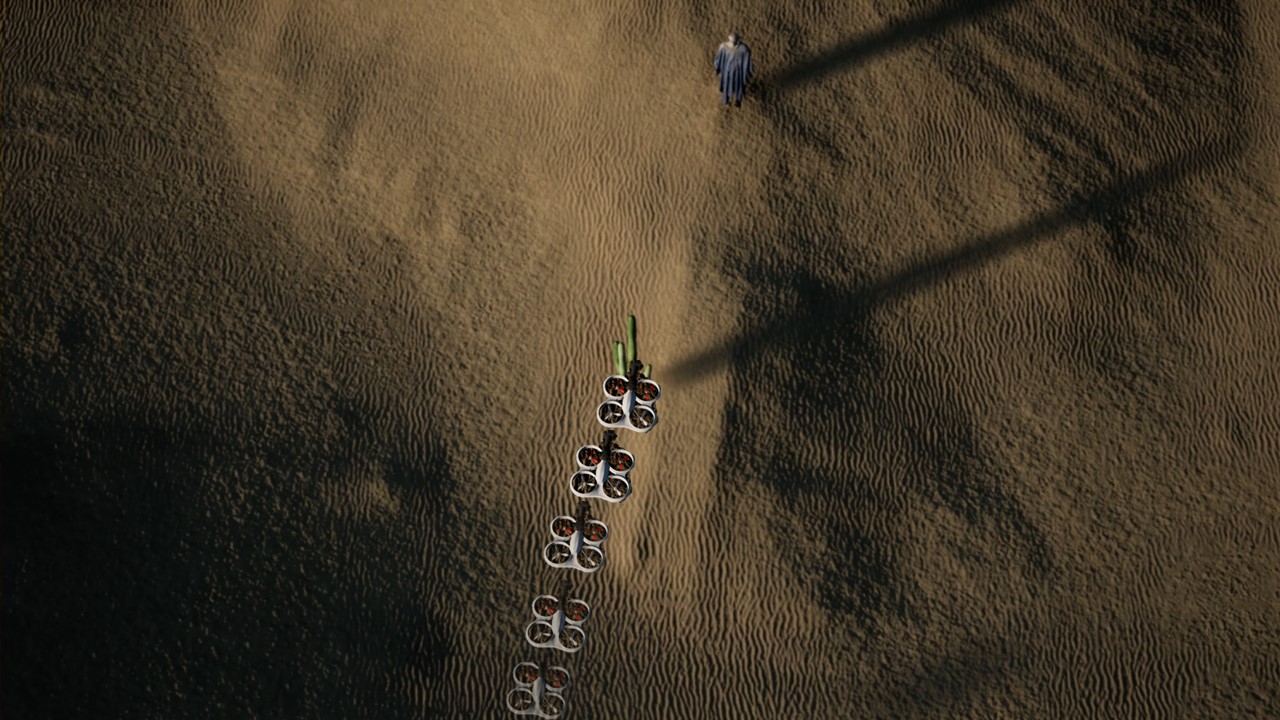}
    & 
    \includegraphics[width=0.46\columnwidth,height=2.4cm]{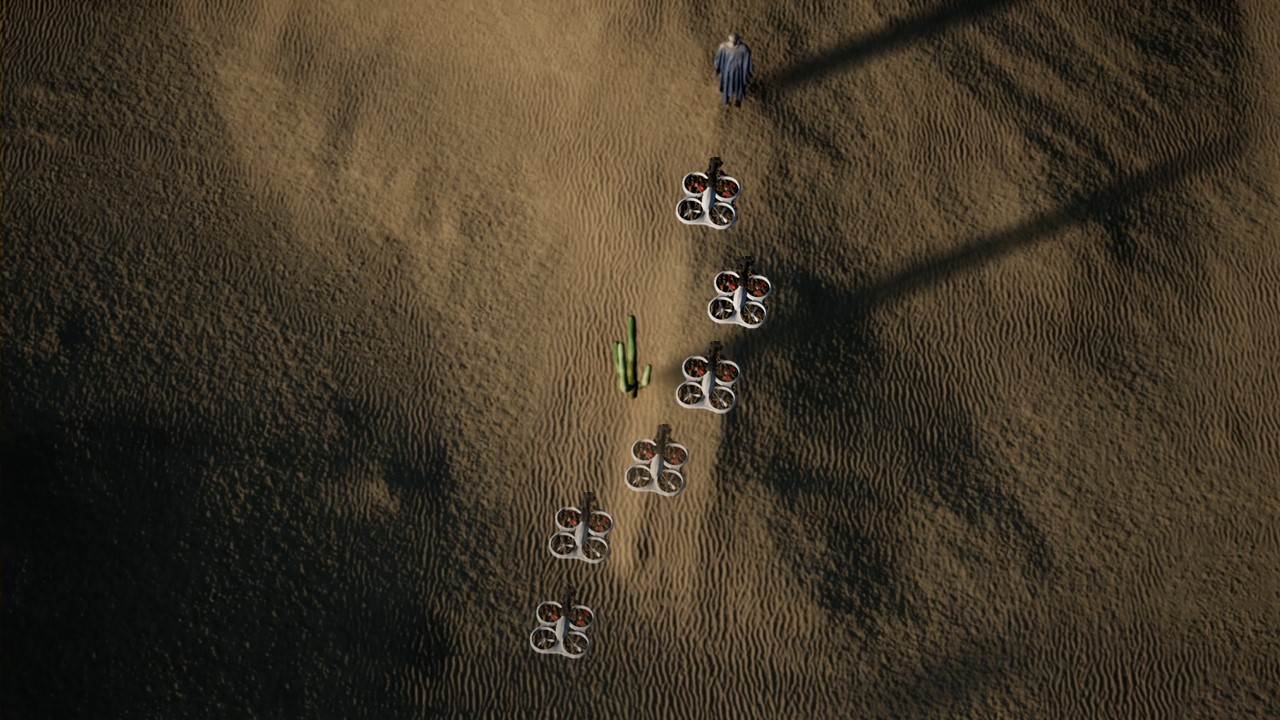} 
    \\\footnotesize (a) & \footnotesize (b) 
\\
\includegraphics[width=0.46\columnwidth,height=2.4cm]{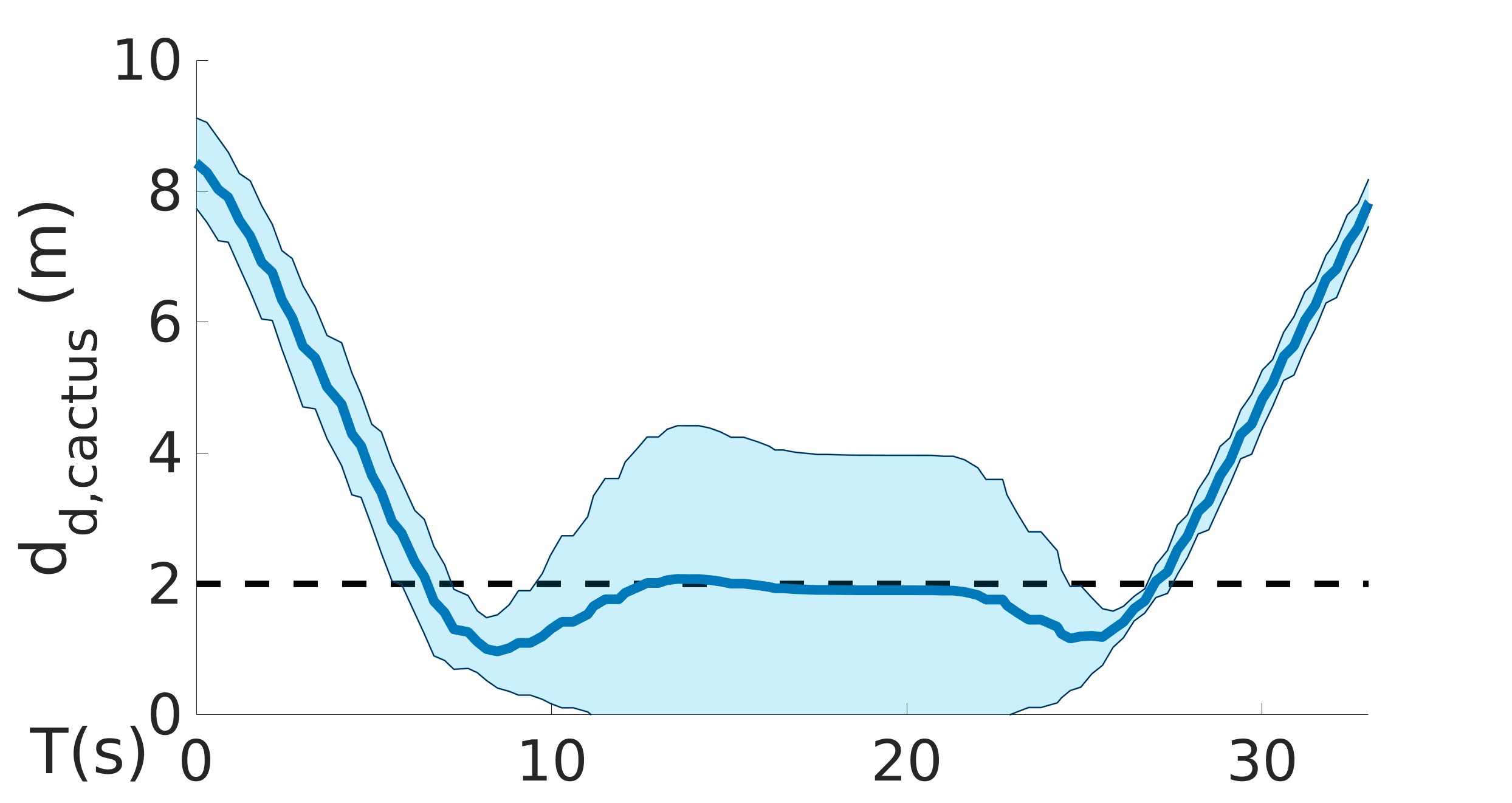}
    & 
    \includegraphics[width=0.46\columnwidth,height=2.4cm]{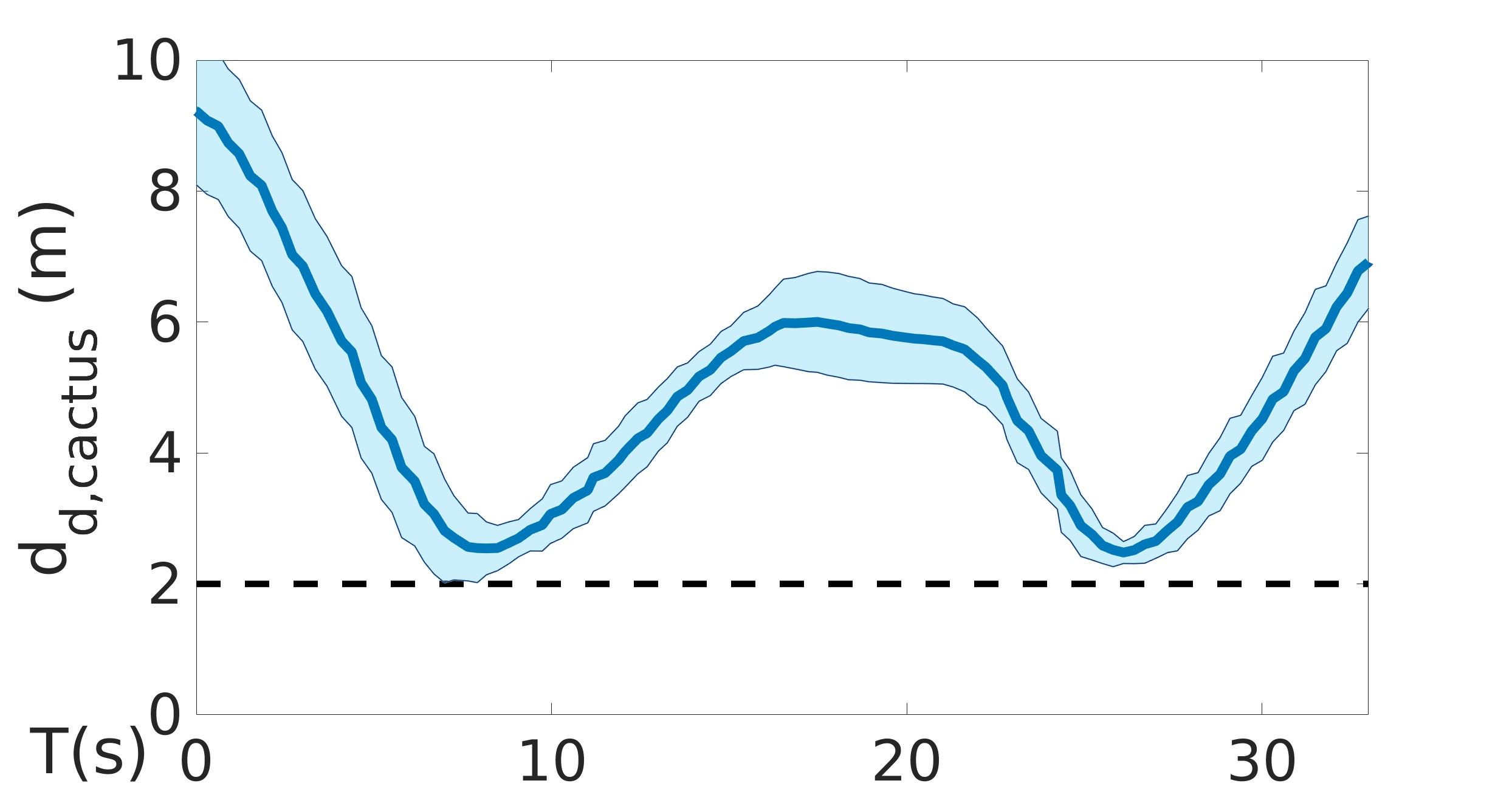} 
    \\\footnotesize (c) & \footnotesize (d)
 
\end{tabular}
\caption{\textbf{Experiment 4, Collision avoidance:} The left column illustrates the experiment conducted without collision avoidance constraints, leading to a collision with a cactus serving as an obstacle. In the right column, collision avoidance constraints prevent the drone from colliding with the cactus by readjusting its trajectory. (a,b) Third-person view of the drone trajectory. (c,d) Plots with average (solid line) and standard deviation (light blue area) of the distance drone-cactus. The dashed black line represents the security distance ($d_{min}= 2m$). In the right column, this security distance is introduced as a constraint in the control module. This causes the drone never to approach the cactus closer than $d_{min}$, avoiding the collision and ensuring proper footage.  }

\label{fig:collision_test}
\end{figure}

Second, we test the \textbf{collision avoidance constraints} using the same scenario. 
We conducted the experiment 10 times, with the recording instructions for each sequence remaining consistent.
Without collision avoidance constraints, the drone impacts the cactus, stopping the recording. When collision constraints are in place (Sec. \ref{sec_constraints}), the drone avoids approaching the cactus closer than $d_{min}$ (2 m. in this experiment) and continues recording. Figure \ref{fig:collision_test}-a displays the trajectory of the drone in the first sequence of the experiment without collision avoidance constraints, resulting in a collision with the cactus. Figure \ref{fig:collision_test}-b shows the trajectory of the drone for the same sequence if the collision constraints are included, altering the trajectory to avoid a collision. Figure \ref{fig:collision_test}-c shows the mean distance between the drone and the cactus over time without collision constraints. The distance goes sometimes below the security distance (dashed black line), causing a collision. Figure \ref{fig:collision_test}-d represents the drone-cactus distance over time when collision constraints are in place.

\subsubsection{User study}
 We conducted a small user study to gauge the potential interest in CineMPC among the general public and evaluate its user-friendliness. We engaged eight non-expert users in a task involving manual control of a drone and camera within a simulation environment. The objective was to capture an image resembling a predefined frame while adhering to specific cinematographic guidelines. The study proceeded in two phases: first, users manually adjusted both the camera's extrinsic and intrinsic parameters using a simple keyboard interface. Then, they utilized the CineMPC user interface to input recording instructions, allowing our solution to position the drone and camera to fulfill these instructions while replicating the reference frame. To quantify the results, we compared the time required for manual adjustments with the time spent configuring all parameters within the CineMPC user interface. Additionally, we assessed the final cost of CineMPC using Equation~\eqref{eq:main_cost} for each approach. Table \ref{table:user} presents the mean and standard deviation (\textit{std}) of the times and costs for each case.

The results show that the users take much less time to understand and set the values in the user interface of CineMPC than controlling a camera manually, producing an image that is worse according to the CineMPC cost function in the last case.
Supplementary material includes a document with all the times and costs for every case as well as the final image of every user when they control the camera manually and the final image with CineMPC given as a reference.

\begin{table}[!t]
\begin{center}
\caption{\textbf{User Study evaluation.} Execution time and CineMPC cost}
\begin{tabular}{| c | c | c | c | c |}
\hline
 & \textit{mean}, time (s) & \textit{std}, time (s) & \textit{mean}, cost & \textit{std}, cost
\\
\hline 
CineMPC & 81 & 15 & 1512 & 754 \\ 
\hline
Manual & 224 & 86 & 6850 & 1205\\
\hline
\end{tabular}
\label{table:user}
\end{center}
\end{table}

%% file: 08B_Real_Experiments_Real.tex
We conducted the real experiments in two different scenarios: outdoors (C) and indoors (D). In the first experiment, we only controlled the camera's intrinsics, enabling us to conduct the experiment outdoors (scenario C). We focus this experiment as an ablation study to show the qualitative differences in the image when controlling the intrinsics versus not controlling them, or the effect of incorporating only some cost terms of Eq.~\eqref{eq:main_cost} in the control problem. 
Simulation experiments about the influence of the intrinsics are available in the conference paper \cite{pueyo2021cinempc}. 
In the second real experiment, we operated the drone indoors within a controlled area with safety nets (scenario D). This setup allows us to fly the drone safely, and to present this experiment as a full test of the cinematographic platform, as we control all the cost terms of Eq.~\eqref{eq:main_cost} and the intrinsics and extrinsics of the camera and drone. 

\subsubsection{Experimental Setup}
\label{sec:exp_setup}
\begin{figure}[!tb]
\centering
\begin{tabular}{cc}
    \includegraphics[width=0.46\columnwidth]{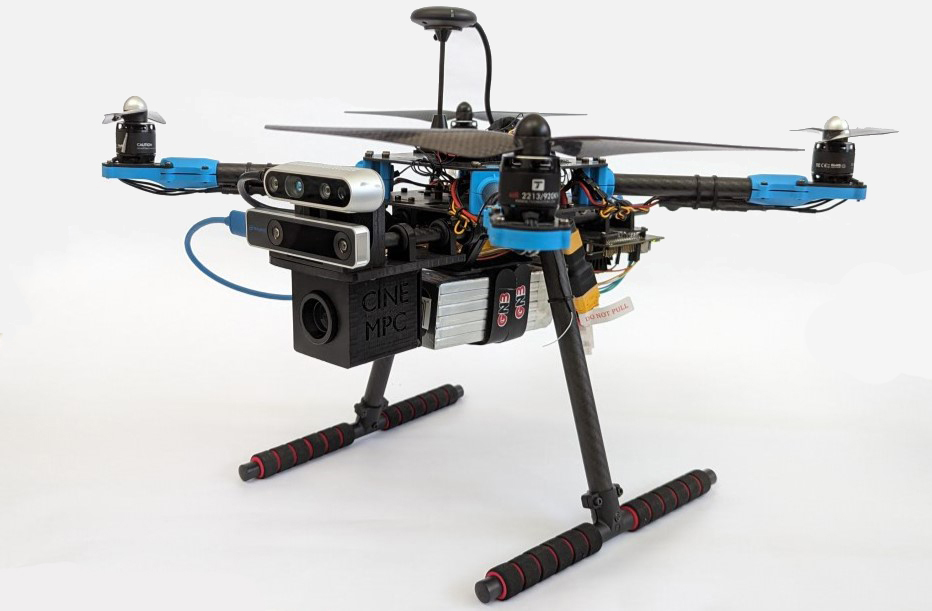}
    & 
    \includegraphics[width=0.46\columnwidth]{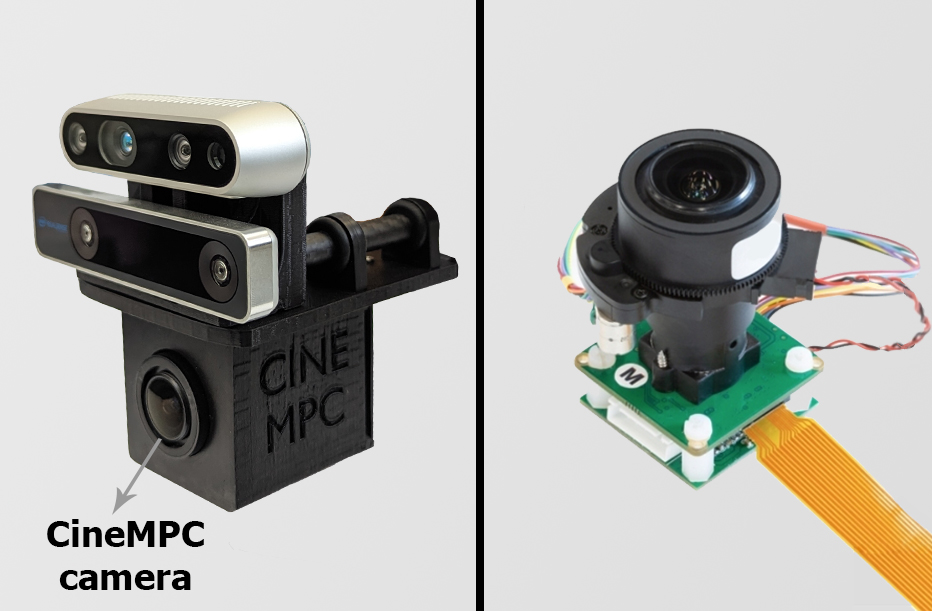} 
    \\\footnotesize (a) & \footnotesize (b) 
 
\end{tabular}
\caption{\textbf{Experimental setup of hardware experiments}. (a) CineMPC drone platform. (b) Left - Close view of the cameras. Top to bottom: RGB-D Camera, Odometry camera, and Cinematographic camera (inside the CineMPC box).  Right - Close view of Cinematographic camera (outside the CineMPC box). Note the inclusion of motors designed to control zoom and focus in real-time. }
\label{fig:real_setup}
\end{figure}

\begin{table}[!h]
\begin{center}
\caption{\textbf{Real experiments}. Camera Parameters of the real camera}
\begin{tabular}{| c | c | c | c | c | c | c | c | c | c | }
\hline
\multicolumn{2}{|c|}{$px$} & \multicolumn{2}{|c|}{$mm$} & $\beta_x$ & $\beta_y$ & $c_u$ & $c_v$& $s$& $c$
\\
\hline 

W & H & W & H & \multirow{ 2}{*}{107.3} & \multirow{ 2}{*}{80.6}   & \multirow{ 2}{*}{337}   & \multirow{ 2}{*}{190}  &\multirow{ 2}{*}{0} &\multirow{ 2}{*}{0.001} 

 \\
 \cline{0-3}

675 & 380 & 6.29 & 4.71 &  &  &  & & &\\ 
\hline
\end{tabular}
\label{table:camera-params-real}
\end{center}
\end{table}
The main hardware components of the drone used in the real setup are:
\begin{itemize}
  \item \textit{Drone frame:} Quadrotor Holybro x500 v2.
  \item \textit{Onboard Computer:} NVIDIA Jetson Nano.
  \item \textit{Flight Controller:} Pixhawk 4.
  \item \textit{Odometry-SLAM Camera:} Realsense T265. Calculates the odometry and the pose of the drone in real-time.
    \item \textit{LiDAR:} Benewake TFMINI-S. Measures altitude.
  \item \textit{RGB-D Camera:}  Realsense D435i. Determines the depth of the targets of the scene. 
  \item \textit{Cine-Camera:} Arducam PTZ 12 MP. This camera module incorporates a lens with two motors to the base camera chip Sony 12MP IMX477. The motors allow the modification of the focus and the zoom of the lens with different steps. We calibrated and transformed these steps into the actual focal length and focus distance. Aperture is not controlled in real experiments.
\end{itemize}
The real setup is depicted in Figure \ref{fig:real_setup}. Figure \ref{fig:real_setup}-a shows the CineMPC drone platform. 
The RGB-D Camera, Odometry Camera, and Cine-Camera are shown on the right side of Fig. \ref{fig:real_setup}-b, top to bottom. The Cine-Camera is embedded into a personalized box for safety and aesthetic reasons. This camera module is shown in detail on the left side of Figure \ref{fig:real_setup}-b. The total weight of the setup is $2.25$ Kg. 

The execution flow of the program is similar to the experiments in simulation. The onboard computer reads and sends the current state of the drone and the camera, along with the images recorded/taken by the cameras, to the desktop computer that is described in Sec.\ref{sec:experimental_desktop}. Since the Jetson Nano lacks a GPU powerful enough to run the entire pipeline, this computer operates the CineMPC framework and computes the next values for the extrinsics and intrinsics to fulfill the experiment's instructions. These commands are then sent back to the onboard computer, which interprets and sends them to the drone and Cine-Camera. 
The computers communicate using ROS. The Flight Controller implements the MAVLink communication protocol, allowing it to communicate with the onboard computer through MAVROS. The software controlling the Cine-Camera is implemented in Python. 
In Experiment 7, we employed an Nvidia Jetson AGX Xavier to demonstrate the capability of running the entire pipeline onboard, eliminating the need for an external desktop computer. However, our current drone setup cannot accommodate this board due to the substantial weight of the Nvidia Xavier together with its development kit, which is 667 g, in stark contrast to the Nano's weight of 241 g.
For the real experiments, we use a sample period of $\Delta_T = 0.5 s.$ and time-horizon of $N = 5$ time-steps.

Table \ref{table:camera-params-real} contains the set of constants described in Sec. \ref{sec_agents} that describe the real cinematographic camera. Table \ref{table:system-constraints-real} describes the lower and upper bounds of the system constraints that are determined by the hardware limits or stabilization requirements of the real experiments. Table \ref{table:weights-real} shows the cost function weights for each sequence of experiments E5 and E6.

\begin{table}[!b]
\begin{center}
\caption{\textbf{Real experiments}. System Constraints.}
\scriptsize{
\begin{tabular}{| c | c | c | c | c | c | c | c | c | c | c | }
\hline
&$\mathbf{p}_d, \mathbf{v}_d$  & $f, F$ & $v_f, v_F$\\
\hline
E5-min & $0, 0$& $5, 0.4$  & $-0.5, -0.5$ \\
\hline 
E5-max & $0, 0$& $10, 7$ & $+0.5, +0.5$ \\
\hline 
E6-min & $-0.15, -0.15$& $5, 0.4$ & $-0.1, -0.1$ \\
\hline 
E6-max & $+0.15, +0.15$& $10, 13$ & $+0.1, +0.1$ \\
\hline
\end{tabular}
}
\label{table:system-constraints-real}
\end{center}
\end{table}

\begin{table}[!tb]
\begin{center}
\caption{\textbf{Real experiments}. Weights of cost function terms.}
\begin{tabular}{| c | c | c | c | c | c | c | c |}
\hline
   & $w_{D_{n}}$ & ${w_{im.x}}$ & ${w_{im.y}}$ & ${w_{d}}$ & ${w_{f}}$\\ \hline
E5 & 300 & 1 & 0 & 0 &  1 \\
\hline 
E6 & 5 &  2  & 2 & 400  & 400 \\

 \hline
\end{tabular}
\label{table:weights-real}
\end{center}
\end{table}

\begin{figure}[!bh]
\centering
\begin{tabular}{cc}
    \includegraphics[width=0.46\columnwidth,height=2.4cm]{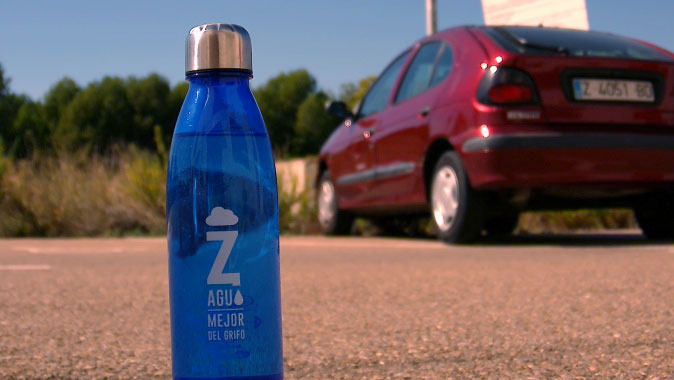}
    & 
    \includegraphics[width=0.46\columnwidth,height=2.4cm]{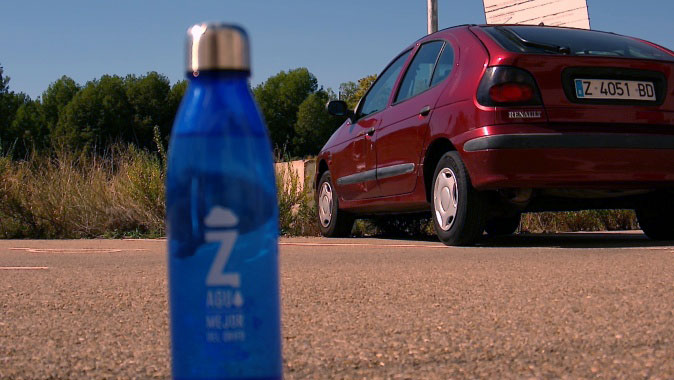}
    \\\footnotesize (a) & \footnotesize (b) 
\\
    \includegraphics[width=0.46\columnwidth,height=2.4cm]{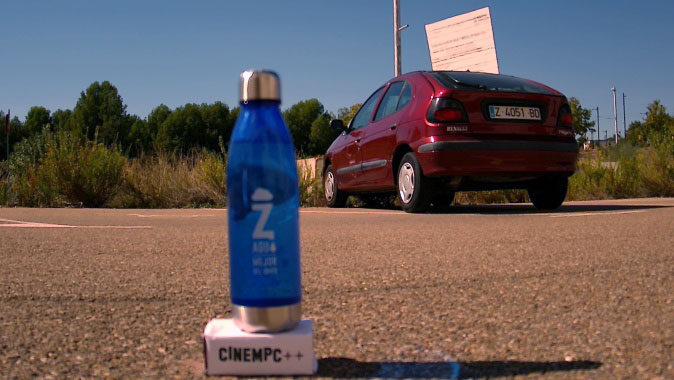}
    & 
    \includegraphics[width=0.46\columnwidth,height=2.4cm]{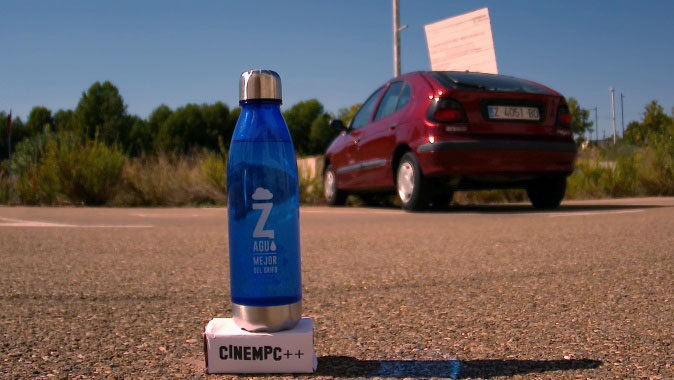}
    \\\footnotesize (c) & \footnotesize (d) 
\\
    
    \includegraphics[width=0.46\columnwidth,height=2.4cm]
    {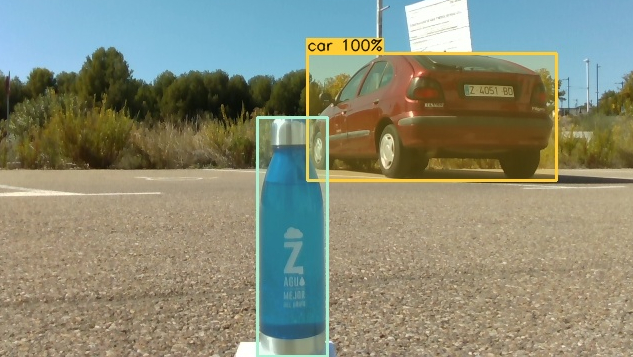}
    & 
    \includegraphics[width=0.46\columnwidth,height=2.4cm]{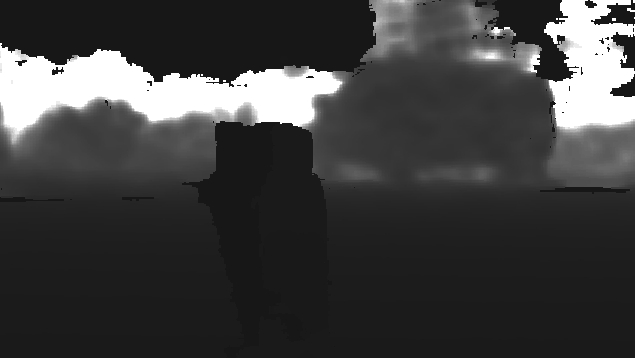} 
    \\\footnotesize (e) & \footnotesize (f)
    \\
    
    \includegraphics[width=0.46\columnwidth,height=2.4cm]{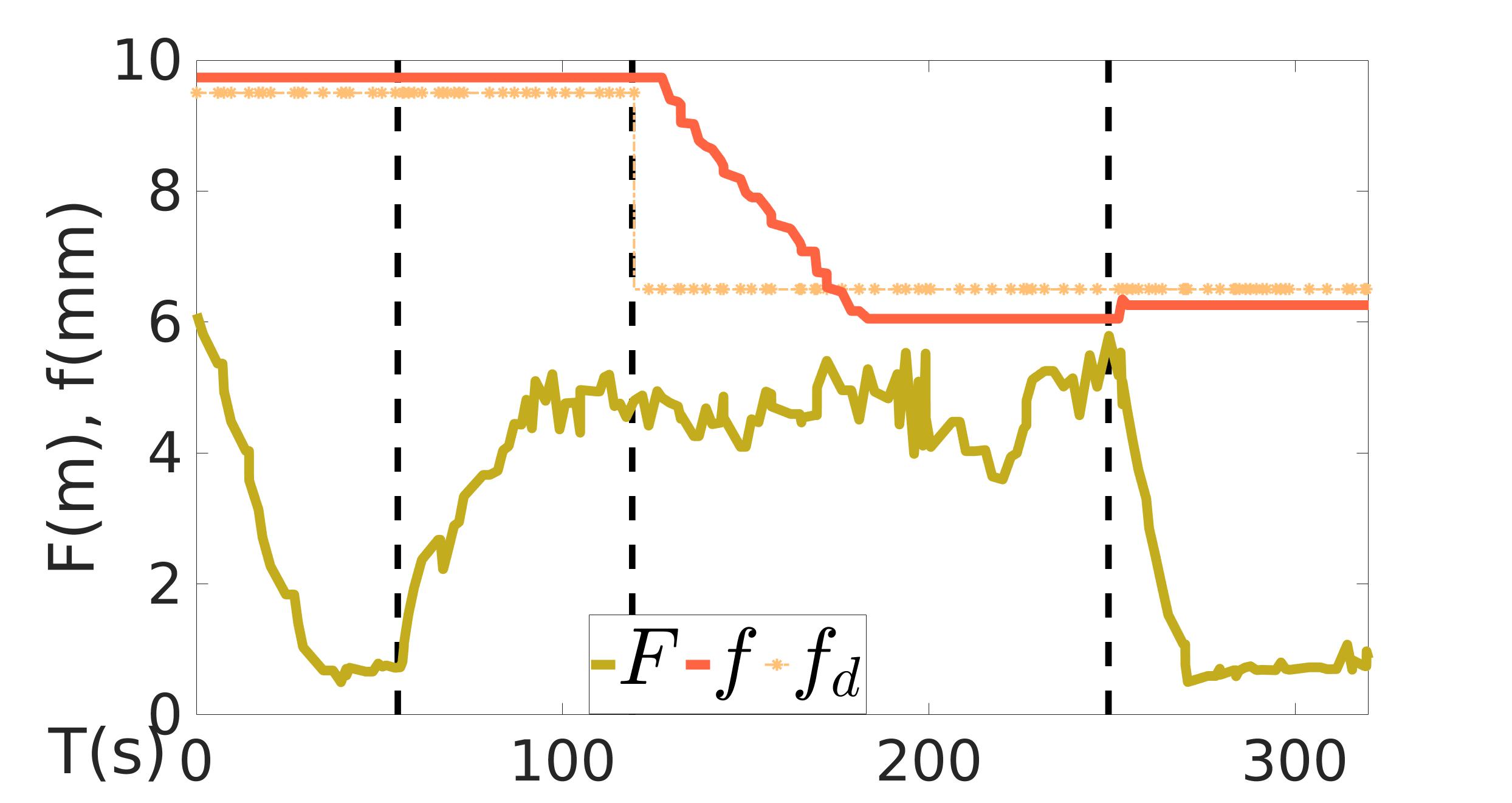} &
\includegraphics[width=0.46\columnwidth,height=2.4cm]{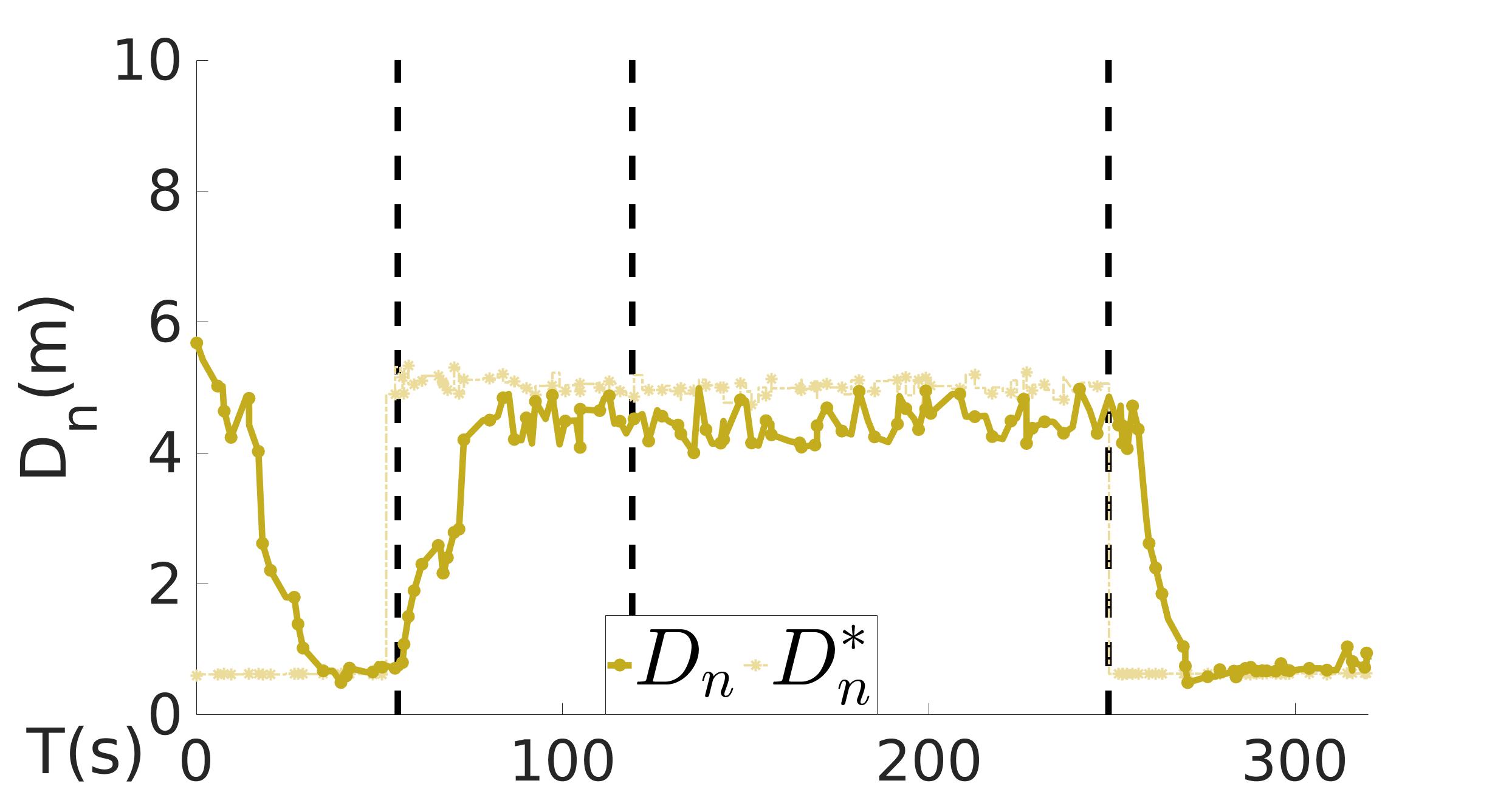}

    \\\footnotesize (g) & \footnotesize (h)
    \\
    
    \includegraphics[width=0.46\columnwidth,height=2.4cm]{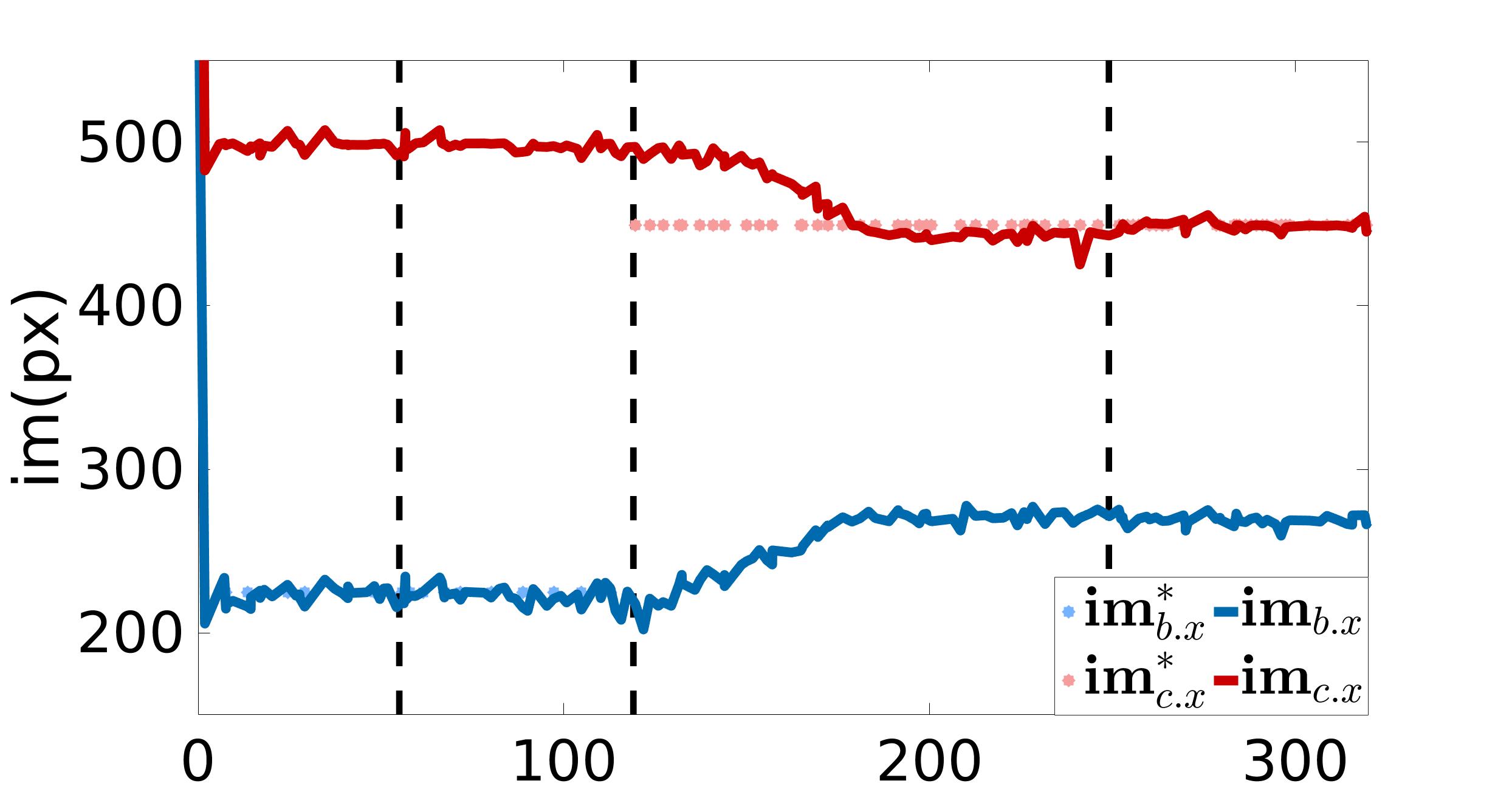} 

    & 
\includegraphics[width=0.46\columnwidth,height=2.4cm]{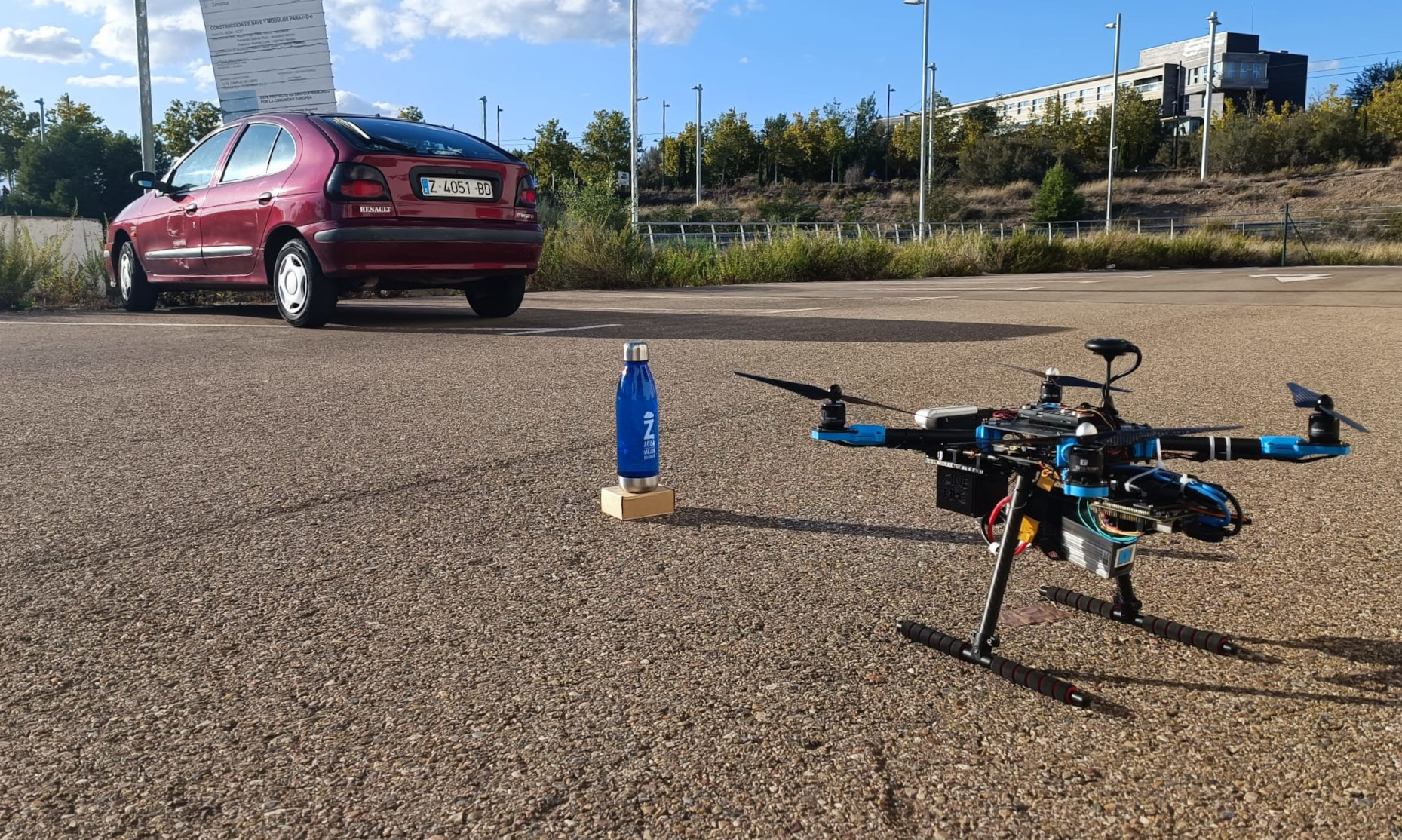}

    \\\footnotesize (i) & \footnotesize (j)
 
\end{tabular}
\caption{\textbf{Experiment 5: Controlling the Depth of Field and image position of two
targets using only intrinsics}. Lighter lines represent desired values and darker lines represent actual values. Dashed black lines represent a change of sequence. (a,b,c,d) Frames captured by the cinematographic camera in each sequence (e) RGB output captured by the RGB-D camera, showcasing bounding boxes around objects detected by the perception module during the experiment. (f) Depth output captured by the RGB-D camera.
 (g) Intrinsics of the camera. The orange line is the focal length ($f_d$ and $f$) and the yellow line focus distance ($F$). (h) Near distance ($D_n$) of the Depth of Field. (i) Controlled vertical pixel for each target ($\mathbf{im}_{t.x}$).  The blue line represents the image position of the bottle (controlled in Seqs. 1 and 2) and the red line represents the car (controlled in Seqs. 3 and 4). (j) Third-person view of the setup of this experiment, which remains static throughout the entire execution.}

\label{fig:real_intrinsics}
\end{figure}

\subsubsection{Experiment 5 (E5): Controlling the Depth of Field and image position of two targets using only intrinsics} 
This experiment is presented as an ablation study, where we exclusively control the cost terms of Eq. \eqref{eq:main_cost} where the intrinsics play a vital role. This approach showcases how controlling these parameters can positively impact the final recording. In this experiment, the position of the drone remains static, as shown in Figure \ref{fig:real_intrinsics}-h.
Therefore, the position in the image and focus of the elements of the scene in real-time are uniquely determined using the focus distance and focal length of the camera. The aperture is not controllable in this camera.  Consequently, we just control $J_{im}$,  $J_{DoF}$ and $J_{f}$ from Eq. \eqref{eq:main_cost} to meet the goals. CineMPC uses the RGB-D camera to calculate the position of the targets and adjust the desired depth of field. First, the bounding box of the targets is detected from the RGB image (Fig. \ref{fig:real_intrinsics}-e). Next, the distance to the object is calculated obtaining the depth of these bounding boxes from the depth image (Fig.\ref{fig:real_intrinsics}-f). Finally, we adjust the requested near distance ($D_n^*$) to focus on the distance to the target. The focus distance of the camera is changed in real-time to satisfy this requirement.  
The experiment is divided into four sequences, each with two targets to position in the image and focus on: a bottle in the foreground and a car in the background. In the first sequence, the bottle should be placed in the left horizontal third, and the position of the car is not controlled. The bottle should be shown in focus, and the car out of focus. In the second sequence, the focus is the only parameter that changes, focusing on the car and showing the bottle out of focus. The third sequence maintains the focus but the position of the car should match the right horizontal third of the image. The last sequence maintains the positions of the elements but changes the focus, showing the bottle in focus and the car and background out of focus.  

The results of this experiment are presented in Fig. \ref{fig:real_intrinsics}. Figure \ref{fig:real_intrinsics}-g illustrates the values of the intrinsics (focus distance and desired and actual focal length). 
Figure \ref{fig:real_intrinsics}-h displays the evolution of the near distance along with the desired values.  Figure \ref{fig:real_intrinsics}-i depicts the desired and actual horizontal image positions of both targets. These quantitative experiments demonstrate the direct influence of the focus distance and the focal length on achieving the desired depth of field and position of the targets in the image, respectively.

A representative frame of each sequence is shown in Figs \ref{fig:real_intrinsics}-a,b,c,d as qualitative results.
Notice how the target's sizes vary in the frames due to alterations in the focal length (e.g. Fig.\ref{fig:real_intrinsics}-a,b vs Fig.\ref{fig:real_intrinsics}-c,d), and how the focus of the scene changes due to the effect of the focus distance (e.g. car focused in Fig.\ref{fig:real_intrinsics}-a,c vs not focused in  Fig.\ref{fig:real_intrinsics}-b,d).
All the cinematographic effects performed in this experiment, and represented in these frames, differ significantly from the output of the fixed camera shown in Fig. \ref{fig:real_intrinsics}-e where the image remains static during the whole experiment. This ablation study helps to demonstrate the importance of controlling the intrinsics of the camera in the final image result.

\subsubsection{Experiment 6 (E6): Full platform test with a flying Cinematographic Drone - Assessing performance and integration}
The goal of this experiment is to demonstrate how CineMPC can control all cost terms by adjusting the extrinsics and intrinsics of a real drone and cinematographic camera to meet various cinematographic objectives. Each cost term in Equation \ref{eq:main_cost} plays a crucial role in filming this recording:

\begin{itemize}
\item{\bm{$J_{DoF}$}}: Manages the focus distance ($F$) to ensure that the main target is always shown in focus, respecting the requested depth of field ($D_n^*$).
\item{\bm{$J_{im}$}}: Controls the position of the drone ($\mathbf{p}_d$) and the focal length ($f$), placing the target in the requested image position ($\mathbf{im}_t^*$). In this experiment, the associated weight ($w_{im}$) is proportionally higher to highlight its effect.
\item{\bm{$J_p$}}: Places the drone at a desired recording distance ($d_{dt}^*$) by controlling its position $\mathbf{p}_d$.
\item{\bm{$J_f$}}: Controls the focal length ($f_d$) avoiding aggressive zoom variations.
\end{itemize}

To reduce complexity, and due to the hardware limitations, we do not control the orientation of the camera and the aperture of the camera in this experiment.

The experiment is divided into two sequences. In the first sequence, the chair in the scene should align with the left vertical third and the top of the chair should match the top horizontal third according to the rule of thirds. Moreover, the drone should be placed 4.5 meters away from the target, keeping a stable focal length of 7 mm.
In the second sequence, the chair should match the right vertical third, and the top of the chair the horizontal center of the image, while the drone flies at 3.5 meters from the target, keeping a focal length of 5.4 mm. In both sequences, the control of the depth of field ensures the chair appears in focus in the image.

\begin{figure}[!tb]
\centering
\begin{tabular}{cc}
    \includegraphics[width=0.46\columnwidth,height=2.4cm]{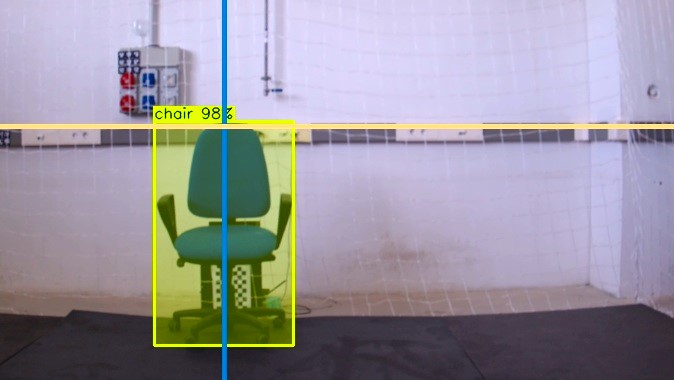}
    & 
    \includegraphics[width=0.46\columnwidth,height=2.4cm]{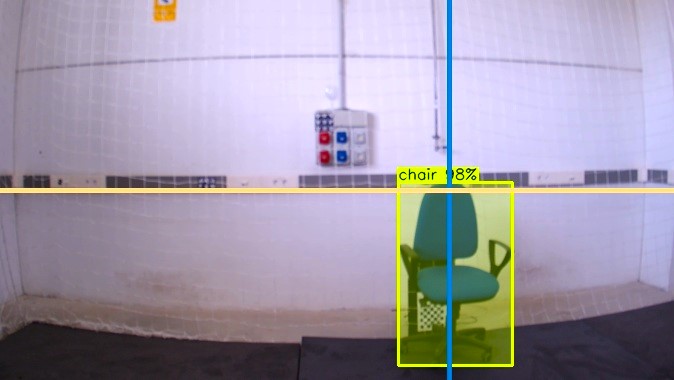}
    \\\footnotesize (a) & \footnotesize (b) 
\\
\includegraphics[width=0.46\columnwidth,height=2.4cm]{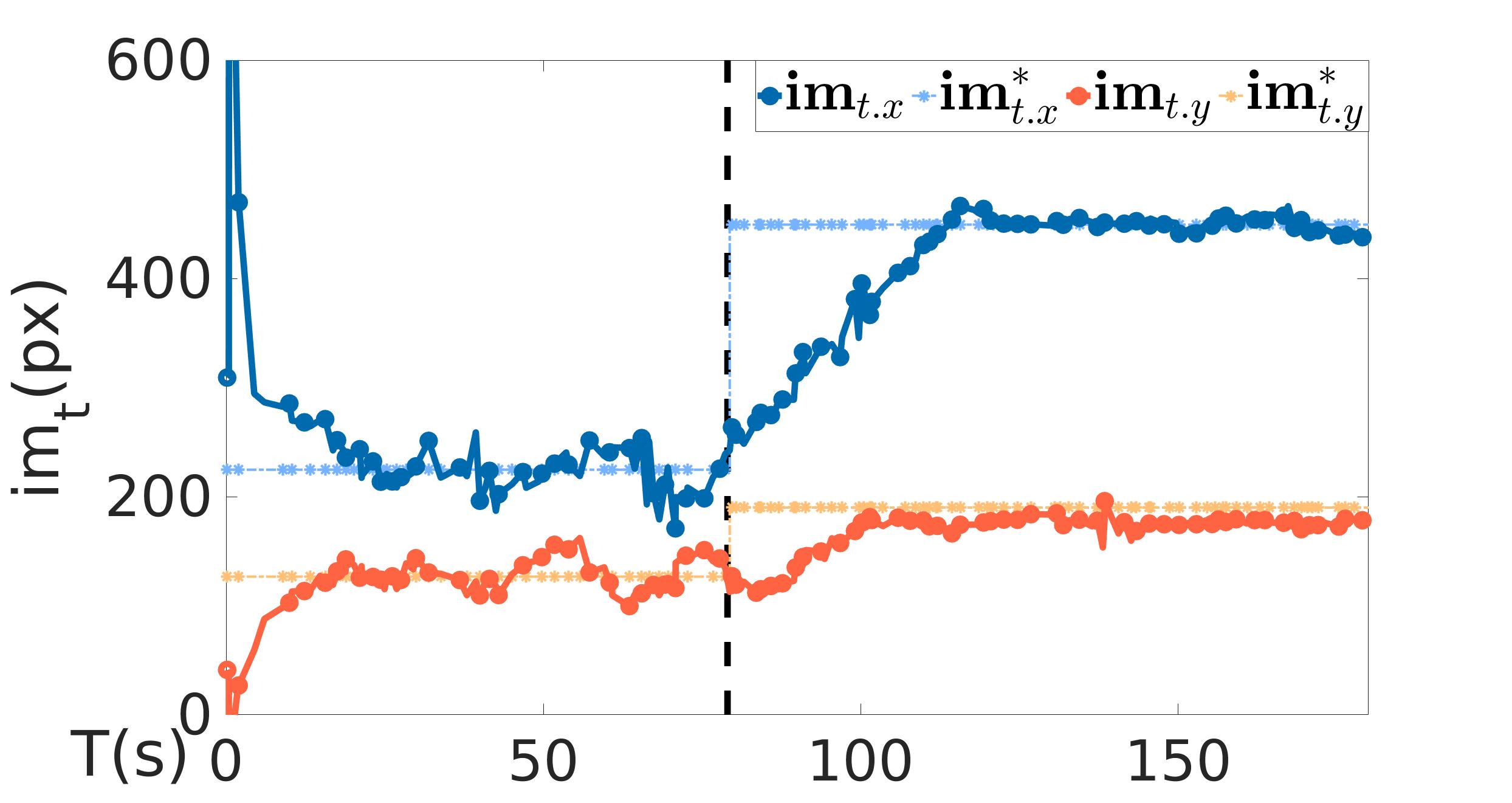}
    & 
    \includegraphics[width=0.46\columnwidth,height=2.4cm]{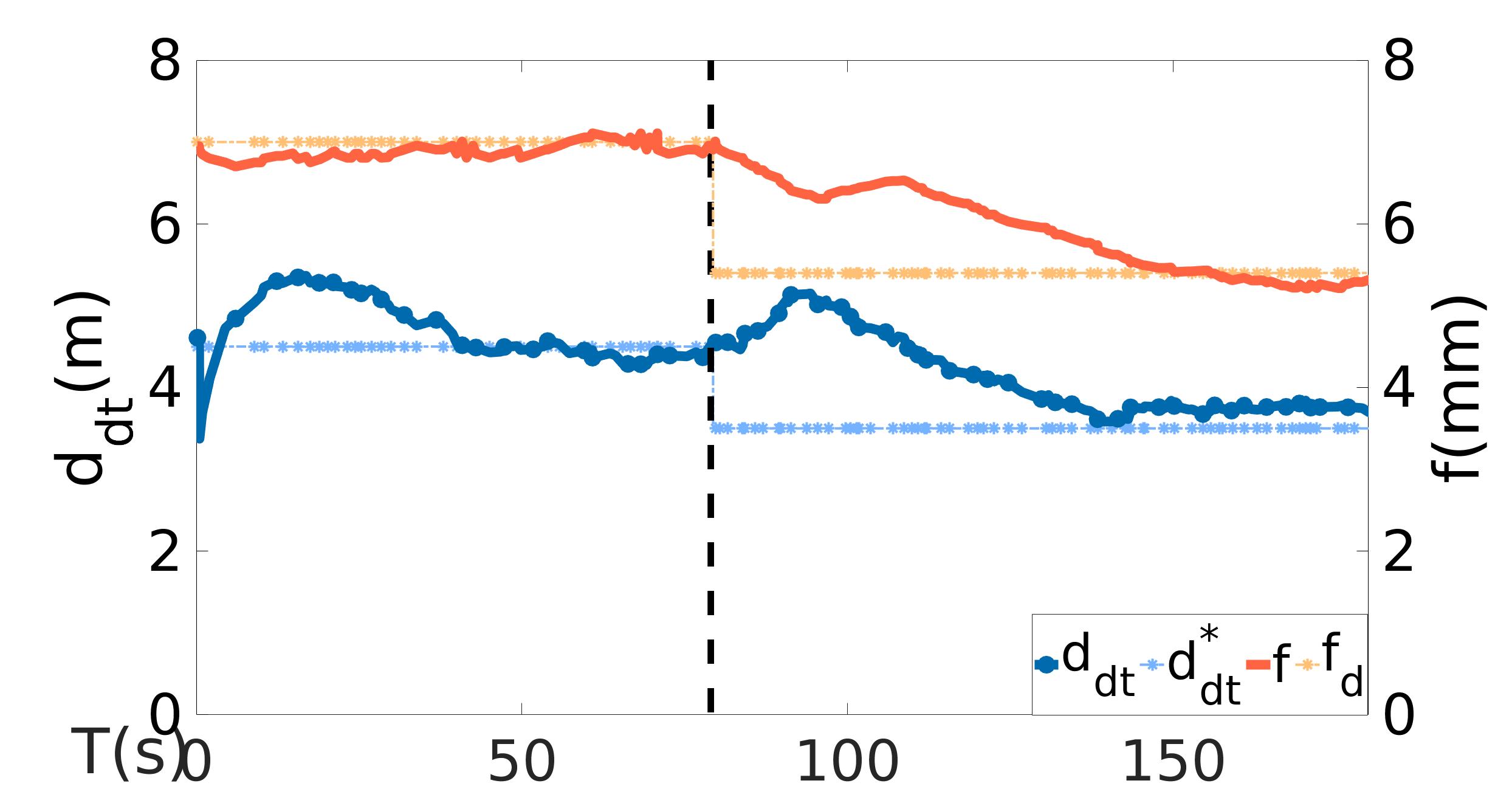} 
    \\\footnotesize (c) & \footnotesize (d)
    \\
\includegraphics[width=0.46\columnwidth,height=2.4cm]{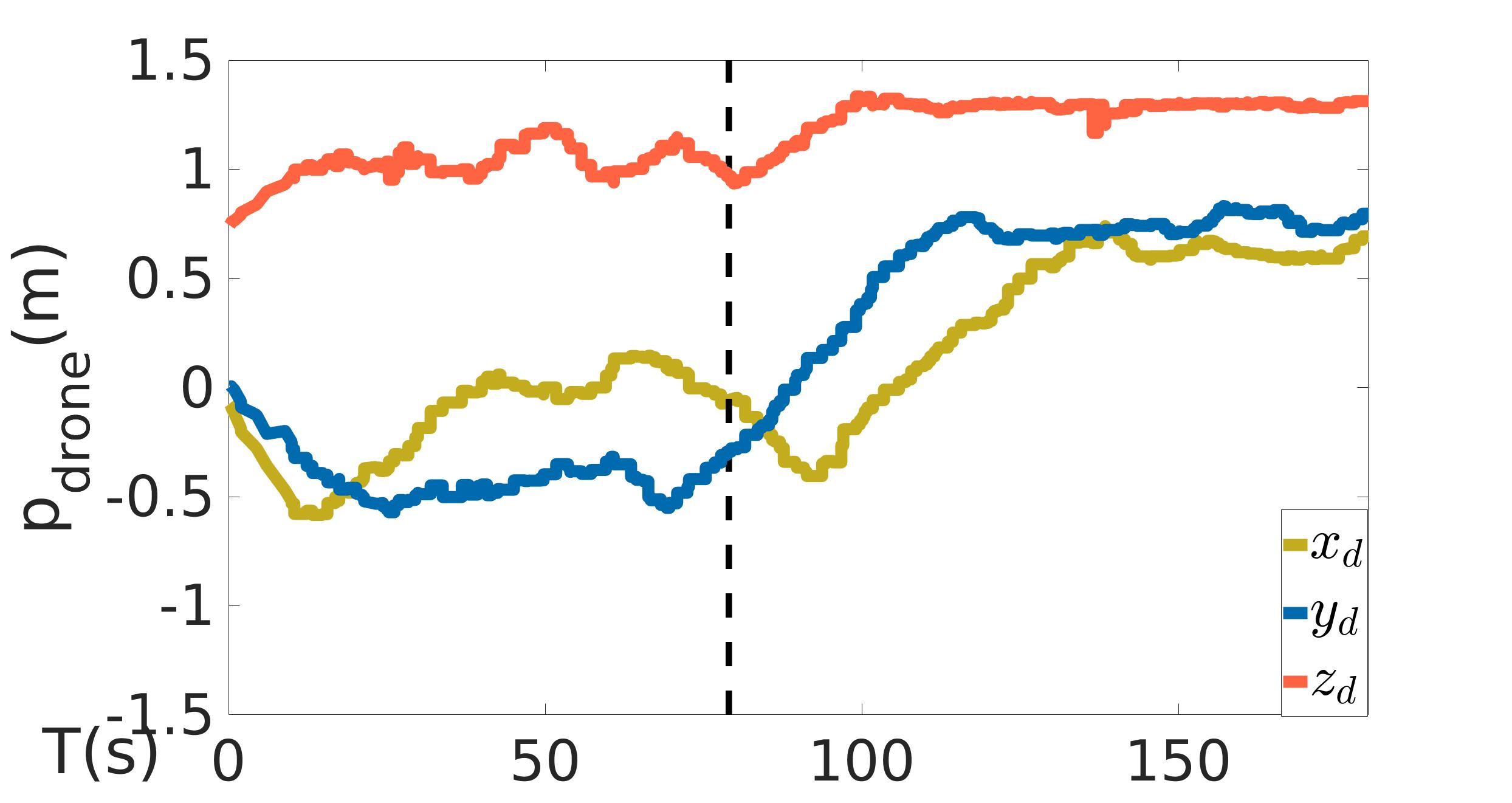}
    & 
    \includegraphics[width=0.46\columnwidth,height=2.4cm]{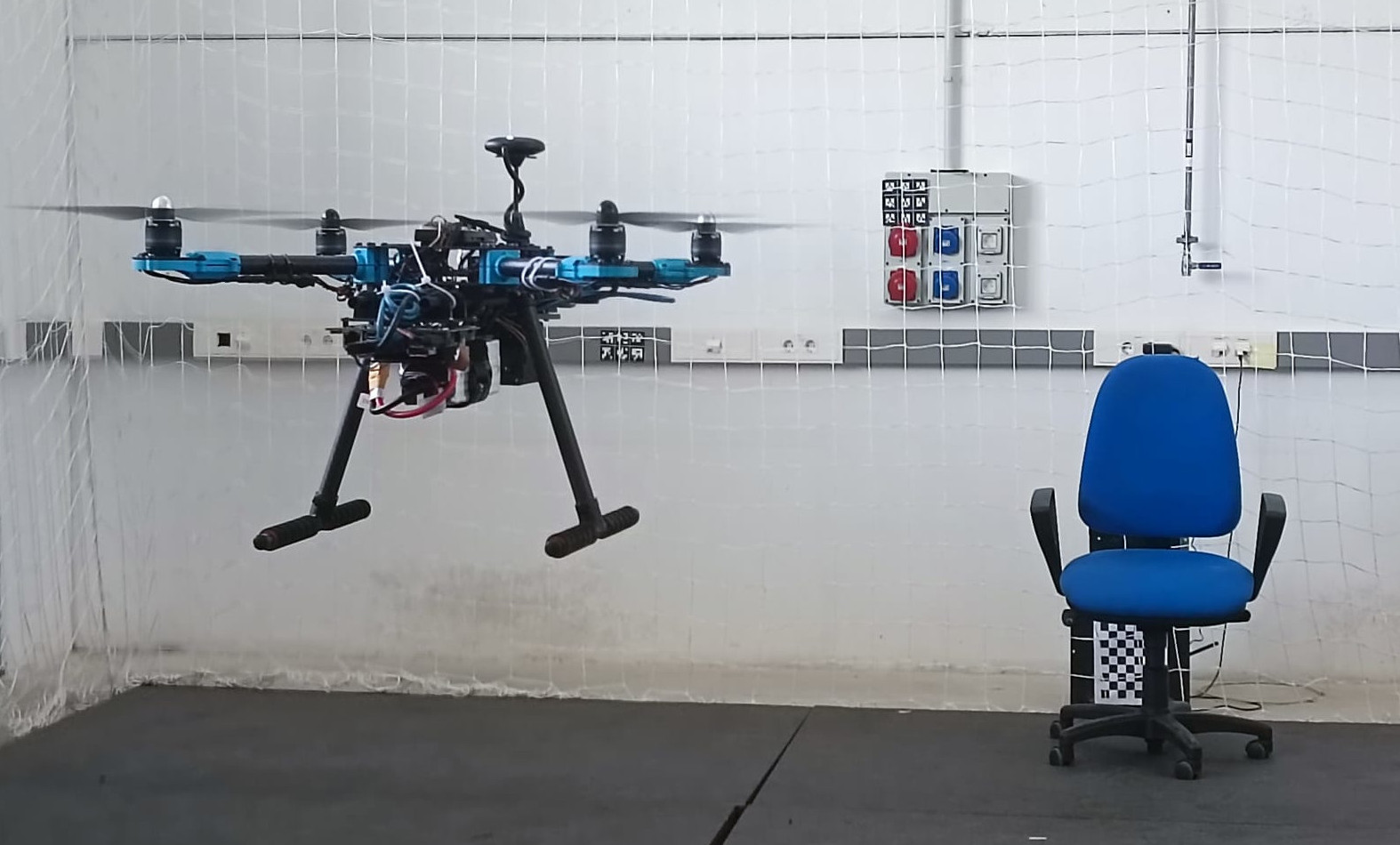} 
    \\\footnotesize (e) & \footnotesize (f)
 
\end{tabular}
\caption{\textbf{Experiment 6: Full platform test with a flying Cinematographic Drone - Assessing performance and integration.}  Lighter lines represent desired
values and darker lines represent actual values.  (a,b) Frames captured by the cinematographic camera
in each sequence. (c) Image position of the target ($\mathbf{im}_{t}$) and desired value ($\mathbf{im}_{t}^*$). The top blue line represents the horizontal pixel and the bottom orange line depicts the vertical pixel. (d) Focal length ($f$), in orange, and desired and actual distance drone-target ($d_{dt}$), in blue. (e) Position of the drone ($\mathbf{p}_d$), where $x$ is yellow, $y$ is blue, and $z$ is orange. (f) Third-person view of platform and target while experimenting. }

\label{fig:real_extrinsincs}
\end{figure}

Figure \ref{fig:real_extrinsincs} presents both qualitative and quantitative results of this experiment. Figures \ref{fig:real_extrinsincs}-a and Fig.\ref{fig:real_extrinsincs}-b show a frame of the recording for each sequence. The bounding box of the target, as well as the horizontal and vertical guidelines denoting its desired position in the image, are depicted for clarification. Third-person views of the experiment with the drone in the air are illustrated in Fig \ref{fig:real_extrinsincs}-f. For quantitative results, Fig \ref{fig:real_extrinsincs}-c depicts the actual and desired image position of the target. The plot of Fig.\ref{fig:real_extrinsincs}-d shows the focal length and the actual and desired distance to the target. Finally, Fig.\ref{fig:real_extrinsincs}-e depicts the evolution of the actual position of the drone. These plots demonstrate how CineMPC modifies the extrinsic parameters of the drone and the focal length of the camera to place the target in the desired image position while satisfying the other coupled requisites.

\begin{table}[!h]
\begin{center}
\caption{\textbf{Computational load}. Mean and standard deviation of the execution times (in ms) of the perception and control modules}
\scriptsize{
\begin{tabular}{|c|c|c|}
\hline
&mean & std    \\
    \hline
Perception & 373 & 104   \\
    \hline 
Control & 258 & 84    \\
\hline
\end{tabular}
}
\label{tab:comp_load}
\end{center}
\end{table}

\subsubsection{Experiment 7 (E7) - Computational load}
This experiment demonstrates the feasibility of using CineMPC on a real cinematographic platform. We installed and ran CineMPC on an Nvidia Jetson AGX Xavier. 
 Utilizing a pre-recorded rosbag containing all camera images and drone state data (as described in Section \ref{sec:exp_setup}), we executed the entire pipeline on the aforementioned board. 
To quantify the computational load, we measured the time required by the control and perception modules on this board, as these modules consume the most computational resources within the pipeline. The results (mean and standard deviation) are depicted in Table \ref{tab:comp_load}.
The sum of these two times is 631 ms. To ensure that the drone and camera execute the correct trajectory, the calculation time of each iteration should be shorter than the time horizon $\Delta_T N$ which is 2500 ms in the real experiments. The gap between the time horizon and computational time affirms the algorithm's feasibility for onboard execution.

%% file: 09_Future.tex
CineMPC paves the way for new interesting and challenging problems in the context of autonomous cinematography.
For example, on the control side, there is the issue of defining suitable values of $\bm\mu$ to achieve the different cinematographic effects.
Currently, this requires human expertise, but it is something that could be automated as well.
Some works like \cite{gebhardt2016airways, lan2017xpose} approach this problem and present novel user interfaces that are more intuitive and visual, using selectors or images. Another interesting approach could be the use of data-driven techniques like Imitation or Reinforcement Learning, partly explored in \cite{gschwindt2019can,huang2021one, jiang2021camera, bonatti2021batteries}, to determine all the intrinsic and extrinsic parameters from experts or existing footage.
Consideration of the intrinsic parameters in a multi-drone setup is also something worth analyzing when compared to existing multi-drone approaches. 
CineMPC also raises new questions in the field of perception.
The proposed perception module focuses on providing the necessary information for the controller to work, but it still relies on some human input.
For instance, the current module is only able to estimate the orientation of moving targets and needs a preliminary orientation of static ones ($t_R$). The human also needs to specify the type of target to film and is limited to those available in Yolo.
More advanced segmentation techniques~\cite{openpose} could be used to overcome these limitations, but the increase of computational load should also be considered.

%% file: 10_Conclusions.tex
We have presented CineMPC, a complete cinematographic platform that uses perception to track multiple types of targets from RGB-D thin-lens produced images and implements a model predictive control approach to control the intrinsic and extrinsic parameters of a camera for autonomous cinematography. This is the first approach to date to include the intrinsic information in this kind of control.  

We have described the main modules of the system, namely the perception and control modules and the cinematographic agents, and the role that they play in the cinematographic platform. The perception module is able to localize the targets that are present in the images taken by a thin-lens camera, that can be blurred or distorted, and extract their position and orientation that are filtered and processed with a Kalman Filter. The control module is implemented inside an MPC framework whose cost function includes four different cost terms to achieve several artistic guidelines, like the depth of field and artistic composition of the image, the relative recording pose and canonical shots, and desired focal length. The optimization of these terms returns camera control values that generate semantically expressive images, closer to the ones seen in actual movies. We also release the system implementation of all these features
to the community.
A variety of experiments have been used to illustrate the potential of CineMPC in photorealistic simulation and in a real setup, successfully considering different guidelines and kinds of shots and a variety of targets in nature and dynamics.